\pdfoutput=1
\documentclass{article}

\def\CLEANVERSION{1}

\usepackage{iclr2027_conference,times}

\iclrfinalcopy

\usepackage[utf8]{inputenc} 
\usepackage[T1]{fontenc}    
\usepackage{graphicx}       
\usepackage{hyperref}       
\usepackage{url}            
\usepackage{booktabs}       
\usepackage{amsfonts}       
\usepackage{nicefrac}       
\usepackage{microtype}      
\usepackage[table]{xcolor}  
\usepackage{multirow}       

\usepackage{amsmath}   
\usepackage{enumitem}  
\usepackage{listings}     
\usepackage{longtable}    
\usepackage{pdflscape}    
\usepackage{array}        
\usepackage{arydshln}     
\usepackage{float}        
\usepackage{tikz}
\usepackage[normalem]{ulem} 

\usepackage{algpseudocode}
\usepackage[ruled,vlined]{algorithm2e}

\usepackage{caption}

\usepackage{subcaption}

\usepackage{tcolorbox}
\tcbuselibrary{breakable} 

\usepackage{wrapfig} 
\usepackage{needspace} 

\usetikzlibrary{tikzmark,calc}

\definecolor{dropPurpleCell}{HTML}{D6E4FF}
\definecolor{riseBlueCell}{HTML}{E7D0FF}
\definecolor{conflictOrange}{HTML}{D97706}
\definecolor{citationOrange}{HTML}{B86A3C}
\definecolor{revisionViolet}{HTML}{6F2DA8}
\definecolor{revisionGreen}{HTML}{2E7D32}
\hypersetup{
  colorlinks=true,
  citecolor=citationOrange,
  linkcolor=black,
  urlcolor=blue
}
\ifdefined\CLEANVERSION
  \newcommand{\revision}[1]{\textcolor{black}{#1}}
  \newcommand{\postreviewrevision}[1]{\textcolor{black}{#1}}
  \newcommand{\greenrevision}[1]{\textcolor{black}{#1}}
  \newcommand{\unchangednumber}[1]{{\color{black}#1}}
  \newcommand{\postreviewnumber}[1]{{\color{black}#1}}
  \newcommand{\greennumber}[1]{{\color{black}#1}}
  \newcommand{\revisiontablecolor}{\color{black}}
  
  \newcommand{\revisioncaptionstyle}{\captionsetup{labelfont={color=black},textfont={color=black}}}
  \newcommand{\postreviewcaptionstyle}{\captionsetup{labelfont={color=black},textfont={color=black}}}
\else
  \newcommand{\revision}[1]{\textcolor{red}{#1}}
  \newcommand{\postreviewrevision}[1]{\textcolor{red}{#1}}
  \newcommand{\greenrevision}[1]{\textcolor{red}{#1}}
  \newcommand{\unchangednumber}[1]{{\color{black}#1}}
  \newcommand{\postreviewnumber}[1]{{\color{black}#1}}
  \newcommand{\greennumber}[1]{{\color{black}#1}}
  \newcommand{\revisiontablecolor}{\color{black}}
  
  \newcommand{\revisioncaptionstyle}{\captionsetup{labelfont={color=black},textfont={color=black}}}
  \newcommand{\postreviewcaptionstyle}{\captionsetup{labelfont={color=black},textfont={color=red}}}
\fi

\ifdefined\CLEANVERSION
  \newcommand{\scorerevision}[1]{\textcolor{black}{#1}}
\else
  \newcommand{\scorerevision}[1]{\textcolor{red}{#1}}
\fi

\newcommand{\fictitiousterm}{fictitious}
\newcommand{\Fictitiousterm}{Fictitious}
\newcommand{\fictitiousmath}{\mathrm{fictitious}}
\newcommand{\fictitiousidentifier}[1]{\texttt{fictitious}\texttt{#1}}

\usepackage{tabularx,pifont}

\newcommand{\benchmarkname}{MemoReason}
\newcommand{\benchmarkfullname}{\textbf{Memo}rization in \textbf{Reason}ing}
\newcommand{\ResultMaxSignificantVariantDrop}{15.7}

\newcommand{\ResultTemporalSignificantModels}{5}
\newcommand{\ResultTemporalDropMin}{4.4}
\newcommand{\ResultTemporalDropMax}{15.6}
\newcommand{\ResultArithmeticSignificantModels}{6}
\newcommand{\ResultArithmeticDropMin}{4.7}
\newcommand{\ResultArithmeticDropMax}{7.2}

\newcommand{\ResultVariantExtractiveModels}{9}
\newcommand{\ResultVariantExtractiveSignificantModels}{6}
\newcommand{\ResultShortcutReasoningMin}{1.0}
\newcommand{\ResultShortcutReasoningMax}{3.2}
\newcommand{\ResultShortcutExtractiveMin}{0.0}
\newcommand{\ResultShortcutExtractiveMax}{1.0}

\newcommand{\ResultPartialRecoveryMin}{0.07}
\newcommand{\ResultPartialClearRecoveryMin}{0.42}
\newcommand{\ResultPartialRecoveryMax}{0.87}
\newcommand{\ResultPartialModelCount}{7}
\newcommand{\ResultClaudePartialFactual}{95.67}

\newcommand{\ResultClaudePartialMinimum}{91.88}
\newcommand{\ResultClaudePartialMinimumPercent}{80}

\newcommand{\ResultClaudePartialRecovery}{0.42}
\newcommand{\ResultEffortMediumMinusLowFactual}{3.50}
\newcommand{\ResultEffortMediumMinusLowFictional}{2.25}
\newcommand{\ResultEffortMediumMinusHighFactual}{1.58}
\newcommand{\ResultEffortMediumMinusHighFictional}{1.59}

\newcommand{\ResultAllQuestionFlipModelCount}{9}
\newcommand{\ResultAllQuestionFlipSignificantPositiveCount}{7}
\newcommand{\ResultReverseModelCount}{4}
\newcommand{\ResultReverseShortcutReasoningMin}{1.95}
\newcommand{\ResultReverseShortcutReasoningMax}{3.22}
\newcommand{\ResultReverseShortcutAtomicMax}{5.00}
\newcommand{\ResultFrequencyPearson}{-0.103}
\newcommand{\ResultFrequencyPearsonLow}{-0.252}
\newcommand{\ResultFrequencyPearsonHigh}{0.028}

\newcommand{\ResultTemperatureGapChangeMin}{0.03}
\newcommand{\ResultTemperatureGapChangeMax}{0.29}

\newcommand{\ResultTemperatureInconclusiveContrastCount}{4}
\newcommand{\ResultTemperatureSignificantGapCellCount}{6}

\DeclareRobustCommand{\memoreasonbrain}{%
  \raisebox{-0.12em}{\includegraphics[height=0.95em]{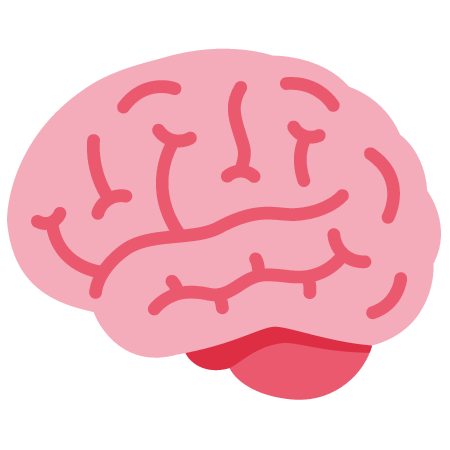}}%
}
\DeclareRobustCommand{\leaderboardicon}[1]{%
  \raisebox{-0.12em}{\csname leaderboardimage#1\endcsname}%
}
\expandafter\newcommand\csname leaderboardimagetrophy\endcsname{\includegraphics[height=1em]{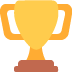}}
\expandafter\newcommand\csname leaderboardimagegold_medal\endcsname{\includegraphics[height=1em]{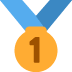}}
\expandafter\newcommand\csname leaderboardimagesilver_medal\endcsname{\includegraphics[height=1em]{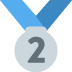}}
\expandafter\newcommand\csname leaderboardimagebronze_medal\endcsname{\includegraphics[height=1em]{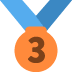}}
\newcommand{\modellogo}[1]{%
  \makebox[1.15em][c]{\raisebox{-0.15em}{%
    \csname modelimage#1\endcsname}}%
  \hspace{0.25em}%
}
\expandafter\newcommand\csname modelimageanthropic\endcsname{\includegraphics[width=1em,height=1em,keepaspectratio]{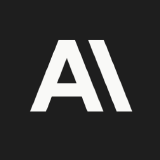}}
\expandafter\newcommand\csname modelimageopenai\endcsname{\includegraphics[width=1em,height=1em,keepaspectratio]{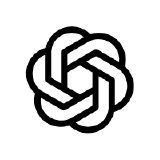}}
\expandafter\newcommand\csname modelimageqwen\endcsname{\includegraphics[width=1em,height=1em,keepaspectratio]{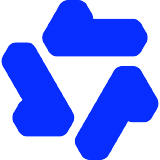}}
\expandafter\newcommand\csname modelimagegoogle-deepmind\endcsname{\includegraphics[width=1em,height=1em,keepaspectratio]{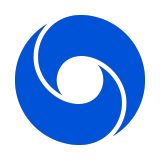}}
\expandafter\newcommand\csname modelimagemeta-llama\endcsname{\includegraphics[width=1em,height=1em,keepaspectratio]{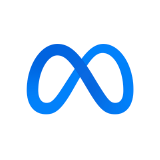}}
\expandafter\newcommand\csname modelimageai2\endcsname{\includegraphics[width=1em,height=1em,keepaspectratio]{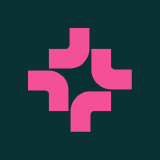}}
\title{\texorpdfstring{\benchmarkname\,\memoreasonbrain}{MemoReason}: Evaluating the Effect of Parametric Memory on Contextual Reasoning in LLMs}

\newcommand{\memoreasonwebsiteurl}{https://memoreason.github.io/}
\newcommand{\memoreasondataseturl}{https://huggingface.co/datasets/zineddine/MemoReason}
\newcommand{\memoreasoncodeurl}{https://github.com/Zineddine-Tighidet/MemoReason}
\definecolor{resourceBlue}{HTML}{169BC4}
\DeclareRobustCommand{\websiteicon}{%
  \tikz[x=1.15em,y=1.15em,baseline=0.2em]{%
    \draw[resourceBlue,line width=0.55pt] (0.5,0.5) circle[radius=0.5];
    \draw[resourceBlue,line width=0.55pt] (0.5,0.5) ellipse[x radius=0.23,y radius=0.5];
    \draw[resourceBlue,line width=0.55pt] (0,0.5) -- (1,0.5);
    \draw[resourceBlue,line width=0.45pt] (0.06,0.27) .. controls (0.33,0.20) and (0.67,0.20) .. (0.94,0.27);
    \draw[resourceBlue,line width=0.45pt] (0.06,0.73) .. controls (0.33,0.80) and (0.67,0.80) .. (0.94,0.73);
  }%
}
\newcommand{\memoreasonresources}{%
  \begingroup
  \normalfont\small
  \hypersetup{urlcolor=resourceBlue}%
  \begin{tabular*}{\dimexpr\textwidth-2\tabcolsep\relax}{@{}l@{\extracolsep{\fill}}l l@{}}
    \websiteicon\,:\;\href{\memoreasonwebsiteurl}{\texttt{memoreason.github.io}} &
    \raisebox{-0.2em}{\includegraphics[height=1.45em]{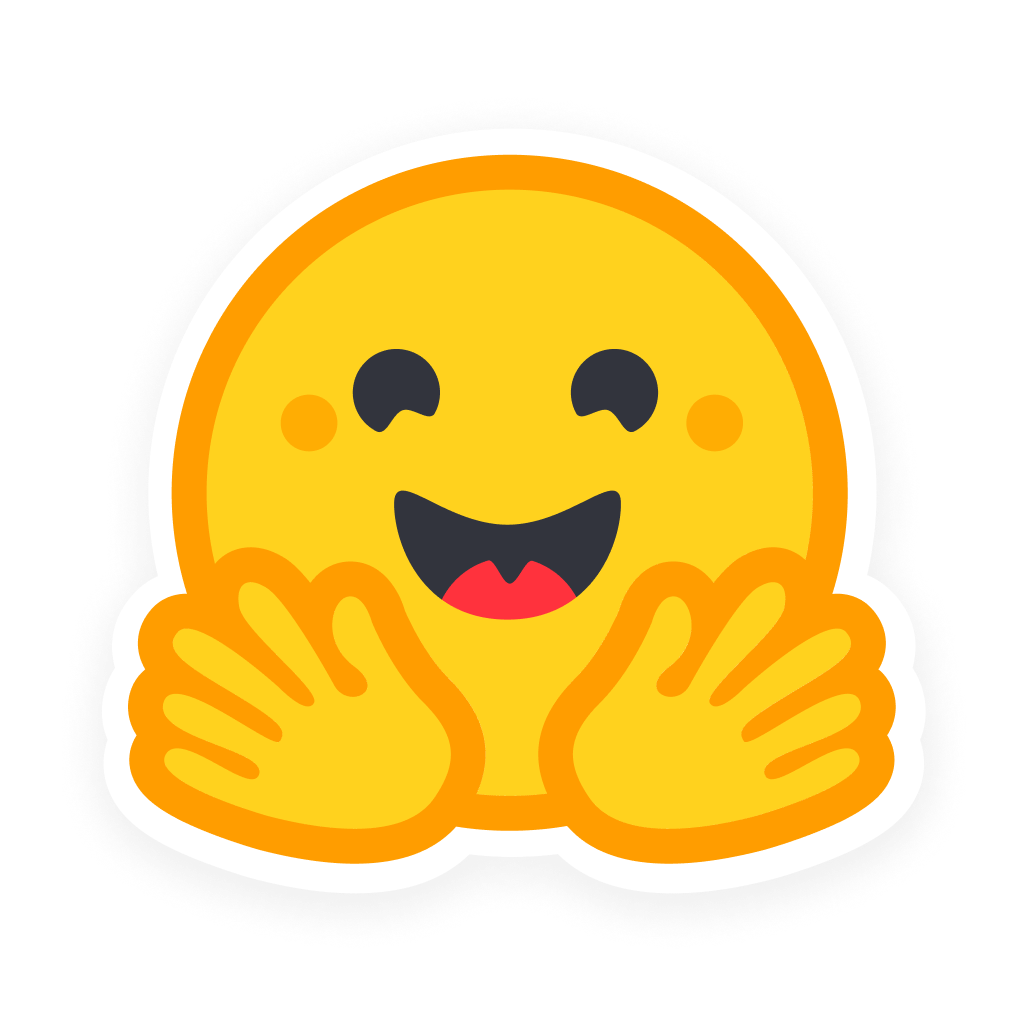}}\;\href{\memoreasondataseturl}{\texttt{MemoReason}} &
    \raisebox{-0.2em}{\includegraphics[height=1.15em]{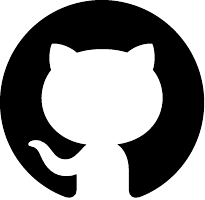}}\;\href{\memoreasoncodeurl}{\texttt{Zineddine-Tighidet/MemoReason}}
  \end{tabular*}%
  \endgroup
}

\author{%
  Zineddine Tighidet\textsuperscript{1,2}%
  \thanks{Corresponding author: zineddine.tighidet@sorbonne-universite.com}%
  \qquad Andrea Mogini\textsuperscript{1}%
  \qquad Jiali Mei\textsuperscript{1}\\
  \textbf{Patrick Gallinari}\textsuperscript{2,3}%
  \qquad\textbf{Benjamin Piwowarski}\textsuperscript{2}\\
  \textsuperscript{1}BNP Paribas\quad
  \textsuperscript{2}Sorbonne Université\quad
  \textsuperscript{3}Criteo AI Lab\\
  {\normalfont Paris, France}\\[0.8ex]
  \memoreasonresources
}

\begin{document}

\maketitle
\lhead{MemoReason\,\memoreasonbrain: \underline{Memo}rization in \underline{Reason}ing}

\begin{abstract}

Large Language Models (LLMs) perform well on reasoning benchmarks, but it remains unclear whether this reflects genuine contextual reasoning or reliance on facts memorized in their parameters. We investigate this by distinguishing two possibilities: a broad \textit{memorization bias}, where familiar content improves reasoning performance, and the \textit{Strong Parametric Shortcut Hypothesis}, where models skip reasoning entirely and recall stored answers. To test these effects, we introduce \textbf{MemoReason}, a human-curated benchmark that pairs factual reasoning tasks with structurally identical \fictitiousterm{} versions where real entities like people, companies, or dates are systematically replaced by \fictitiousterm{} ones of the same type. This \scorerevision{preserves task structure and specified reasoning operations} while varying the familiarity of the context, allowing controlled measurement of how the parametric memory affects reasoning. \revision{Our evaluation of recent LLMs reveals consistent and statistically significant performance drops of up to \unchangednumber{\ResultMaxSignificantVariantDrop\%} in the fictitious setting, demonstrating a clear memorization bias.} However, a targeted analysis of \revision{questions failed in the fictitious setting} shows that models rarely respond with the corresponding factual answer, indicating that direct parametric shortcuts are not the dominant failure mode. These findings suggest that parametric memory influences reasoning through mechanisms more complex than simple factual recall. \textbf{\benchmarkname} provides a controlled framework for studying these mechanisms and for extending paired factual--\fictitiousterm{} evaluation to broader reasoning settings.

\end{abstract}

\section{Introduction}

Large Language Models (LLMs) have demonstrated remarkable performance in complex reasoning tasks by scoring highly on benchmarks involving math \citep{hendrycks2021measuring, lightman2023lets, alshammari2026mathnet}, code generation \citep{chen2021codex, jimenez2024swebench, jain2024livecodebench, zhuo2025bigcodebench, ye2025verina}, commonsense inference \citep{zellers-etal-2019-hellaswag, talmor-etal-2019-commonsenseqa}, multi-hop question answering \citep{yang-etal-2018-hotpotqa} and professional domain exams such as law and medicine \citep{10.1098/rsta.2023.0254, nori2023capabilitiesgpt4medicalchallenge}. 
However, the extent to which this performance is due to abstract reasoning, pattern matching or factual memorization remains an open question. In this work, we propose an approach to better elucidate this question by studying LLM performance on reasoning tasks under controlled contextual information environments.

\textbf{memorization bias in Reasoning \& the Strong Parametric Shortcut Hypothesis.} Some studies show that, when performing reasoning tasks, LLMs can over-rely on their parametric memory -- i.e. the knowledge stored in the model parameters during training -- rather than on contextual knowledge \citep{longpre-etal-2021-entity, wu-etal-2024-reasoning}. In this work, we begin by testing whether familiarity with the contextual information being processed biases the ability of a model to perform reasoning tasks. This relates to the question: \textit{what is the impact of parametric memory on the \textbf{reasoning performance} of LLMs?} If memory plays an important role, reasoning performance should degrade when the model is exposed to new information that has similar textual patterns to the training data but fundamentally different meaning. Henceforth we will refer to such information as \textit{\fictitiousterm{}} (as opposed to \textit{factual} information).

Additionally, we explore the \textit{Strong Parametric Shortcut Hypothesis}: the idea that LLMs leverage embedded memory as an easy way out of reasoning tasks. In short, the hypothesis states that, rather than performing all the intermediary reasoning steps logically necessary to arrive at the final answer (e.g., arithmetic, relational inference, temporal reasoning), models proceed by wrongly recalling a memorized response.

For mathematical reasoning, works such as \texttt{GSM-Symbolic} \citep{mirzadeh2025gsmsymbolic} and \texttt{MATH()} \citep{srivastava2024functionalbenchmarksrobustevaluation} have successfully isolated true reasoning from memorization (due to benchmark leakage into training data) by replacing entities within fixed problem templates, thereby maintaining the same task structure and complexity.

For reasoning tasks over semantically-rich documents, recent literature has sought to investigate shortcuts \citep{glockner-etal-2025-neoqa} and \textit{memorization biases} \citep{wu2024cofcastepwisecounterfactualmultihop}. In the case of the \texttt{CofCa} benchmark by \cite{wu2024cofcastepwisecounterfactualmultihop}, a difference in reasoning performance between factual and \fictitiousterm{} examples was reported. However, the gap they reported could be fully explained by the unaccounted systematic error introduced by the fact they used two \textbf{distinct} sets of evidence-question pairs for the factual and the \fictitiousterm{} data. Hence, a strong conclusion on the role of parametric memory in reasoning cannot be drawn from their work.

In this paper, we set out to close this gap by applying the robust approach by \cite{mirzadeh2025gsmsymbolic} to semantically-rich documents. We do so by introducing \textbf{\benchmarkname} (\textbf{\benchmarkfullname}), a novel, high quality human-curated benchmark designed specifically to test \textit{memorization biases} in reasoning and the \textit{Strong Parametric Shortcut Hypothesis} under strictly controlled conditions.

We evaluate LLMs on reasoning tasks using known factual entities and subsequently test them on the exact same tasks where those familiar entities are replaced with \fictitiousterm{} ones under rule constraints. To the best of our knowledge, we are the first to propose a benchmark that compares LLM performance on a complex, human-curated, multi-domain question-answering dataset under these controlled conditions. We make the following main contributions:
\begin{itemize}
\item We introduce \textbf{\benchmarkname}, a novel reasoning question-answering benchmark that pairs factual tasks with structurally identical \fictitiousterm{} counterparts.
\item We formulate a rigorous framework to evaluate how much LLMs rely on contextual reasoning and parametric shortcuts using this novel benchmark.
\item We open-source the customized annotation interface used to build \textbf{\benchmarkname} as a commitment to help future work aiming to extend our approach.
\item We show that LLMs exhibit significant performance drops on reasoning questions when moving from factual to \fictitiousterm{} settings.
\item We verify that direct factual information recall (the \textit{Strong Parametric Shortcut Hypothesis}) is not the dominant memorization bias in recent LLMs
\end{itemize}

\section{Benchmark Construction}

To examine familiarity biases in reasoning, we define a setup that allows us to systematically isolate the parametric memory without changing the underlying reasoning. To this end, we start by consolidating a factual dataset based on Wikipedia excerpts (Section \ref{subsec:factual_dataset_construction}). We then use this dataset to build templates that preserve the text structure while abstracting its content (Section \ref{subsec:template_annotation_process}). Finally, we generate \fictitiousterm{} variants of the factual dataset (Section \ref{subsec:fictional_dataset_generation}).

\subsection{Factual Dataset}
\label{subsec:factual_dataset_construction}

\revision{We start by building a factual dataset $\mathcal{D} = \{(d_i, q_{i}, a_{i})\}_{i=1}^{N}$ where $d_i$ is an excerpt~(introduction section) of a Wikipedia article, and $(q_{i}, a_{i})$ a question-answer tuple based on $d_i$.} \revision{Similarly to \texttt{GSM-Symbolic}, which instantiates \unchangednumber{100} templates, \texttt{MemoReason} consists of 100 semantically-rich template documents with 12 corresponding question-answer (Q-A) tasks which results in N=1,200 factual tasks that are later paired with fictitious counterparts. We report in Table~\ref{tab:theme-dataset-stats} detailed statistics about \texttt{MemoReason}.}

We chose Wikipedia excerpts treating famous topics, as it is reasonable to assume that most LLMs were exposed to them during training. The selected documents contain a high density of entities, making them ideal for probing the parametric memory of LLMs. Finally, the dataset spans nine diverse themes for which the statistics are reported in Table \ref{tab:theme-dataset-stats} (additional details available in Appendix \ref{sec:dataset-themes}).

The question-answer pairs associated with each document are chosen to cover four \textit{question classes} and three \textit{answer classes} (see Section \ref{subsec:fictional_dataset_generation}). 


\begin{figure*}[t]
\centering
\begingroup
\scriptsize
\setlength{\tabcolsep}{0pt}

\definecolor{mrInk}{RGB}{31,41,55}
\definecolor{mrMuted}{RGB}{100,116,139}
\definecolor{mrFactBg}{RGB}{255,255,255}
\definecolor{mrFictBg}{RGB}{255,255,255}
\definecolor{mrFactBorder}{RGB}{207,219,234}
\definecolor{mrFictBorder}{RGB}{216,207,228}
\definecolor{mrProbe}{RGB}{222,208,248}
\definecolor{mrProbeText}{RGB}{123,103,161}
\definecolor{mrLander}{RGB}{209,233,219}
\definecolor{mrLanderText}{RGB}{95,129,113}
\definecolor{mrAgency}{RGB}{205,228,248}
\definecolor{mrAgencyText}{RGB}{93,123,157}
\definecolor{mrComet}{RGB}{245,222,190}
\definecolor{mrCometText}{RGB}{155,126,81}
\definecolor{mrAstA}{RGB}{244,214,200}
\definecolor{mrAstAText}{RGB}{161,117,99}
\definecolor{mrAstB}{RGB}{219,234,200}
\definecolor{mrAstBText}{RGB}{122,136,96}
\definecolor{mrProgram}{RGB}{229,214,239}
\definecolor{mrProgramText}{RGB}{136,108,151}
\definecolor{mrDate}{RGB}{209,222,238}
\definecolor{mrDateText}{RGB}{109,121,142}
\definecolor{mrOrdinal}{RGB}{216,225,236}
\definecolor{mrOrdinalText}{RGB}{113,121,132}
\definecolor{mrCost}{RGB}{221,228,203}
\definecolor{mrCostText}{RGB}{123,131,98}
\definecolor{mrEntA}{RGB}{225,203,249}
\definecolor{mrEntB}{RGB}{247,208,219}
\definecolor{mrEntC}{RGB}{248,218,186}
\definecolor{mrEntC2}{RGB}{248,180,180}
\definecolor{mrEntD}{RGB}{199,232,248}
\definecolor{mrEntE}{RGB}{255,235,157}
\definecolor{mrEntF}{RGB}{211,236,201}
\definecolor{mrEntG}{RGB}{177,222,240}
\definecolor{mrEntH}{RGB}{163,195,243}
\definecolor{mrEntI}{RGB}{239,197,244}
\definecolor{mrEntJ}{RGB}{191,240,229}
\definecolor{mrEntK}{RGB}{190,214,245}
\definecolor{mrEntL}{RGB}{178,233,200}
\definecolor{mrEntM}{RGB}{249,120,208}
\definecolor{mrEntN}{RGB}{255,210,137}
\definecolor{mrEntO}{RGB}{255,194,148}
\definecolor{mrEntP}{RGB}{157,229,234}
\definecolor{mrEntQ}{RGB}{218,189,248}
\definecolor{mrEntR}{RGB}{212,190,238}
\definecolor{mrEntS}{RGB}{247,194,214}
\definecolor{mrEntT}{RGB}{255,223,194}
\definecolor{mrEntU}{RGB}{223,241,171}

\newcommand{\mrchip}[2]{%
  \tikz[baseline=(mrnode.base)]\node[
    fill=#1,
    rounded corners=1.1pt,
    inner xsep=1.3pt,
    inner ysep=0.15pt,
    text=black,
    font=\ttfamily\fontsize{6.8}{6.8}\selectfont
  ] (mrnode) {\strut #2};%
}
\newcommand{\mrtxt}[2]{{\ttfamily\color{#1}#2}}
\newcommand{\mrref}[2]{\mrchip{#1}{#2}}

\newcommand{\Probe}[1]{\mrchip{mrProbe}{#1}}
\newcommand{\ProbeText}[1]{\mrtxt{mrProbeText}{#1}}
\newcommand{\Lander}[1]{\mrchip{mrLander}{#1}}
\newcommand{\LanderText}[1]{\mrtxt{mrLanderText}{#1}}
\newcommand{\Agency}[1]{\mrchip{mrAgency}{#1}}
\newcommand{\AgencyText}[1]{\mrtxt{mrAgencyText}{#1}}
\newcommand{\Comet}[1]{\mrchip{mrComet}{#1}}
\newcommand{\CometText}[1]{\mrtxt{mrCometText}{#1}}
\newcommand{\AstA}[1]{\mrchip{mrAstA}{#1}}
\newcommand{\AstAText}[1]{\mrtxt{mrAstAText}{#1}}
\newcommand{\AstB}[1]{\mrchip{mrAstB}{#1}}
\newcommand{\AstBText}[1]{\mrtxt{mrAstBText}{#1}}
\newcommand{\Program}[1]{\mrchip{mrProgram}{#1}}
\newcommand{\ProgramText}[1]{\mrtxt{mrProgramText}{#1}}
\newcommand{\Date}[1]{\mrchip{mrDate}{#1}}
\newcommand{\DateText}[1]{\mrtxt{mrDateText}{#1}}
\newcommand{\Ord}[1]{\mrchip{mrOrdinal}{#1}}
\newcommand{\OrdText}[1]{\mrtxt{mrOrdinalText}{#1}}
\newcommand{\Cost}[1]{\mrchip{mrCost}{#1}}
\newcommand{\CostText}[1]{\mrtxt{mrCostText}{#1}}
\newcommand{\NatSite}[1]{\mrchip{mrComet}{#1}}
\newcommand{\NatSiteText}[1]{\mrtxt{mrCometText}{#1}}
\newcommand{\CountryA}[1]{\mrchip{mrAstA}{#1}}
\newcommand{\CountryAText}[1]{\mrtxt{mrAstAText}{#1}}
\newcommand{\CountryB}[1]{\mrchip{mrAstB}{#1}}
\newcommand{\CountryBText}[1]{\mrtxt{mrAstBText}{#1}}
\newcommand{\Vehicle}[1]{\mrchip{mrProgram}{#1}}
\newcommand{\VehicleText}[1]{\mrtxt{mrProgramText}{#1}}
\newcommand{\Count}[1]{\mrchip{mrCost}{#1}}
\newcommand{\CountText}[1]{\mrtxt{mrCostText}{#1}}
\newcommand{\ProgMain}[1]{\mrchip{mrEntA}{#1}}
\newcommand{\ProgAlias}[1]{\mrchip{mrEntB}{#1}}
\newcommand{\Country}[1]{\mrchip{mrEntC}{#1}}
\newcommand{\AgencyRef}[1]{\mrchip{mrEntD}{#1}}
\newcommand{\World}[1]{\mrchip{mrEntE}{#1}}
\newcommand{\Home}[1]{\mrchip{mrEntF}{#1}}
\newcommand{\YearConceived}[1]{\mrchip{mrEntG}{#1}}
\newcommand{\YearCrewed}[1]{\mrchip{mrEntH}{#1}}
\newcommand{\MissionRef}[1]{\mrchip{mrEntI}{#1}}
\newcommand{\VehicleRef}[1]{\mrchip{mrEntJ}{#1}}
\newcommand{\MissionDateRef}[1]{\mrchip{mrEntK}{#1}}
\newcommand{\OrbiterRef}[1]{\mrchip{mrEntL}{#1}}
\newcommand{\RegionRef}[1]{\mrchip{mrEntM}{#1}}
\newcommand{\PlannedRef}[1]{\mrchip{mrEntN}{#1}}
\newcommand{\CanceledRef}[1]{\mrchip{mrEntO}{#1}}
\newcommand{\SuccessfulRef}[1]{\mrchip{mrEntP}{#1}}
\newcommand{\RemainingRef}[1]{\mrchip{mrEntQ}{#1}}
\newcommand{\AbortRef}[1]{\mrchip{mrEntR}{#1}}
\newcommand{\FireRef}[1]{\mrchip{mrEntS}{#1}}
\newcommand{\CrewARef}[1]{\mrchip{mrEntC2}{#1}}
\newcommand{\CrewBRef}[1]{\mrchip{mrEntU}{#1}}

\newcommand{\mrhead}[1]{{\normalfont\bfseries\large\color{mrInk} #1}\par\vspace{2pt}}
\newcommand{\mrsubhead}[1]{{\normalfont\bfseries\color{mrInk} #1}\par\vspace{1.0pt}}
\newcommand{\mrpanelhead}[1]{{\normalfont\bfseries\color{mrInk} #1}\par\vspace{2pt}}
\newcommand{\mrtitle}[1]{{\centering\normalfont\bfseries\fontsize{18}{18}\selectfont\color{mrInk} #1\par}\vspace{4pt}}
\newcommand{\mrdivider}{\par{\color{mrMuted}\rule{\linewidth}{0.25pt}}\par\vspace{1.4pt}}
\newcommand{\docfont}{\ttfamily\fontsize{6.15}{6.7}\selectfont}
\newcommand{\qafont}{\ttfamily\fontsize{6.15}{6.15}\selectfont}
\newcommand{\mrblock}[4]{%
  \begingroup
  \setlength{\fboxsep}{4pt}%
  \setlength{\fboxrule}{0.35pt}%
  \fcolorbox{#2}{#1}{%
    \begin{minipage}[t][#3][t]{\dimexpr\linewidth-2\fboxsep-2\fboxrule\relax}
      #4
    \end{minipage}%
  }%
  \endgroup
}
\newcommand{\mrqatype}[2]{%
  \parbox[t]{\linewidth}{%
    \raggedright
    \textbf{#1}\par
    \vspace{-1.8pt}%
    {\color{mrMuted}\fontsize{5.5}{4.1}\selectfont(#2)}%
  }%
}
\newcommand{\mrqa}[4]{%
  \mrqatype{#1}{#2} &
  \parbox[t]{\linewidth}{%
    \raggedright
    #3\par
    \vspace{1.3pt}%
    {\color{mrMuted}Answer:}~#4%
  }\\[2.2pt]
}
\newcommand{\mrruleline}[1]{#1\\[2.3pt]}
\newlength{\mrDocHeight}
\newlength{\mrQAHeight}
\newlength{\mrTopHeight}
\newlength{\mrBottomHeight}
\setlength{\mrDocHeight}{4.95cm}
\setlength{\mrQAHeight}{2.7cm}
\setlength{\mrTopHeight}{8.75cm}
\setlength{\mrBottomHeight}{3.2cm}

\newcommand{\FactProgram}{\ProgMain{Apollo program}}
\newcommand{\FictProgram}{\ProgMain{Project Heliond}}
\newcommand{\FactAlias}{\ProgAlias{Project Apollo}}
\newcommand{\FictAlias}{\ProgAlias{Stellara 17}}
\newcommand{\FactCountry}{\Country{United States}}
\newcommand{\FictCountry}{\Country{Krostovia}}
\newcommand{\FactAgency}{\AgencyRef{NASA}}
\newcommand{\FictAgency}{\AgencyRef{Ossander Launch Complex}}
\newcommand{\FactWorld}{\World{Moon}}
\newcommand{\FictWorld}{\World{Cerulan Depths}}
\newcommand{\FactWorldsecond}{\World{lunar}}
\newcommand{\FictWorldsecond}{\World{Cerulan Depths}}
\newcommand{\FactHome}{\Home{Earth}}
\newcommand{\FictHome}{\Home{Terrath}}
\newcommand{\FactMission}{\MissionRef{Apollo 11}}
\newcommand{\FictMission}{\MissionRef{Stellara 1}}
\newcommand{\FactConceivedYear}{\YearConceived{1960}}
\newcommand{\FictConceivedYear}{\YearConceived{1950}}
\newcommand{\FactCrewedYear}{\YearCrewed{1968}}
\newcommand{\FictCrewedYear}{\YearCrewed{1981}}
\newcommand{\FactMissionDate}{\MissionDateRef{July 20, 1969}}
\newcommand{\FictMissionDate}{\MissionDateRef{2 January 1982}}
\newcommand{\FactCrewA}{\CrewARef{Neil Armstrong}}
\newcommand{\FictCrewA}{\CrewARef{Halzan Zenkinnor}}
\newcommand{\FactCrewB}{\CrewBRef{Buzz Aldrin}}
\newcommand{\FictCrewB}{\CrewBRef{Korhal Rindor}}
\newcommand{\FactOrbiter}{\OrbiterRef{Michael Collins}}
\newcommand{\FictOrbiter}{\OrbiterRef{Dorvan K. Ellisway}}
\newcommand{\FactRegion}{\RegionRef{lunar}}
\newcommand{\FictRegion}{\RegionRef{Myrgryvale}}
\newcommand{\FactVehicle}{\VehicleRef{Apollo Lunar Module (LM)}}
\newcommand{\FactVehicleAbrrev}{\VehicleRef{Lunar Module}}
\newcommand{\FictVehicle}{\VehicleRef{Stellara Command Vessel}}
\newcommand{\FactCanceled}{\CanceledRef{three}}
\newcommand{\FictCanceled}{\CanceledRef{two}}
\newcommand{\FactPlanned}{\PlannedRef{nine}}
\newcommand{\FictPlanned}{\PlannedRef{six}}
\newcommand{\FactRemaining}{\RemainingRef{six}}
\newcommand{\FictRemaining}{\RemainingRef{four}}
\newcommand{\FactSuccessful}{\SuccessfulRef{five}}
\newcommand{\FictSuccessful}{\SuccessfulRef{three}}
\newcommand{\FactNotCanceled}{\texttt{6} (\FactPlanned{} - \FactCanceled{})}
\newcommand{\FictNotCanceled}{\texttt{4} (\FictPlanned{} - \FictCanceled{})}
\newcommand{\FactConceptionGap}{\texttt{8} (\FactCrewedYear{} - \FactConceivedYear{})}
\newcommand{\FictConceptionGap}{\texttt{31} (\FictCrewedYear{} - \FictConceivedYear{})}
\newcommand{\FactAbortMission}{\AbortRef{Apollo 13}}
\newcommand{\FictAbortMission}{\AbortRef{Stellara 13}}
\newcommand{\FactFireMission}{\FireRef{Apollo 1}}
\newcommand{\FictFireMission}{\FireRef{Stellara 8}}
\newcommand{\FactBeyond}{\texttt{Yes}}
\newcommand{\FictBeyond}{\texttt{Yes}}

\mrtitle{\benchmarkname{}}
\begin{tabularx}{0.985\textwidth}{@{}>{\hsize=1\hsize}X@{\hspace{0.014\textwidth}}>{\hsize=1\hsize}X@{}}

\begin{minipage}[t]{\linewidth}
\mrblock{mrFactBg}{mrFactBorder}{\mrTopHeight}{%
\mrpanelhead{\large Factual}
\begin{minipage}[t][\mrDocHeight][t]{\linewidth}
{\raggedright
\FactProgram{}, also known as \FactAlias{}, was the \FactCountry{} human spaceflight program led by \FactAgency{}, which landed the first humans on the \FactWorld{} in \Date{1969}. It was conceived in \FactConceivedYear{}, and the first crewed flight was in \FactCrewedYear{}. On the \FactMission{} mission, \FactCrewA{} and \FactCrewB{} landed their \FactVehicle{} on \FactMissionDate{}, while \FactOrbiter{} remained in \FactWorldsecond{} orbit in the command and service module (CSM). After the first \FactWorldsecond{} landing, flight hardware remained for \FactPlanned{} follow-on landings, but budget cuts cancelled \FactCanceled{}. \FactSuccessful{} of the remaining \FactRemaining{} missions achieved landings; the \FactAbortMission{} crew used the \FactVehicleAbrrev{} as a ``lifeboat.'' \FactAlias{} sent crewed missions beyond low \FactHome{} orbit. [...] It encountered a major setback in \Date{1967} when the \FactFireMission{} cabin fire killed the entire crew [...]

\color{mrMuted}\textit{Rest of the document omitted.}
}
\end{minipage}\par
\mrdivider
{\raggedright
  \setlength{\tabcolsep}{0.9pt}
  \renewcommand{\arraystretch}{0.84}
  \begin{tabularx}{\linewidth}{@{}>{\normalfont\bfseries\raggedright\arraybackslash}p{0.195\linewidth}X@{}}
\mrqa{Temporal}{Variant Answer}{How many years passed between the conception of \FactAlias{} and its first crewed flight?}{\FactConceptionGap{}}
\mrqa{Arithmetic}{Variant Answer}{How many landings were not canceled?}{\FactNotCanceled{}}
\mrqa{Inference}{Invariant Answer}{Were astronauts able to test the Apollo spacecraft in flight prior to the \FactFireMission{} cabin fire?}{No}
\mrqa{Extractive}{Refusal}{\vspace{0.00001cm}What specifically caused the cabin fire during the \FactFireMission{} incident?}{Cannot be determined}
\end{tabularx}
}}
\end{minipage}
&
\begin{minipage}[t]{\linewidth}
\mrblock{mrFictBg}{mrFictBorder}{\mrTopHeight}{%
\mrpanelhead{\large \Fictitiousterm{}}
\begin{minipage}[t][\mrDocHeight][t]{\linewidth}
{\raggedright
\FictProgram{}, also known as \FictAlias{}, was the \FictCountry{} human spaceflight program led by \FictAgency{}, which landed the first humans on the \FictWorld{} in \Date{1982}. It was conceived in \FictConceivedYear{}, and the first crewed flight was in \FictCrewedYear{}. On the \FictMission{} mission, \FictCrewA{} and \FictCrewB{} landed their \FictVehicle{} on \FictMissionDate{}, while \FictOrbiter{} remained in \FictWorldsecond{} orbit in the command and service module (CSM). After the first \FictWorldsecond{} landing, flight hardware remained for \FictPlanned{} follow-on landings, but budget cuts cancelled \FictCanceled{}. \FictSuccessful{} of the remaining \FictRemaining{} missions achieved landings; the \FictAbortMission{} crew used the \FictVehicle{} as a ``lifeboat.'' \FictAlias{} sent crewed missions beyond low \FictHome{} orbit. [...] It encountered a major setback in \Date{1974} when the \FictFireMission{} cabin fire killed the entire crew [...]

\color{mrMuted}\textit{Rest of the document omitted.}
}
\end{minipage}\par
\mrdivider
{\raggedright
  \setlength{\tabcolsep}{0.9pt}
  \renewcommand{\arraystretch}{0.84}
  \begin{tabularx}{\linewidth}{@{}>{\normalfont\bfseries\raggedright\arraybackslash}p{0.195\linewidth}X@{}}
\mrqa{Temporal}{Variant Answer}{How many years passed between the conception of \FictAlias{} and its first crewed flight?}{\FictConceptionGap{}}
\mrqa{Arithmetic}{Variant Answer}{How many landings were not canceled?}{\FictNotCanceled{}}
\mrqa{Inference}{Invariant Answer}{Were astronauts able to test the spacecraft in flight prior to the \FictFireMission{} cabin fire?}{No}
\mrqa{Extractive}{Refusal}{\vspace{0.00001cm}What specifically caused the cabin fire during the \FictFireMission{} incident?}{Cannot be determined}
\end{tabularx}
}}
\end{minipage}
\\[2pt]
\vspace{-0.55cm}%
\begin{minipage}[t]{\linewidth}
\mrpanelhead{\large Replacements}
\begin{minipage}[t][\mrBottomHeight][t]{\linewidth}
{\docfont\fontsize{5.15}{5.55}\selectfont\raggedright
\setlength{\tabcolsep}{2.2pt}
\renewcommand{\arraystretch}{0.9}
\begin{tabularx}{\linewidth}{@{}l@{\hspace{4pt}}>{\raggedright\arraybackslash}X@{}}
\toprule
Entity reference & factual $\rightarrow$ \fictitiousterm{} \\
\midrule
\mrref{mrEntC2}{person\_1.full\_name} & \FactCrewA{} $\rightarrow$ \raisebox{-6.8pt}{\CrewARef{\shortstack[l]{Halzan\\Zenkinnor}}} \\
\mrref{mrEntG}{temporal\_2.year} & \YearConceived{1960} $\rightarrow$ \YearConceived{1950} \\
\mrref{mrEntC}{place\_1.country} & \FactCountry{} $\rightarrow$ \FictCountry{} \\
\mrref{mrEntH}{temporal\_10.year} & \YearCrewed{1968} $\rightarrow$ \YearCrewed{1981} \\
\mrref{mrEntN}{number\_4.str} & \PlannedRef{nine} $\rightarrow$ \PlannedRef{six} \\
\mrref{mrEntO}{number\_5.str} & \CanceledRef{three} $\rightarrow$ \CanceledRef{two} \\
\mrref{mrEntQ}{number\_7.str} & \RemainingRef{six} $\rightarrow$ \RemainingRef{four} \\
\mrref{mrEntP}{number\_6.str} & \SuccessfulRef{five} $\rightarrow$ \SuccessfulRef{three} \\
\addlinespace[1pt]
\multicolumn{2}{@{}l@{}}{\color{mrMuted}\textit{etc. (other entity replacements omitted)}} \\
\bottomrule
\end{tabularx}
}\vfill
\end{minipage}
\end{minipage}
&
\vspace{-0.55cm}%
\begin{minipage}[t]{\linewidth}
\mrpanelhead{\large Rules}
\begin{minipage}[t][\mrBottomHeight][t]{\linewidth}
{\docfont\raggedright
\setlength{\tabcolsep}{0pt}
\renewcommand{\arraystretch}{1.02}
\begin{tabularx}{\linewidth}{@{}X@{}}
\mrruleline{\mrref{mrEntN}{number\_4.str} - \mrref{mrEntO}{number\_5.str} == \mrref{mrEntQ}{number\_7.str}}
\mrruleline{\mrref{mrEntQ}{number\_7.str} - \texttt{1} == \mrref{mrEntP}{number\_6.str}}
\mrruleline{\mrref{mrEntG}{temporal\_2.year} $\in [1900, 1970]$}
\mrruleline{\mrref{mrEntH}{temporal\_10.year} $\in [1960, 1990]$}
\mrruleline{\mrref{mrEntN}{number\_4.str} $\in [1, 19]$}
\mrruleline{\mrref{mrEntO}{number\_5.str} $\in [1, 13]$}
\vspace{-0.2cm}
\color{mrMuted}\textit{etc. (other rules omitted)}
\end{tabularx}
\vfill
}
\end{minipage}
\end{minipage}

\end{tabularx}
\caption{Illustration of the \textbf{\benchmarkname{}} benchmark on the \texttt{Apollo} \texttt{Program} template showcasing the factual excerpt from Wikipedia on the top left with its annotated entities, a paired \fictitiousterm{} document on the top right, the factual and \fictitiousterm{} question-answer pairs below, the replacement table that maps each factual entity to its \fictitiousterm{} value and some of the associated rules that should be satisfied when generating the \fictitiousterm{} variants on the bottom. The same highlighting color is used to represent two paired entities between factual and \fictitiousterm{}.}
\label{fig:main_memoreason_example}

\endgroup
\end{figure*}

\textbf{Question Classes.} Four classes of questions are proposed: simple extractive questions, arithmetic questions involving calculations, inference questions that are not numerical, and temporal reasoning. We propose these question classes to test \textit{memorization biases} on different reasoning tasks~(see Figure \ref{fig:main_memoreason_example} for examples of each):
\begin{itemize}[leftmargin=*,itemsep=3pt]
    \item \textbf{Extractive (baseline):} the answer is explicitly stated in the document, so the main challenge is to identify and extract the relevant evidence.
    \item \textbf{Arithmetic:} answering this type of question requires performing arithmetic operations on the numerical entities mentioned in the document.
    \item \textbf{Temporal:} this type of question aims to test the ability of models to reason over temporal entities such as years, dates, ages, or durations.
    \item \textbf{Inference:} answering this type of question requires reasoning that is not primarily arithmetic or temporal.
\end{itemize}

\subsection{Template Annotation}
\label{subsec:template_annotation_process}

The template annotation process consists of transforming the factual dataset $\mathcal{D}$ into a dataset of templates $\mathcal{D}^{\mathrm{template}}=\{(d_i^{\mathrm{template}},q_{i}^{\mathrm{template}},a_{i}^{\mathrm{template}},\mathcal{E}_i^{\mathrm{factual}},\mathcal{R}_i)\}_{i=1}^{N}$ where $d_i^{\mathrm{template}}$ is a document with annotated entities following a taxonomy that we define in Appendix \ref{appendix:taxonomy}, $q_i^{\mathrm{template}}$ the question, $a_i^{\mathrm{template}}$ the answer, $\mathcal{E}_i^{\mathrm{factual}}$ the set of factual entities, and $\mathcal{R}_i$ the set of rules that must be satisfied when replacing factual entities with \fictitiousterm{} ones in $d_i^{\mathrm{template}}$. Unlike \texttt{GSM-Symbolic} \citep{mirzadeh2025gsmsymbolic}, which operates on short arithmetic statements, our setting involves entity-rich documents with substantially greater complexity, requiring us to design more sophisticated and expressive templates.

\textbf{Entity Annotation.} A factual document $d_i$ is transformed into a template document $d_i^{\mathrm{template}}$ by identifying entities and assigning the right entity type and attribute (e.g., \textit{Albert Einstein} gets assigned the \texttt{person} entity type with a \texttt{full\_name} attribute). This annotation process ensures that factual entities are replaced with plausible \fictitiousterm{} ones that preserve their type (e.g., person $\rightarrow$ person, number $\rightarrow$ number). To this end, we developed a taxonomy of 14 entity types and their associated attributes, designed to cover all entities present in the factual dataset (see Appendix \ref{appendix:taxonomy} for details).

\newcommand{\DatasetLengthAwardWinners}{\shortstack{2{,}974\\{\scriptsize $\pm1{,}036$}}}
\newcommand{\DatasetLengthBiographies}{\shortstack{3{,}160\\{\scriptsize $\pm952$}}}
\newcommand{\DatasetLengthPlaces}{\shortstack{2{,}606\\{\scriptsize $\pm600$}}}
\newcommand{\DatasetLengthCompanies}{\shortstack{1{,}943\\{\scriptsize $\pm799$}}}
\newcommand{\DatasetLengthNaturalDisasters}{\shortstack{2{,}644\\{\scriptsize $\pm969$}}}
\newcommand{\DatasetLengthPublicAttacks}{\shortstack{2{,}454\\{\scriptsize $\pm1{,}337$}}}
\newcommand{\DatasetLengthRetailBanking}{\shortstack{1{,}265\\{\scriptsize $\pm498$}}}
\newcommand{\DatasetLengthSpaceMissions}{\shortstack{1{,}503\\{\scriptsize $\pm698$}}}
\newcommand{\DatasetLengthSportEvents}{\shortstack{2{,}268\\{\scriptsize $\pm750$}}}
\newcommand{\DatasetLengthTotal}{\shortstack{2{,}298\\{\scriptsize $\pm1{,}037$}}}

\begin{table*}[t]
    \centering
    \footnotesize
    \renewcommand{\arraystretch}{1.18}
    \setlength{\tabcolsep}{3pt}
        \caption{\textbf{Dataset statistics by theme category.} \revision{Task-template counts are reported by theme; entity and rule counts are per document ($\pm$ sample standard deviation).}}
    \label{tab:theme-dataset-stats}

    \resizebox{\textwidth}{!}{%
    \begin{tabular}{@{}lcccccccccc@{}}
        \toprule
         & \rotatebox{60}{\textbf{Award Winners}} 
         & \rotatebox{60}{\textbf{Biographies}} 
         & \rotatebox{60}{\textbf{Places}} 
         & \rotatebox{60}{\textbf{Companies}} 
         & \rotatebox{60}{\textbf{Natural Disasters}} 
         & \rotatebox{60}{\textbf{Public Attacks}} 
         & \rotatebox{60}{\textbf{Retail Banking}} 
         & \rotatebox{60}{\textbf{Space Missions}} 
         & \rotatebox{60}{\textbf{Sport Events}} 
         & \rotatebox{60}{\textbf{Total}} \\
        \midrule
        \revision{\textbf{\#Task templates}}
        & \unchangednumber{\shortstack{144\\{\scriptsize (12.0\%)}}}
        & \unchangednumber{\shortstack{120\\{\scriptsize (10.0\%)}}}
        & \unchangednumber{\shortstack{120\\{\scriptsize (10.0\%)}}}
        & \unchangednumber{\shortstack{144\\{\scriptsize (12.0\%)}}}
        & \unchangednumber{\shortstack{144\\{\scriptsize (12.0\%)}}}
        & \unchangednumber{\shortstack{120\\{\scriptsize (10.0\%)}}}
        & \unchangednumber{\shortstack{132\\{\scriptsize (11.0\%)}}}
        & \unchangednumber{\shortstack{144\\{\scriptsize (12.0\%)}}}
        & \unchangednumber{\shortstack{132\\{\scriptsize (11.0\%)}}}
        & \unchangednumber{\textbf{1{,}200}} \\
        \midrule
        \revision{\textbf{Avg length (chars)}}
        & \unchangednumber{\DatasetLengthAwardWinners}
        & \unchangednumber{\DatasetLengthBiographies}
        & \unchangednumber{\DatasetLengthPlaces}
        & \unchangednumber{\DatasetLengthCompanies}
        & \unchangednumber{\DatasetLengthNaturalDisasters}
        & \unchangednumber{\DatasetLengthPublicAttacks}
        & \unchangednumber{\DatasetLengthRetailBanking}
        & \unchangednumber{\DatasetLengthSpaceMissions}
        & \unchangednumber{\DatasetLengthSportEvents}
        & \unchangednumber{\DatasetLengthTotal} \\
        \midrule
        \textbf{Avg \#Entities} 
        & \unchangednumber{\shortstack{58.0\\{\scriptsize $\pm26.1$}}}
        & \unchangednumber{\shortstack{51.4\\{\scriptsize $\pm14.6$}}}
        & \unchangednumber{\shortstack{48.7\\{\scriptsize $\pm14.4$}}}
        & \unchangednumber{\shortstack{41.4\\{\scriptsize $\pm14.9$}}}
        & \unchangednumber{\shortstack{39.8\\{\scriptsize $\pm14.2$}}}
        & \unchangednumber{\shortstack{22.3\\{\scriptsize $\pm9.5$}}}
        & \unchangednumber{\shortstack{14.3\\{\scriptsize $\pm5.5$}}}
        & \unchangednumber{\shortstack{26.2\\{\scriptsize $\pm9.8$}}}
        & \unchangednumber{\shortstack{39.4\\{\scriptsize $\pm12.6$}}}
        & \unchangednumber{\shortstack{38.0\\{\scriptsize $\pm19.6$}}} \\
        \midrule
        \textbf{Avg \#Rules} 
        & \unchangednumber{\shortstack{57.2\\{\scriptsize $\pm34.7$}}}
        & \unchangednumber{\shortstack{35.5\\{\scriptsize $\pm9.6$}}}
        & \unchangednumber{\shortstack{31.4\\{\scriptsize $\pm8.8$}}}
        & \unchangednumber{\shortstack{21.7\\{\scriptsize $\pm13.0$}}}
        & \unchangednumber{\shortstack{38.5\\{\scriptsize $\pm16.4$}}}
        & \unchangednumber{\shortstack{17.4\\{\scriptsize $\pm9.6$}}}
        & \unchangednumber{\shortstack{8.4\\{\scriptsize $\pm5.6$}}}
        & \unchangednumber{\shortstack{23.5\\{\scriptsize $\pm12.8$}}}
        & \unchangednumber{\shortstack{44.9\\{\scriptsize $\pm15.7$}}}
        & \unchangednumber{\shortstack{31.2\\{\scriptsize $\pm21.4$}}} \\
        \bottomrule
    \end{tabular}%
    }
\end{table*}

\textbf{Rules.} It is necessary to make sure that replacing the factual entities does not alter the logic and core semantics of the document. We therefore manually encode constraints or rules that must be satisfied when replacing with \fictitiousterm{} entities. Those rules include:
\begin{itemize}
    \item Plural mentions: "[...] \textit{He was among the} [\textbf{\textcolor{blue}{10}}; \texttt{number\_1.int}] \textit{selected player}\underline{\textcolor{blue}{s}} [...]" $\rightarrow \texttt{number\_1.int} \geq 2$
    \item Numbers that should sum to a total: "[...] \textit{Among the} [\textbf{\textcolor{blue}{fourteen}}; \texttt{number\_2.str}] \textit{casualties are} [\textbf{\textcolor{blue}{9}}; \texttt{number\_3.int}] \textit{injured and} [\textbf{\textcolor{blue}{5}}; \texttt{number\_4.int}] \textit{dead} [...]"\\ $\rightarrow \texttt{number\_2.str} = \texttt{number\_3.int} + \texttt{number\_4.int}$
    \item Order preservation: we define rules to ensure that the order of numerical and temporal entities (e.g. years, dates, etc.) is maintained between factual and \fictitiousterm{} settings. This is crucial especially for temporal entities to avoid breaking the logical timeline of events.
    \item Sampling intervals: for number and temporal entities, we sample the \fictitiousterm{} replacements from restricted intervals around the factual values to avoid situations where replacing a small number with a very large one would create unnatural situations such as asserting a historical figure lived to be 800 years old instead of 80 or that a stock price went negative.
\end{itemize}

\textbf{Annotation Process.} All the annotation steps, including the entity annotation, rule definition, and question-answer writing are first drafted by an AI Agent backed by Claude Opus 4.6 \citep{anthropic2026claude46opus}. Then 15 human annotators (qualified colleagues with knowledge in LLMs) independently check the generated annotations and modify them if needed. Finally, the annotators meet in a dedicated agreement session. Even with a highly capable AI Agent, systematic human annotations proved indispensable for correcting, refining and validating all initial AI drafts. As a point of reference, almost 50\% of the Q-A pairs required human intervention \revision{and we estimate the total time spent in the annotation process around 200 hours (i.e., 2 hours per document)}. We stress that the annotation and review process is both intensive and of critical importance. We show in Figure \ref{fig:annotation_interface} of Appendix \ref{sec:annotation_interface} a screenshot of the annotation interface used to inspect entity spans, replacement rules, and question-answer fields for the \texttt{Toyota} template.
\postreviewrevision{Although Claude Opus \unchangednumber{4.6} drafts fictitious named entities and Claude Sonnet \unchangednumber{4.6} is among the evaluated models, candidate acceptance is model-independent: entities undergo web and Wikipedia non-existence checks, human review, and deterministic rule validation before evaluation.}

\subsection{\texorpdfstring{\Fictitiousterm{}}{Fictitious} Dataset}
\label{subsec:fictional_dataset_generation}

For each template, we generate $K=10$ \fictitiousterm{} variants $\{(d_{i,s}^{\fictitiousmath},q_{i,s}^{\fictitiousmath},a_{i,s}^{\fictitiousmath})\}_{s=1}^{K}$ by replacing all entities under rule constraints.
We categorize the entities into \textbf{named} entities (\textit{e.g.} persons, places, etc.) and \textbf{numerical} entities (\textit{e.g.} numbers, temporal entities such as dates, etc.).
To ensure that the generated entity values are \fictitiousterm{} and satisfy the replacement rules, the generation process includes 2 steps for each document:

\begin{itemize}
    \item \textbf{Named entity sampling.}  We use an AI Agent backed by \texttt{Claude Opus 4.6} \citep{anthropic2026claude46opus} to build a document-specific pool of \fictitiousterm{} named-entity candidates. The generated candidates must be non-existing, pronounceable \citep{BAUER_2015}, type-compatible, and internally coherent across attributes. When a document links several attributes of a single entity, such as a place name and its demonym, the candidate pool keeps the linked attributes aligned so that a sampled \fictitiousterm{} document remains coherent. The prompt used is provided in Appendix~\ref{appendix:fictional_pool_prompt}. We also ensure that the generated entities do not exist by checking their existence in Wikipedia and via a web search.
    \item \textbf{Numerical entity sampling.} To sample numerical and temporal entities that satisfy the rule constraints, we use the Mixed Integer Linear Programming solver implemented in \texttt{SciPy} \citep{Virtanen_2020}. We also make sure not to sample the same values across the $K$ \fictitiousterm{} variants. Days and months are also replaced and verified manually against the rules.
    
\end{itemize}

\textbf{Answer Classes.} Orthogonally to the question class, and in order to systematically isolate the influence of parametric memory, we define three answer classes. Questions with \textit{variant} answers depend on the entities described in the document and, therefore, their answers change when those entities are replaced. Questions with \textit{invariant} answers are independent of the \fictitiousterm{} substitutions and remain identical between factual and \fictitiousterm{} settings. \revision{Invariant questions are still evaluated on the fully fictitious document---both named identities and numerical/temporal values are replaced---but their answer literal is unchanged.} \textit{Refusal} questions ask for information that is not provided in the document, testing whether the model stays grounded in the provided context.

\revision{Each document-template is used to generate $K=\unchangednumber{10}$ fictitious variants for each of its $\unchangednumber{12}$ Q-A pairs, yielding \unchangednumber{12{,}000} fictitious pairs.} Figure \ref{fig:main_memoreason_example} illustrates a template on the \texttt{Apollo} \texttt{Program} example.

\section{Evaluation Framework}
\label{sec:eval_framework}

\paragraph{Measuring Performance Variation.} Given the performance of a model on a factual example $(d_i, q_{i}, a_{i})$, noted $Y^{\mathrm{factual}}_{i}$, we are interested in how this performance changes towards the set of $K$ corresponding \fictitiousterm{} examples $\{(d_{i,s}^{\fictitiousmath}, q_{i,s}^{\fictitiousmath}, a_{i,s}^{\fictitiousmath})\}_{s=1}^{K}$. We define the paired difference as $\Delta_{i} = \hat{Y}_{i}^{\fictitiousmath} - Y_{i}^{\mathrm{factual}}$, where $\hat{Y}_{i}^{\fictitiousmath}$ denotes the average performance across the $K$ corresponding \fictitiousterm{} examples. \revision{To account for dependence among questions and variants from the same document, confidence intervals are computed by bootstrapping complete documents.}

\paragraph{Exact and Judge Match.} We assess prediction correctness using two complementary metrics. We first apply Exact Match (EM) and for predictions deemed incorrect we follow up with Judge Match (JM), in which an LLM is prompted to determine whether the prediction and the ground truth are semantically equivalent. To validate the reliability of the JM, we calibrated it on 200 randomly sampled examples, achieving 100\% human-machine agreement. This confirms that the JM not only captures correct predictions missed by EM, but also reliably identifies incorrect ones, suggesting it is not prone to the confirmation bias toward affirmative judgments that has been reported in LLMs \citep{jain2025consensusmitigatingagreeablenessbias, he2026martingale}. We provide more details on the calibration process in Appendix \ref{sec:judge_calibration}.

\paragraph{Model Selection.} We evaluate nine recent, high-performing reasoning models spanning multiple providers: \texttt{GPT-OSS-20B} and \texttt{GPT-OSS-120B} \citep{openai2025gptoss120bgptoss20bmodel}, \texttt{OLMo-3-7B-Think} and \texttt{OLMo-3-7B-Instruct} \citep{olmo2026olmo3}, \texttt{Gemma-4-26B-A4B-it}, \texttt{Qwen3.5-27B} and \texttt{Qwen3.5-35B-A3B} \citep{qwen35blog}, \texttt{Llama-3.1-8B-Instruct} \citep{dubey2024llama3}, and \texttt{Claude Sonnet 4.6} \citep{anthropic2026claude46sonnet, anthropic2026claude46opus}. This set combines open-weight models with a frontier API model, allowing us to compare memorization biases and parametric-shortcuts across model families and scales.

\section{Experiments \& Results}

In this section, we describe the experiments conducted on the \textbf{\benchmarkname} benchmark. In Section~\ref{subsec:parametric_memory_impact} we examine the impact of replacing entities in the document on LLMs' ability to answer questions based on contextual information, testing the existence of a \textit{memorization bias} in reasoning. In Section~\ref{subsec:varying_proportion}, we study how performance evolves as the proportion of replaced entities increases, testing whether models perform worse in the knowledge-conflict setting of partial entity replacement than when all entities are replaced (Section~4.2). \postreviewrevision{Section~\ref{sec:cot_effect} tests chain-of-thought effects and reasoning effort.} Finally, in Section~\ref{sec:parametric_shortcut}, we put the \textit{Strong Parametric Shortcut Hypothesis} to the test to ascertain whether prediction errors are due to pure recall of learned facts. \revision{Additional robustness and mechanism checks are reported in Appendix~\ref{appendix:robustness_checks}.}

\definecolor{dropPurpleCell}{HTML}{E7D0FF}
\definecolor{riseBlueCell}{HTML}{D6E4FF}
\begin{table}[!htbp]
\revisiontablecolor
\centering
\scriptsize
\renewcommand{\arraystretch}{0.98}
\setlength{\tabcolsep}{2.0pt}
\caption{Mean performance change (fictitious minus factual, \%) by answer and question type. We report aggregated reasoning question performance (i.e. \textit{Arith} for Arithmetic, \textit{Temp} for Temporal, and \textit{Infer} for Inference) in the \textit{Reason} column. Extractive questions are reported in the \textit{Extr} column. Statistically significant results are framed and indicated in bold. We also report the 95\% confidence intervals below each value. Results that decrease and increase from factual to fictitious are highlighted in
\protect\colorbox{dropPurpleCell}{\textcolor{black}{purple}} and
\protect\colorbox{riseBlueCell}{\textcolor{black}{blue}}, respectively.}
\label{tab:question-answer-type-drop}
\resizebox{\textwidth}{!}{%
\begin{tabular}{@{}l*{5}{c}@{\hspace{8pt}}*{5}{c}@{\hspace{8pt}}*{5}{c}@{}}
\toprule
\textit{Answer} & \multicolumn{5}{c}{\textbf{Variant}} & \multicolumn{5}{c}{\textbf{Invariant}} & \multicolumn{5}{c}{\textbf{Refusal}} \\
\cmidrule(lr){2-6} \cmidrule(lr){7-11} \cmidrule(lr){12-16}
\textit{Question} & Arith. & Temp. & Infer. & \tikzmarknode[inner sep=1.4pt]{dataset100_variant_reason_top}{\textbf{Reason.}} & Extr. & Arith. & Temp. & Infer. & \tikzmarknode[inner sep=1.4pt]{dataset100_invariant_reason_top}{\textbf{Reason.}} & Extr. & Arith. & Temp. & Infer. & \tikzmarknode[inner sep=1.4pt]{dataset100_refusal_reason_top}{\textbf{Reason.}} & Extr. \\
\midrule
\shortstack[l]{\textbf{OLMO-3}\\\textbf{7B-THINK}} & \cellcolor{dropPurpleCell}\shortstack{-1.4\\{\fontsize{3.8}{4.1}\selectfont [-9.2, +6.5]}} & \cellcolor{dropPurpleCell}\tikz[baseline=(sig.base)]\node[draw=black!80,fill=white!35!dropPurpleCell,line width=0.28pt,rounded corners=1.1pt,inner sep=1.0pt,outer sep=0pt] (sig) {\shortstack{\textbf{-15.6}\\{\fontsize{3.8}{4.1}\selectfont [-23.2, -7.8]}}}; & \cellcolor{dropPurpleCell}\shortstack{-7.4\\{\fontsize{3.8}{4.1}\selectfont [-14.6, +0.0]}} & \cellcolor{dropPurpleCell}\tikz[baseline=(sig.base)]\node[draw=black!80,fill=white!35!dropPurpleCell,line width=0.28pt,rounded corners=1.1pt,inner sep=1.0pt,outer sep=0pt] (sig) {\shortstack{\textbf{-8.1}\\{\fontsize{3.8}{4.1}\selectfont [-12.4, -3.8]}}}; & \cellcolor{dropPurpleCell}\tikz[baseline=(sig.base)]\node[draw=black!80,fill=white!35!dropPurpleCell,line width=0.28pt,rounded corners=1.1pt,inner sep=1.0pt,outer sep=0pt] (sig) {\shortstack{\textbf{-9.3}\\{\fontsize{3.8}{4.1}\selectfont [-13.6, -4.8]}}}; & \cellcolor{dropPurpleCell}\shortstack{-0.8\\{\fontsize{3.8}{4.1}\selectfont [-8.3, +6.7]}} & \cellcolor{dropPurpleCell}\shortstack{-6.6\\{\fontsize{3.8}{4.1}\selectfont [-13.2, +0.4]}} & \cellcolor{dropPurpleCell}\shortstack{-2.2\\{\fontsize{3.8}{4.1}\selectfont [-10.2, +6.1]}} & \cellcolor{dropPurpleCell}\shortstack{-3.2\\{\fontsize{3.8}{4.1}\selectfont [-7.7, +1.4]}} & \cellcolor{dropPurpleCell}\shortstack{-1.6\\{\fontsize{3.8}{4.1}\selectfont [-7.7, +4.6]}} & \cellcolor{dropPurpleCell}\shortstack{-3.6\\{\fontsize{3.8}{4.1}\selectfont [-9.1, +2.0]}} & \cellcolor{dropPurpleCell}\tikz[baseline=(sig.base)]\node[draw=black!80,fill=white!35!dropPurpleCell,line width=0.28pt,rounded corners=1.1pt,inner sep=1.0pt,outer sep=0pt] (sig) {\shortstack{\textbf{-10.8}\\{\fontsize{3.8}{4.1}\selectfont [-17.9, -3.6]}}}; & \cellcolor{dropPurpleCell}\shortstack{-2.0\\{\fontsize{3.8}{4.1}\selectfont [-7.9, +4.0]}} & \cellcolor{dropPurpleCell}\tikz[baseline=(sig.base)]\node[draw=black!80,fill=white!35!dropPurpleCell,line width=0.28pt,rounded corners=1.1pt,inner sep=1.0pt,outer sep=0pt] (sig) {\shortstack{\textbf{-5.5}\\{\fontsize{3.8}{4.1}\selectfont [-9.1, -1.9]}}}; & \cellcolor{dropPurpleCell}\shortstack{-2.4\\{\fontsize{3.8}{4.1}\selectfont [-7.1, +2.5]}} \\
\arrayrulecolor{black!18}\specialrule{0.25pt}{1.0pt}{1.0pt}\arrayrulecolor{black}
\shortstack[l]{\textbf{OLMO-3}\\\textbf{7B-INSTRUCT}} & \cellcolor{dropPurpleCell}\shortstack{-4.0\\{\fontsize{3.8}{4.1}\selectfont [-12.9, +4.6]}} & \cellcolor{dropPurpleCell}\shortstack{-3.7\\{\fontsize{3.8}{4.1}\selectfont [-11.5, +4.1]}} & \cellcolor{dropPurpleCell}\tikz[baseline=(sig.base)]\node[draw=black!80,fill=white!35!dropPurpleCell,line width=0.28pt,rounded corners=1.1pt,inner sep=1.0pt,outer sep=0pt] (sig) {\shortstack{\textbf{-15.7}\\{\fontsize{3.8}{4.1}\selectfont [-24.8, -6.5]}}}; & \cellcolor{dropPurpleCell}\tikz[baseline=(sig.base)]\node[draw=black!80,fill=white!35!dropPurpleCell,line width=0.28pt,rounded corners=1.1pt,inner sep=1.0pt,outer sep=0pt] (sig) {\shortstack{\textbf{-7.8}\\{\fontsize{3.8}{4.1}\selectfont [-12.6, -2.9]}}}; & \cellcolor{dropPurpleCell}\shortstack{-6.0\\{\fontsize{3.8}{4.1}\selectfont [-12.0, +0.1]}} & \cellcolor{dropPurpleCell}\shortstack{-1.6\\{\fontsize{3.8}{4.1}\selectfont [-8.0, +4.7]}} & \cellcolor{dropPurpleCell}\tikz[baseline=(sig.base)]\node[draw=black!80,fill=white!35!dropPurpleCell,line width=0.28pt,rounded corners=1.1pt,inner sep=1.0pt,outer sep=0pt] (sig) {\shortstack{\textbf{-11.1}\\{\fontsize{3.8}{4.1}\selectfont [-18.5, -4.1]}}}; & \cellcolor{dropPurpleCell}\shortstack{-5.4\\{\fontsize{3.8}{4.1}\selectfont [-12.6, +1.7]}} & \cellcolor{dropPurpleCell}\tikz[baseline=(sig.base)]\node[draw=black!80,fill=white!35!dropPurpleCell,line width=0.28pt,rounded corners=1.1pt,inner sep=1.0pt,outer sep=0pt] (sig) {\shortstack{\textbf{-6.0}\\{\fontsize{3.8}{4.1}\selectfont [-10.1, -2.1]}}}; & \cellcolor{dropPurpleCell}\shortstack{-0.3\\{\fontsize{3.8}{4.1}\selectfont [-5.7, +5.2]}} & \cellcolor{riseBlueCell}\shortstack{+2.1\\{\fontsize{3.8}{4.1}\selectfont [-2.8, +7.4]}} & \cellcolor{riseBlueCell}\tikz[baseline=(sig.base)]\node[draw=black!80,fill=white!35!riseBlueCell,line width=0.28pt,rounded corners=1.1pt,inner sep=1.0pt,outer sep=0pt] (sig) {\shortstack{\textbf{+6.5}\\{\fontsize{3.8}{4.1}\selectfont [+0.0, +13.2]}}}; & \cellcolor{riseBlueCell}\shortstack{+3.9\\{\fontsize{3.8}{4.1}\selectfont [-1.7, +9.9]}} & \cellcolor{riseBlueCell}\tikz[baseline=(sig.base)]\node[draw=black!80,fill=white!35!riseBlueCell,line width=0.28pt,rounded corners=1.1pt,inner sep=1.0pt,outer sep=0pt] (sig) {\shortstack{\textbf{+4.2}\\{\fontsize{3.8}{4.1}\selectfont [+0.8, +7.6]}}}; & \cellcolor{riseBlueCell}\shortstack{+1.4\\{\fontsize{3.8}{4.1}\selectfont [-2.1, +5.3]}} \\
\arrayrulecolor{black!18}\specialrule{0.25pt}{1.0pt}{1.0pt}\arrayrulecolor{black}
\shortstack[l]{\textbf{GPT-OSS}\\\textbf{20B}} & \cellcolor{dropPurpleCell}\tikz[baseline=(sig.base)]\node[draw=black!80,fill=white!35!dropPurpleCell,line width=0.28pt,rounded corners=1.1pt,inner sep=1.0pt,outer sep=0pt] (sig) {\shortstack{\textbf{-4.8}\\{\fontsize{3.8}{4.1}\selectfont [-9.3, -0.5]}}}; & \cellcolor{dropPurpleCell}\tikz[baseline=(sig.base)]\node[draw=black!80,fill=white!35!dropPurpleCell,line width=0.28pt,rounded corners=1.1pt,inner sep=1.0pt,outer sep=0pt] (sig) {\shortstack{\textbf{-4.8}\\{\fontsize{3.8}{4.1}\selectfont [-9.3, -0.6]}}}; & \cellcolor{dropPurpleCell}\tikz[baseline=(sig.base)]\node[draw=black!80,fill=white!35!dropPurpleCell,line width=0.28pt,rounded corners=1.1pt,inner sep=1.0pt,outer sep=0pt] (sig) {\shortstack{\textbf{-11.3}\\{\fontsize{3.8}{4.1}\selectfont [-18.2, -4.5]}}}; & \cellcolor{dropPurpleCell}\tikz[baseline=(sig.base)]\node[draw=black!80,fill=white!35!dropPurpleCell,line width=0.28pt,rounded corners=1.1pt,inner sep=1.0pt,outer sep=0pt] (sig) {\shortstack{\textbf{-7.0}\\{\fontsize{3.8}{4.1}\selectfont [-10.2, -3.8]}}}; & \cellcolor{dropPurpleCell}\tikz[baseline=(sig.base)]\node[draw=black!80,fill=white!35!dropPurpleCell,line width=0.28pt,rounded corners=1.1pt,inner sep=1.0pt,outer sep=0pt] (sig) {\shortstack{\textbf{-7.0}\\{\fontsize{3.8}{4.1}\selectfont [-12.0, -2.6]}}}; & \cellcolor{dropPurpleCell}\shortstack{-0.3\\{\fontsize{3.8}{4.1}\selectfont [-6.6, +6.2]}} & \cellcolor{dropPurpleCell}\tikz[baseline=(sig.base)]\node[draw=black!80,fill=white!35!dropPurpleCell,line width=0.28pt,rounded corners=1.1pt,inner sep=1.0pt,outer sep=0pt] (sig) {\shortstack{\textbf{-9.7}\\{\fontsize{3.8}{4.1}\selectfont [-15.6, -4.1]}}}; & \cellcolor{riseBlueCell}\shortstack{+0.2\\{\fontsize{3.8}{4.1}\selectfont [-4.2, +4.8]}} & \cellcolor{dropPurpleCell}\tikz[baseline=(sig.base)]\node[draw=black!80,fill=white!35!dropPurpleCell,line width=0.28pt,rounded corners=1.1pt,inner sep=1.0pt,outer sep=0pt] (sig) {\shortstack{\textbf{-3.3}\\{\fontsize{3.8}{4.1}\selectfont [-6.5, -0.1]}}}; & \cellcolor{dropPurpleCell}\shortstack{-2.2\\{\fontsize{3.8}{4.1}\selectfont [-6.4, +1.8]}} & \cellcolor{dropPurpleCell}\shortstack{-3.6\\{\fontsize{3.8}{4.1}\selectfont [-8.6, +1.1]}} & \cellcolor{riseBlueCell}\shortstack{+1.8\\{\fontsize{3.8}{4.1}\selectfont [-3.7, +7.4]}} & \cellcolor{riseBlueCell}\shortstack{+4.1\\{\fontsize{3.8}{4.1}\selectfont [-2.6, +11.0]}} & \cellcolor{riseBlueCell}\shortstack{+0.8\\{\fontsize{3.8}{4.1}\selectfont [-2.7, +4.3]}} & \cellcolor{riseBlueCell}\shortstack{+0.4\\{\fontsize{3.8}{4.1}\selectfont [-3.5, +4.5]}} \\
\arrayrulecolor{black!18}\specialrule{0.25pt}{1.0pt}{1.0pt}\arrayrulecolor{black}
\shortstack[l]{\textbf{GEMMA-4}\\\textbf{26B-A4B-IT}} & \cellcolor{dropPurpleCell}\tikz[baseline=(sig.base)]\node[draw=black!80,fill=white!35!dropPurpleCell,line width=0.28pt,rounded corners=1.1pt,inner sep=1.0pt,outer sep=0pt] (sig) {\shortstack{\textbf{-6.9}\\{\fontsize{3.8}{4.1}\selectfont [-13.2, -0.7]}}}; & \cellcolor{dropPurpleCell}\shortstack{-4.6\\{\fontsize{3.8}{4.1}\selectfont [-10.3, +1.0]}} & \cellcolor{dropPurpleCell}\tikz[baseline=(sig.base)]\node[draw=black!80,fill=white!35!dropPurpleCell,line width=0.28pt,rounded corners=1.1pt,inner sep=1.0pt,outer sep=0pt] (sig) {\shortstack{\textbf{-8.1}\\{\fontsize{3.8}{4.1}\selectfont [-15.1, -1.4]}}}; & \cellcolor{dropPurpleCell}\tikz[baseline=(sig.base)]\node[draw=black!80,fill=white!35!dropPurpleCell,line width=0.28pt,rounded corners=1.1pt,inner sep=1.0pt,outer sep=0pt] (sig) {\shortstack{\textbf{-6.5}\\{\fontsize{3.8}{4.1}\selectfont [-10.2, -2.9]}}}; & \cellcolor{dropPurpleCell}\tikz[baseline=(sig.base)]\node[draw=black!80,fill=white!35!dropPurpleCell,line width=0.28pt,rounded corners=1.1pt,inner sep=1.0pt,outer sep=0pt] (sig) {\shortstack{\textbf{-4.2}\\{\fontsize{3.8}{4.1}\selectfont [-7.7, -1.4]}}}; & \cellcolor{riseBlueCell}\shortstack{+0.1\\{\fontsize{3.8}{4.1}\selectfont [-5.8, +6.0]}} & \cellcolor{dropPurpleCell}\shortstack{-2.0\\{\fontsize{3.8}{4.1}\selectfont [-7.2, +3.3]}} & \cellcolor{riseBlueCell}\shortstack{+0.5\\{\fontsize{3.8}{4.1}\selectfont [-5.0, +6.1]}} & \cellcolor{dropPurpleCell}\shortstack{-0.5\\{\fontsize{3.8}{4.1}\selectfont [-3.7, +2.7]}} & \cellcolor{dropPurpleCell}\shortstack{-0.0\\{\fontsize{3.8}{4.1}\selectfont [-3.3, +3.3]}} & \cellcolor{riseBlueCell}\tikz[baseline=(sig.base)]\node[draw=black!80,fill=white!35!riseBlueCell,line width=0.28pt,rounded corners=1.1pt,inner sep=1.0pt,outer sep=0pt] (sig) {\shortstack{\textbf{+3.8}\\{\fontsize{3.8}{4.1}\selectfont [+0.7, +7.8]}}}; & \cellcolor{riseBlueCell}\shortstack{+3.4\\{\fontsize{3.8}{4.1}\selectfont [-1.0, +8.2]}} & \cellcolor{dropPurpleCell}\shortstack{-0.2\\{\fontsize{3.8}{4.1}\selectfont [-4.5, +4.2]}} & \cellcolor{riseBlueCell}\tikz[baseline=(sig.base)]\node[draw=black!80,fill=white!35!riseBlueCell,line width=0.28pt,rounded corners=1.1pt,inner sep=1.0pt,outer sep=0pt] (sig) {\shortstack{\textbf{+2.3}\\{\fontsize{3.8}{4.1}\selectfont [+0.1, +4.7]}}}; & \cellcolor{riseBlueCell}\shortstack{+2.1\\{\fontsize{3.8}{4.1}\selectfont [-0.6, +5.6]}} \\
\arrayrulecolor{black!18}\specialrule{0.25pt}{1.0pt}{1.0pt}\arrayrulecolor{black}
\shortstack[l]{\textbf{GPT-OSS}\\\textbf{120B}} & \cellcolor{dropPurpleCell}\tikz[baseline=(sig.base)]\node[draw=black!80,fill=white!35!dropPurpleCell,line width=0.28pt,rounded corners=1.1pt,inner sep=1.0pt,outer sep=0pt] (sig) {\shortstack{\textbf{-4.7}\\{\fontsize{3.8}{4.1}\selectfont [-9.4, -0.2]}}}; & \cellcolor{dropPurpleCell}\tikz[baseline=(sig.base)]\node[draw=black!80,fill=white!35!dropPurpleCell,line width=0.28pt,rounded corners=1.1pt,inner sep=1.0pt,outer sep=0pt] (sig) {\shortstack{\textbf{-7.4}\\{\fontsize{3.8}{4.1}\selectfont [-11.3, -4.0]}}}; & \cellcolor{dropPurpleCell}\tikz[baseline=(sig.base)]\node[draw=black!80,fill=white!35!dropPurpleCell,line width=0.28pt,rounded corners=1.1pt,inner sep=1.0pt,outer sep=0pt] (sig) {\shortstack{\textbf{-11.8}\\{\fontsize{3.8}{4.1}\selectfont [-18.8, -4.9]}}}; & \cellcolor{dropPurpleCell}\tikz[baseline=(sig.base)]\node[draw=black!80,fill=white!35!dropPurpleCell,line width=0.28pt,rounded corners=1.1pt,inner sep=1.0pt,outer sep=0pt] (sig) {\shortstack{\textbf{-8.0}\\{\fontsize{3.8}{4.1}\selectfont [-11.3, -4.7]}}}; & \cellcolor{dropPurpleCell}\tikz[baseline=(sig.base)]\node[draw=black!80,fill=white!35!dropPurpleCell,line width=0.28pt,rounded corners=1.1pt,inner sep=1.0pt,outer sep=0pt] (sig) {\shortstack{\textbf{-7.8}\\{\fontsize{3.8}{4.1}\selectfont [-12.5, -3.7]}}}; & \cellcolor{dropPurpleCell}\shortstack{-0.8\\{\fontsize{3.8}{4.1}\selectfont [-6.6, +5.2]}} & \cellcolor{riseBlueCell}\shortstack{+2.1\\{\fontsize{3.8}{4.1}\selectfont [-3.2, +7.6]}} & \cellcolor{dropPurpleCell}\shortstack{-0.2\\{\fontsize{3.8}{4.1}\selectfont [-3.3, +3.3]}} & \cellcolor{riseBlueCell}\shortstack{+0.4\\{\fontsize{3.8}{4.1}\selectfont [-2.6, +3.3]}} & \cellcolor{dropPurpleCell}\shortstack{-0.6\\{\fontsize{3.8}{4.1}\selectfont [-4.4, +3.2]}} & \cellcolor{dropPurpleCell}\tikz[baseline=(sig.base)]\node[draw=black!80,fill=white!35!dropPurpleCell,line width=0.28pt,rounded corners=1.1pt,inner sep=1.0pt,outer sep=0pt] (sig) {\shortstack{\textbf{-5.6}\\{\fontsize{3.8}{4.1}\selectfont [-10.7, -0.8]}}}; & \cellcolor{dropPurpleCell}\shortstack{-1.0\\{\fontsize{3.8}{4.1}\selectfont [-5.0, +2.9]}} & \cellcolor{dropPurpleCell}\shortstack{-1.4\\{\fontsize{3.8}{4.1}\selectfont [-6.5, +3.7]}} & \cellcolor{dropPurpleCell}\shortstack{-2.7\\{\fontsize{3.8}{4.1}\selectfont [-5.4, +0.1]}} & \cellcolor{dropPurpleCell}\shortstack{-1.4\\{\fontsize{3.8}{4.1}\selectfont [-5.5, +2.7]}} \\
\arrayrulecolor{black!18}\specialrule{0.25pt}{1.0pt}{1.0pt}\arrayrulecolor{black}
\shortstack[l]{\textbf{QWEN3.5}\\\textbf{27B}} & \cellcolor{dropPurpleCell}\tikz[baseline=(sig.base)]\node[draw=black!80,fill=white!35!dropPurpleCell,line width=0.28pt,rounded corners=1.1pt,inner sep=1.0pt,outer sep=0pt] (sig) {\shortstack{\textbf{-5.3}\\{\fontsize{3.8}{4.1}\selectfont [-10.2, -0.8]}}}; & \cellcolor{dropPurpleCell}\tikz[baseline=(sig.base)]\node[draw=black!80,fill=white!35!dropPurpleCell,line width=0.28pt,rounded corners=1.1pt,inner sep=1.0pt,outer sep=0pt] (sig) {\shortstack{\textbf{-4.4}\\{\fontsize{3.8}{4.1}\selectfont [-9.2, -0.0]}}}; & \cellcolor{dropPurpleCell}\tikz[baseline=(sig.base)]\node[draw=black!80,fill=white!35!dropPurpleCell,line width=0.28pt,rounded corners=1.1pt,inner sep=1.0pt,outer sep=0pt] (sig) {\shortstack{\textbf{-10.0}\\{\fontsize{3.8}{4.1}\selectfont [-17.7, -2.5]}}}; & \cellcolor{dropPurpleCell}\tikz[baseline=(sig.base)]\node[draw=black!80,fill=white!35!dropPurpleCell,line width=0.28pt,rounded corners=1.1pt,inner sep=1.0pt,outer sep=0pt] (sig) {\shortstack{\textbf{-6.6}\\{\fontsize{3.8}{4.1}\selectfont [-10.1, -3.1]}}}; & \cellcolor{dropPurpleCell}\tikz[baseline=(sig.base)]\node[draw=black!80,fill=white!35!dropPurpleCell,line width=0.28pt,rounded corners=1.1pt,inner sep=1.0pt,outer sep=0pt] (sig) {\shortstack{\textbf{-5.0}\\{\fontsize{3.8}{4.1}\selectfont [-9.0, -1.7]}}}; & \cellcolor{riseBlueCell}\shortstack{+0.6\\{\fontsize{3.8}{4.1}\selectfont [-5.5, +6.9]}} & \cellcolor{dropPurpleCell}\shortstack{-3.4\\{\fontsize{3.8}{4.1}\selectfont [-7.4, +0.5]}} & \cellcolor{dropPurpleCell}\shortstack{-2.1\\{\fontsize{3.8}{4.1}\selectfont [-5.5, +0.7]}} & \cellcolor{dropPurpleCell}\shortstack{-1.6\\{\fontsize{3.8}{4.1}\selectfont [-4.6, +1.3]}} & \cellcolor{dropPurpleCell}\tikz[baseline=(sig.base)]\node[draw=black!80,fill=white!35!dropPurpleCell,line width=0.28pt,rounded corners=1.1pt,inner sep=1.0pt,outer sep=0pt] (sig) {\shortstack{\textbf{-3.1}\\{\fontsize{3.8}{4.1}\selectfont [-6.5, -0.5]}}}; & \cellcolor{dropPurpleCell}\shortstack{-4.6\\{\fontsize{3.8}{4.1}\selectfont [-9.8, +0.2]}} & \cellcolor{dropPurpleCell}\shortstack{-1.0\\{\fontsize{3.8}{4.1}\selectfont [-5.3, +3.0]}} & \cellcolor{riseBlueCell}\shortstack{+2.5\\{\fontsize{3.8}{4.1}\selectfont [-2.0, +7.4]}} & \cellcolor{dropPurpleCell}\shortstack{-1.0\\{\fontsize{3.8}{4.1}\selectfont [-3.8, +1.8]}} & \cellcolor{riseBlueCell}\shortstack{+0.4\\{\fontsize{3.8}{4.1}\selectfont [-1.3, +2.9]}} \\
\arrayrulecolor{black!18}\specialrule{0.25pt}{1.0pt}{1.0pt}\arrayrulecolor{black}
\shortstack[l]{\textbf{QWEN3.5}\\\textbf{35B-A3B}} & \cellcolor{dropPurpleCell}\tikz[baseline=(sig.base)]\node[draw=black!80,fill=white!35!dropPurpleCell,line width=0.28pt,rounded corners=1.1pt,inner sep=1.0pt,outer sep=0pt] (sig) {\shortstack{\textbf{-6.5}\\{\fontsize{3.8}{4.1}\selectfont [-11.8, -1.5]}}}; & \cellcolor{dropPurpleCell}\shortstack{-2.6\\{\fontsize{3.8}{4.1}\selectfont [-8.3, +2.9]}} & \cellcolor{dropPurpleCell}\tikz[baseline=(sig.base)]\node[draw=black!80,fill=white!35!dropPurpleCell,line width=0.28pt,rounded corners=1.1pt,inner sep=1.0pt,outer sep=0pt] (sig) {\shortstack{\textbf{-10.4}\\{\fontsize{3.8}{4.1}\selectfont [-16.1, -5.0]}}}; & \cellcolor{dropPurpleCell}\tikz[baseline=(sig.base)]\node[draw=black!80,fill=white!35!dropPurpleCell,line width=0.28pt,rounded corners=1.1pt,inner sep=1.0pt,outer sep=0pt] (sig) {\shortstack{\textbf{-6.5}\\{\fontsize{3.8}{4.1}\selectfont [-9.9, -3.2]}}}; & \cellcolor{dropPurpleCell}\tikz[baseline=(sig.base)]\node[draw=black!80,fill=white!35!dropPurpleCell,line width=0.28pt,rounded corners=1.1pt,inner sep=1.0pt,outer sep=0pt] (sig) {\shortstack{\textbf{-5.4}\\{\fontsize{3.8}{4.1}\selectfont [-10.3, -0.9]}}}; & \cellcolor{dropPurpleCell}\shortstack{-3.7\\{\fontsize{3.8}{4.1}\selectfont [-10.2, +2.5]}} & \cellcolor{dropPurpleCell}\shortstack{-3.7\\{\fontsize{3.8}{4.1}\selectfont [-9.4, +1.9]}} & \cellcolor{dropPurpleCell}\tikz[baseline=(sig.base)]\node[draw=black!80,fill=white!35!dropPurpleCell,line width=0.28pt,rounded corners=1.1pt,inner sep=1.0pt,outer sep=0pt] (sig) {\shortstack{\textbf{-5.1}\\{\fontsize{3.8}{4.1}\selectfont [-10.1, -0.7]}}}; & \cellcolor{dropPurpleCell}\tikz[baseline=(sig.base)]\node[draw=black!80,fill=white!35!dropPurpleCell,line width=0.28pt,rounded corners=1.1pt,inner sep=1.0pt,outer sep=0pt] (sig) {\shortstack{\textbf{-4.2}\\{\fontsize{3.8}{4.1}\selectfont [-7.5, -1.0]}}}; & \cellcolor{dropPurpleCell}\tikz[baseline=(sig.base)]\node[draw=black!80,fill=white!35!dropPurpleCell,line width=0.28pt,rounded corners=1.1pt,inner sep=1.0pt,outer sep=0pt] (sig) {\shortstack{\textbf{-3.3}\\{\fontsize{3.8}{4.1}\selectfont [-7.0, -0.2]}}}; & \cellcolor{riseBlueCell}\shortstack{+2.2\\{\fontsize{3.8}{4.1}\selectfont [-2.4, +7.2]}} & \cellcolor{riseBlueCell}\shortstack{+1.0\\{\fontsize{3.8}{4.1}\selectfont [-2.1, +4.5]}} & \cellcolor{riseBlueCell}\shortstack{+2.0\\{\fontsize{3.8}{4.1}\selectfont [-0.3, +5.2]}} & \cellcolor{riseBlueCell}\shortstack{+1.7\\{\fontsize{3.8}{4.1}\selectfont [-0.4, +4.0]}} & \cellcolor{riseBlueCell}\shortstack{+2.6\\{\fontsize{3.8}{4.1}\selectfont [-0.2, +6.2]}} \\
\arrayrulecolor{black!18}\specialrule{0.25pt}{1.0pt}{1.0pt}\arrayrulecolor{black}
\shortstack[l]{\textbf{LLAMA-3.1}\\\textbf{8B-INSTRUCT}} & \cellcolor{dropPurpleCell}\shortstack{-3.0\\{\fontsize{3.8}{4.1}\selectfont [-9.5, +3.6]}} & \cellcolor{dropPurpleCell}\shortstack{-4.1\\{\fontsize{3.8}{4.1}\selectfont [-9.7, +1.4]}} & \cellcolor{dropPurpleCell}\tikz[baseline=(sig.base)]\node[draw=black!80,fill=white!35!dropPurpleCell,line width=0.28pt,rounded corners=1.1pt,inner sep=1.0pt,outer sep=0pt] (sig) {\shortstack{\textbf{-12.6}\\{\fontsize{3.8}{4.1}\selectfont [-19.1, -6.4]}}}; & \cellcolor{dropPurpleCell}\tikz[baseline=(sig.base)]\node[draw=black!80,fill=white!35!dropPurpleCell,line width=0.28pt,rounded corners=1.1pt,inner sep=1.0pt,outer sep=0pt] (sig) {\shortstack{\textbf{-6.6}\\{\fontsize{3.8}{4.1}\selectfont [-10.4, -2.8]}}}; & \cellcolor{dropPurpleCell}\shortstack{-1.9\\{\fontsize{3.8}{4.1}\selectfont [-6.2, +2.4]}} & \cellcolor{riseBlueCell}\shortstack{+2.5\\{\fontsize{3.8}{4.1}\selectfont [-3.0, +8.2]}} & \cellcolor{riseBlueCell}\shortstack{+2.3\\{\fontsize{3.8}{4.1}\selectfont [-4.9, +9.6]}} & \cellcolor{dropPurpleCell}\shortstack{-0.3\\{\fontsize{3.8}{4.1}\selectfont [-2.7, +2.1]}} & \cellcolor{riseBlueCell}\shortstack{+1.5\\{\fontsize{3.8}{4.1}\selectfont [-1.6, +4.6]}} & \cellcolor{dropPurpleCell}\shortstack{-1.1\\{\fontsize{3.8}{4.1}\selectfont [-5.4, +2.9]}} & \cellcolor{dropPurpleCell}\shortstack{-1.9\\{\fontsize{3.8}{4.1}\selectfont [-6.8, +2.8]}} & \cellcolor{dropPurpleCell}\shortstack{-2.3\\{\fontsize{3.8}{4.1}\selectfont [-9.6, +5.0]}} & \cellcolor{riseBlueCell}\shortstack{+3.1\\{\fontsize{3.8}{4.1}\selectfont [-2.8, +9.4]}} & \cellcolor{dropPurpleCell}\shortstack{-0.4\\{\fontsize{3.8}{4.1}\selectfont [-4.2, +3.5]}} & \cellcolor{riseBlueCell}\shortstack{+2.9\\{\fontsize{3.8}{4.1}\selectfont [-3.5, +9.5]}} \\
\arrayrulecolor{black!18}\specialrule{0.25pt}{1.0pt}{1.0pt}\arrayrulecolor{black}
\tikzmarknode[inner sep=0pt]{dataset100_reason_bottom}{\shortstack[l]{\textbf{CLAUDE}\\\textbf{SONNET 4.6}}} & \cellcolor{dropPurpleCell}\tikz[baseline=(sig.base)]\node[draw=black!80,fill=white!35!dropPurpleCell,line width=0.28pt,rounded corners=1.1pt,inner sep=1.0pt,outer sep=0pt] (sig) {\shortstack{\textbf{-7.2}\\{\fontsize{3.8}{4.1}\selectfont [-12.4, -2.1]}}}; & \cellcolor{dropPurpleCell}\tikz[baseline=(sig.base)]\node[draw=black!80,fill=white!35!dropPurpleCell,line width=0.28pt,rounded corners=1.1pt,inner sep=1.0pt,outer sep=0pt] (sig) {\shortstack{\textbf{-6.7}\\{\fontsize{3.8}{4.1}\selectfont [-10.8, -3.1]}}}; & \cellcolor{dropPurpleCell}\tikz[baseline=(sig.base)]\node[draw=black!80,fill=white!35!dropPurpleCell,line width=0.28pt,rounded corners=1.1pt,inner sep=1.0pt,outer sep=0pt] (sig) {\shortstack{\textbf{-9.6}\\{\fontsize{3.8}{4.1}\selectfont [-15.1, -4.6]}}}; & \cellcolor{dropPurpleCell}\tikz[baseline=(sig.base)]\node[draw=black!80,fill=white!35!dropPurpleCell,line width=0.28pt,rounded corners=1.1pt,inner sep=1.0pt,outer sep=0pt] (sig) {\shortstack{\textbf{-7.8}\\{\fontsize{3.8}{4.1}\selectfont [-10.8, -5.0]}}}; & \cellcolor{dropPurpleCell}\shortstack{-3.1\\{\fontsize{3.8}{4.1}\selectfont [-7.3, +0.8]}} & \cellcolor{dropPurpleCell}\shortstack{-1.5\\{\fontsize{3.8}{4.1}\selectfont [-7.3, +4.4]}} & \cellcolor{dropPurpleCell}\shortstack{-0.4\\{\fontsize{3.8}{4.1}\selectfont [-4.1, +3.3]}} & \cellcolor{dropPurpleCell}\shortstack{-3.2\\{\fontsize{3.8}{4.1}\selectfont [-7.0, +0.4]}} & \cellcolor{dropPurpleCell}\shortstack{-1.7\\{\fontsize{3.8}{4.1}\selectfont [-4.5, +1.0]}} & \cellcolor{dropPurpleCell}\shortstack{-1.7\\{\fontsize{3.8}{4.1}\selectfont [-5.4, +1.8]}} & \cellcolor{dropPurpleCell}\tikz[baseline=(sig.base)]\node[draw=black!80,fill=white!35!dropPurpleCell,line width=0.28pt,rounded corners=1.1pt,inner sep=1.0pt,outer sep=0pt] (sig) {\shortstack{\textbf{-5.3}\\{\fontsize{3.8}{4.1}\selectfont [-9.5, -1.7]}}}; & \cellcolor{riseBlueCell}\shortstack{+1.9\\{\fontsize{3.8}{4.1}\selectfont [-2.7, +6.8]}} & \cellcolor{dropPurpleCell}\shortstack{-2.4\\{\fontsize{3.8}{4.1}\selectfont [-7.0, +2.1]}} & \cellcolor{dropPurpleCell}\shortstack{-1.9\\{\fontsize{3.8}{4.1}\selectfont [-4.5, +0.6]}} & \cellcolor{dropPurpleCell}\shortstack{-1.2\\{\fontsize{3.8}{4.1}\selectfont [-4.3, +1.7]}} \\
\bottomrule
\end{tabular}%
\begin{tikzpicture}[remember picture,overlay]
\draw[black!55,line width=0.8pt,rounded corners=2.8pt] ([xshift=-2.0pt,yshift=1.2pt]dataset100_variant_reason_top.north west) rectangle ([xshift=2.0pt,yshift=-1.2pt]dataset100_variant_reason_top.south east |- dataset100_reason_bottom.south);
\draw[black!55,line width=0.8pt,rounded corners=2.8pt] ([xshift=-2.0pt,yshift=1.2pt]dataset100_invariant_reason_top.north west) rectangle ([xshift=2.0pt,yshift=-1.2pt]dataset100_invariant_reason_top.south east |- dataset100_reason_bottom.south);
\draw[black!55,line width=0.8pt,rounded corners=2.8pt] ([xshift=-2.0pt,yshift=1.2pt]dataset100_refusal_reason_top.north west) rectangle ([xshift=2.0pt,yshift=-1.2pt]dataset100_refusal_reason_top.south east |- dataset100_reason_bottom.south);
\end{tikzpicture}%
}
\end{table}

\subsection{The Impact of Entity Replacement on LLMs' Reasoning Performance}
\label{subsec:parametric_memory_impact}

Our first experiment investigates whether entity replacement influences model reasoning performance. The results are reported in Table \ref{tab:question-answer-type-drop}. 

\paragraph{Reasoning questions with Variant answers} For variant answers (see Section~\ref{subsec:fictional_dataset_generation}), all models exhibit a drop in reasoning performance when working on \fictitiousterm{} documents. Moreover, this drop is statistically significant for all 9 models tested. This result is driven by inference-class questions (see Section~\ref{subsec:factual_dataset_construction}), with significant drops in the 10\% range. Temporal-class questions also contribute, with significant drops for \unchangednumber{\ResultTemporalSignificantModels{}} models in the \unchangednumber{\ResultTemporalDropMin\%} to \unchangednumber{\ResultTemporalDropMax\%} range. Arithmetic-class questions show significant drops for \unchangednumber{\ResultArithmeticSignificantModels{}} models, in the \unchangednumber{\ResultArithmeticDropMin\%} to \unchangednumber{\ResultArithmeticDropMax\%} range. This ranking should not come as a surprise, since arithmetic questions are much more prevalent than other categories of questions on many mathematical reasoning benchmarks and post-training datasets. Temporal questions, though much less prevalent, are relatively straightforward to systematize and generally share similar patterns among each other. Inference questions, however, are much broader in the patterns they follow: they are the most difficult ones to shortcut via pattern matching. Overall, the Variant-class results confirm that there is indeed a familiarity bias in LLM reasoning.

\paragraph{Reasoning questions with Invariant answers} \scorerevision{For invariant answers, aggregate reasoning changes are heterogeneous across models, ranging from a 6.0-point decrease to a 1.5-point increase. Only 3 of the 9 models show a statistically significant decrease at the 95\% confidence level.} \revision{Given that answers don't change, we further investigated whether these drops on invariant answers are due to models wrongly using familiar named or numerical entities to map the context to memorized structure and report the results in Appendix~\ref{appendix:cue_ablation}.}

\paragraph{Refusal answers and Extractive questions.} The picture is significantly more muddled for refusals, with only a couple of barely significant deviations in performance going in different directions. Overall, a non-statistically significant hint of an inverse bias can be gleaned from the data, with models being more likely to venture an answer despite insufficient supporting evidence when in the presence of factual information. In the factual setting, models rely on memorized knowledge even when the document lacks sufficient evidence, causing them to answer when they should refuse. In the fictitious setting this memorized fallback is unavailable, making models more likely to recognize the missing evidence and correctly refuse. \revision{On extractive questions with variant answers, performance decreases for all \unchangednumber{\ResultVariantExtractiveModels{}} models and significantly for \unchangednumber{\ResultVariantExtractiveSignificantModels}} of them. 

\revision{Beyond the aggregate results in Table~\ref{tab:question-answer-type-drop}, Appendix~\ref{appendix:directional_flips} compares paired model responses in both directions: factual-to-fictitious and fictitious-to-factual. Models are more likely to answer a fictitious task incorrectly when they answer its factual counterpart correctly than to answer a factual task incorrectly when they answer its fictitious counterpart correctly. This directional asymmetry favors the factual setting, consistent with models benefiting from entity familiarity when performing contextual reasoning. It does not, however, establish direct factual-answer copying; we test the \textit{Strong Parametric Shortcut Hypothesis} in Section~\ref{sec:parametric_shortcut}.}

\subsection{Varying the Proportion of Replaced Entities}
\label{subsec:varying_proportion}

The previous results from Table \ref{tab:question-answer-type-drop} show that the models yield significant reductions in reasoning performance when replacing all entities. In this experiment, our aim is to probe knowledge conflict when factual anchors and counterfactual relations coexist. To this end, we performed the same analysis described in Section \ref{subsec:parametric_memory_impact} on partially fictitious documents, where a percentage of randomly selected entities are replaced while the others are kept factual. Once again, we ensure that every assessed document respects all the rules associated with its template after partial entity substitution.

\begin{figure}[!t]
\centering
\includegraphics[width=\linewidth]{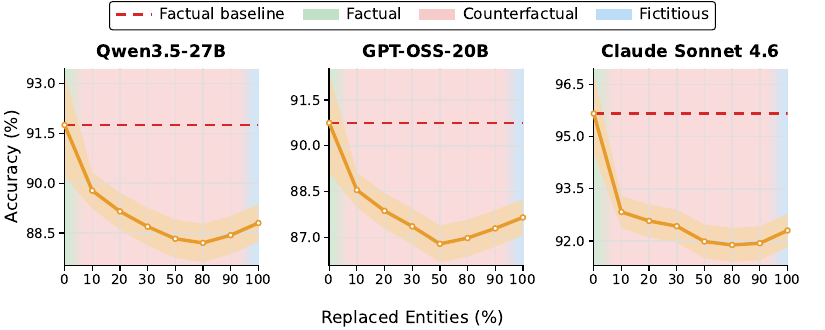}
\caption{\scorerevision{Accuracy of Qwen3.5-27B, GPT-OSS-20B, and Claude Sonnet 4.6 as the proportion of entities replaced with fictitious ones increases. Dashed red lines show factual accuracy; orange shaded bands show 95\% confidence intervals.}}
\label{fig:accuracy_by_fictiobal_replacement_proportions}
\end{figure}

Figure \ref{fig:accuracy_by_fictiobal_replacement_proportions} reports accuracy from the factual baseline (i.e., 0\% replacement, represented with horizontal dashed red lines) to full fictitious replacement (i.e., 100\% replacement, matching the setup from Section \ref{subsec:parametric_memory_impact}) for Qwen3.5-27B, GPT-OSS-20B, and Claude Sonnet 4.6 (we report the remaining four models in Appendix \ref{appendix:partial_replacement_overlays}). 

Across all models, accuracy declines as the replacement proportion rises from 0\% toward 50\%, consistent with the familiarity effect established in Section 4.1. Beyond ~50\%, accuracy partially recovers toward the fully fictitious setting: the interior regime introduces a second factor — conflict between parametric knowledge and contextual assertions about still-familiar entities — that the endpoint comparison of Section 4.1 does not probe. Section 4.2 therefore measures reasoning-task performance under knowledge conflict \citep{longpre-etal-2021-entity}, complementing the pure-familiarity comparison of Section 4.1.

\subsection{Chain-of-Thought Effect}
\label{sec:cot_effect}

To test the effect that chain-of-thought (CoT) post-training has on reasoning performance, we compare OLMo-3-7B-Instruct (no CoT) with OLMo-3-7B-Think (with CoT) on reasoning questions (arithmetic, temporal, and inference). CoT improves absolute performance in both settings, with gains of 15.00\% points on factual reasoning questions and 12.62\% on their fictitious counterparts. However, the factual--fictitious performance drop is also 2.38\% larger: accuracy rises, robustness does not. We show in \revision{Appendix~\ref{appendix:qualitative_cot} a comparative example where OLMo-Think (with CoT) gave the correct answer as opposed to OLMo-Instruct (without CoT).}

\Needspace{14\baselineskip}
\begingroup
\setlength{\intextsep}{4pt}
\setlength{\columnsep}{12pt}

\begin{wraptable}{r}{0.42\linewidth}
\vspace{-0.4\baselineskip}
\begingroup
\centering
\small
\revisiontablecolor
\revisioncaptionstyle

\renewcommand{\arraystretch}{1.14}
\setlength{\tabcolsep}{1.2pt}

\captionsetup{
  hypcap=false,
  justification=raggedright,
  singlelinecheck=false,
  skip=4pt
}

\captionof{table}{%
  \textbf{\texttt{GPT-OSS-20B} reasoning~effort ablation.}%
}
\label{tab:dataset100-reasoning-effort}

\begin{tabularx}{\linewidth}{
  @{}
  l
  *{3}{>{\centering\arraybackslash}X}
  @{}
}
\toprule
Effort &
\shortstack{Factual\\acc. (\%)} &
\shortstack{\Fictitiousterm{}\\acc. (\%)} &
\shortstack{Gap\\(\%)} \\
\midrule

Low &
\shortstack{90.75\\{\tiny [89.17, 92.25]}} &
\shortstack{87.65\\{\tiny [86.15, 89.08]}} &
\shortstack{3.10\\{\tiny [1.44, 4.77]}} \\

Medium &
\shortstack{94.25\\{\tiny [93.00, 95.50]}} &
\shortstack{89.90\\{\tiny [88.35, 91.40]}} &
\shortstack{4.35\\{\tiny [2.89, 5.86]}} \\

High &
\shortstack{92.67\\{\tiny [91.17, 94.17]}} &
\shortstack{88.31\\{\tiny [86.68, 89.87]}} &
\shortstack{4.36\\{\tiny [2.80, 5.98]}} \\

\bottomrule
\end{tabularx}
\endgroup

\vspace{-0.25\baselineskip}
\end{wraptable}

\paragraph{\scorerevision{Does greater reasoning effort improve performance?}}
GPT-OSS models allow us to vary the effort allocated to the reasoning
chain (low, medium, and high), enabling us to test its effect on model
performance \scorerevision{(Table~\ref{tab:dataset100-reasoning-effort})}. Factual/fictitious accuracy rises
\unchangednumber{\ResultEffortMediumMinusLowFactual{}}/
\unchangednumber{\ResultEffortMediumMinusLowFictional{}}
points from low to medium, but falls
\postreviewnumber{\ResultEffortMediumMinusHighFactual{}}/
\postreviewnumber{\ResultEffortMediumMinusHighFictional{}}
at high effort (paired \postreviewnumber{95\% CIs} exclude zero).
Thus, more reasoning effort is not always better. The gap remains
significant at every setting, but its pairwise changes are not;
additional effort does not remove it.

\par
\WFclear
\endgroup

\subsection{Strong Parametric Shortcut Hypothesis}
\label{sec:parametric_shortcut}

The performance drops observed in \fictitiousterm{} variants (see Section \ref{subsec:parametric_memory_impact}) show that models are affected by the familiarity of the entities appearing in the document. However, this result alone does not establish the \textit{Strong Parametric Shortcut Hypothesis}. A drop in performance may indicate a broader memory or familiarity bias, but the hypothesis would require a more specific failure mode: when the model fails on a \fictitiousterm{} variant, it should incorrectly answer with the factual answer associated with the original example, suggesting that it bypassed contextual reasoning completely by recalling memorized parametric knowledge.

\begin{table}[!t]
\centering
\begingroup
\newcommand{\countfont}{\fontsize{7.1}{7.3}\selectfont}
\newcommand{\shortcutcell}[2]{#1\,{\countfont (#2)}}
\fontsize{9.0}{9.2}\selectfont
\revisiontablecolor
\revisioncaptionstyle
\renewcommand{\arraystretch}{1.08}
\setlength{\tabcolsep}{1.3pt}
\captionsetup{hypcap=false,justification=raggedright,singlelinecheck=false}
\begin{minipage}{\linewidth}
\captionof{table}{Shortcut Rate (\%). \postreviewrevision{Rate = judge weighted matches over failed variant examples (counts in parentheses).} \textit{Reason.} aggregates reasoning questions.}
\label{tab:variant-factual-answer-copying}
\end{minipage}
\par\vspace{2pt}
\begin{tabular*}{\linewidth}{@{\extracolsep{\fill}}l*{5}{c}@{}}
\toprule
\hspace{0.15cm}{\scriptsize\itshape Question Type} & \textbf{Arith.} & \textbf{Temp.} & \textbf{Infer.} & \makebox[1.45cm][c]{\tikzmarknode[inner sep=0.5pt,minimum width=1.45cm]{reason_top}{\textbf{Reason.}}} & \textbf{Extr.} \\
\midrule
\textbf{OLMO-3 7B-THINK} & \shortcutcell{1.0}{344} & \shortcutcell{1.4}{336} & \shortcutcell{0.5}{314} & \makebox[1.45cm][c]{\shortcutcell{1.0}{994}} & \shortcutcell{0.8}{123} \\
\arrayrulecolor{black!18}\specialrule{0.25pt}{0.5pt}{0.5pt}\arrayrulecolor{black}
\textbf{OLMO-3 7B-INSTRUCT} & \shortcutcell{2.0}{600} & \shortcutcell{3.7}{567} & \shortcutcell{1.9}{467} & \makebox[1.45cm][c]{\shortcutcell{2.5}{1634}} & \shortcutcell{0.8}{130} \\
\arrayrulecolor{black!18}\specialrule{0.25pt}{0.5pt}{0.5pt}\arrayrulecolor{black}
\textbf{GPT-OSS 20B} & \shortcutcell{1.6}{118} & \shortcutcell{3.9}{98} & \shortcutcell{0.9}{213} & \makebox[1.45cm][c]{\shortcutcell{2.2}{429}} & \shortcutcell{1.0}{80} \\
\arrayrulecolor{black!18}\specialrule{0.25pt}{0.5pt}{0.5pt}\arrayrulecolor{black}
\textbf{GEMMA-4 26B-A4B-IT} & \shortcutcell{1.4}{189} & \shortcutcell{0.6}{186} & \shortcutcell{1.3}{271} & \makebox[1.45cm][c]{\shortcutcell{1.1}{646}} & \shortcutcell{1.0}{52} \\
\arrayrulecolor{black!18}\specialrule{0.25pt}{0.5pt}{0.5pt}\arrayrulecolor{black}
\textbf{GPT-OSS 120B} & \shortcutcell{0.8}{107} & \shortcutcell{2.0}{94} & \shortcutcell{1.2}{188} & \makebox[1.45cm][c]{\shortcutcell{1.3}{389}} & \shortcutcell{1.0}{78} \\
\arrayrulecolor{black!18}\specialrule{0.25pt}{0.5pt}{0.5pt}\arrayrulecolor{black}
\textbf{QWEN3.5 27B} & \shortcutcell{1.9}{253} & \shortcutcell{0.4}{134} & \shortcutcell{1.2}{200} & \makebox[1.45cm][c]{\shortcutcell{1.2}{587}} & \shortcutcell{1.0}{60} \\
\arrayrulecolor{black!18}\specialrule{0.25pt}{0.5pt}{0.5pt}\arrayrulecolor{black}
\textbf{QWEN3.5 35B-A3B} & \shortcutcell{1.6}{305} & \shortcutcell{1.6}{186} & \shortcutcell{3.9}{254} & \makebox[1.45cm][c]{\shortcutcell{2.3}{745}} & \shortcutcell{1.0}{74} \\
\arrayrulecolor{black!18}\specialrule{0.25pt}{0.5pt}{0.5pt}\arrayrulecolor{black}
\textbf{LLAMA-3.1 8B-INSTRUCT} & \shortcutcell{1.8}{360} & \shortcutcell{2.6}{271} & \shortcutcell{5.2}{256} & \makebox[1.45cm][c]{\shortcutcell{3.2}{887}} & \shortcutcell{0.0}{39} \\
\arrayrulecolor{black!18}\specialrule{0.25pt}{0.5pt}{0.5pt}\arrayrulecolor{black}
\textbf{CLAUDE SONNET 4.6} & \shortcutcell{1.2}{102} & \shortcutcell{1.4}{87} & \shortcutcell{3.3}{136} & \makebox[1.45cm][c]{\tikzmarknode[inner sep=0.5pt,minimum width=1.45cm]{reason_bottom}{\shortcutcell{2.0}{325}}} & \shortcutcell{1.0}{41} \\
\bottomrule
\end{tabular*}%
\begin{tikzpicture}[remember picture,overlay]
\draw[black!55,line width=0.8pt,rounded corners=2.8pt] ([xshift=-0.7pt,yshift=1.2pt]reason_top.north west) rectangle ([xshift=0.7pt,yshift=-1.2pt]reason_bottom.south east);
\end{tikzpicture}%
\endgroup

\end{table}

To test this hypothesis directly, we compute the \textit{Shortcut Rate} reported in Table~\ref{tab:variant-factual-answer-copying}. This measures how often the model's incorrect prediction matches the corresponding factual answer.
For each model, we restrict the analysis to failed \fictitiousterm{} variant examples. \revision{The resulting rates are consistently low. Across reasoning questions, the aggregate rate ranges from \unchangednumber{\ResultShortcutReasoningMin\%} to \unchangednumber{\ResultShortcutReasoningMax\%}, and extractive rates range from \unchangednumber{\ResultShortcutExtractiveMin\%} to \unchangednumber{\ResultShortcutExtractiveMax\%}.} Thus, even when models fail under \fictitiousterm{} substitutions, they rarely fail by simply reproducing the factual answer. These results suggest that the memory effect identified in our main performance analysis should not be interpreted as a simple recall shortcut. Parametric memory appears to influence reasoning performance, producing a clear \textit{memory} or \textit{familiarity bias}, but this influence does not usually translate into direct factual-answer copying. \revision{The results do not identify the mechanisms underlying these errors.} The non-trivial interaction between prior knowledge and reasoning ability deserves further investigation, and \textbf{\benchmarkname} provides a controlled setting for analyses.

\subsection{\revision{Evaluation Contract}}
\label{sec:benchmark_contract}

\begin{table}[!b]
  \centering
  \caption{\revision{\leaderboardicon{trophy}\;\textbf{MemoReason leaderboard.} Models are ranked by accuracy in the fully fictitious setting. $\uparrow$~indicates higher is better; $\downarrow$~indicates lower is better.}}
  \label{tab:dataset100-report-card}
  \definecolor{winnerGold}{HTML}{FFF3C4}
\definecolor{runnerSilver}{HTML}{F1F3F5}
\definecolor{thirdBronze}{HTML}{F7E2D2}

\centering
\small
\revisiontablecolor
\renewcommand{\arraystretch}{1.08}
\setlength{\tabcolsep}{3.0pt}

\begin{tabularx}{\linewidth}{@{}r l *{5}{>{\centering\arraybackslash}X}@{}}
\toprule
Rank & Model &
\shortstack{\Fictitiousterm{}\\acc. (\%)\,\revision{$\uparrow$}} &
\shortstack{\revision{Factual}\\acc. (\%)\,\revision{$\uparrow$}} &
\shortstack{Gap\\(\%)\,\revision{$\downarrow$}} &
\shortstack{Error\\ratio\,\revision{$\downarrow$}} &
\shortstack{Shortcut\\Rate (\%)\,\revision{$\downarrow$}} \\
\midrule

\rowcolor{winnerGold}
\leaderboardicon{gold_medal}\,1 &
\modellogo{anthropic}CLAUDE SONNET 4.6 &
92.30 & 95.67 & 3.37 & 1.78$\times$ & 2.01 \\

\rowcolor{runnerSilver}
\leaderboardicon{silver_medal}\,2 &
\modellogo{openai}GPT-OSS 120B &
90.28 & 93.67 & 3.38 & 1.53$\times$ & 1.33 \\

\rowcolor{thirdBronze}
\leaderboardicon{bronze_medal}\,3 &
\modellogo{qwen}QWEN3.5 27B &
88.80 & 91.75 & 2.95 & 1.36$\times$ & 1.18 \\

4 & \modellogo{openai}GPT-OSS 20B &
87.65 & 90.75 & 3.10 & 1.34$\times$ & 2.18 \\

5 & \modellogo{qwen}QWEN3.5 35B-A3B &
86.34 & 89.08 & 2.74 & 1.25$\times$ & 2.33 \\

6 & \modellogo{google-deepmind}GEMMA-4 26B-A4B-IT &
85.91 & 87.25 & 1.34 & 1.11$\times$ & 1.10 \\

7 & \modellogo{meta-llama}LLAMA-3.1 8B-INSTRUCT &
76.88 & 78.25 & 1.37 & 1.06$\times$ & 3.19 \\

8 & \modellogo{ai2}OLMO-3 7B-THINK &
75.61 & 80.92 & 5.31 & 1.28$\times$ & 0.98 \\

9 & \modellogo{ai2}OLMO-3 7B-INSTRUCT &
66.42 & 69.25 & 2.83 & 1.09$\times$ & 2.53 \\

\bottomrule
\end{tabularx}

\end{table}

\revision{MemoReason is intended not only to detect familiarity effects, but also to support meaningful comparisons between models. A single score would be misleading: fictitious accuracy reflects overall reasoning ability, whereas a small factual--fictitious gap may indicate robustness or simply poor performance in both settings. We therefore report complementary diagnostics. Table~\ref{tab:dataset100-report-card} ranks models by fictitious accuracy, while also reporting the factual--fictitious gap, the error ratio \scorerevision{$(1-\mathrm{Acc}_{\mathrm{fict}})/(1-\mathrm{Acc}_{\mathrm{fact}})$}---the factor by which the error rate changes after replacement---and the Shortcut Rate, which measures direct factual-answer reproduction among fictitious failures. These metrics should be interpreted jointly. For example, Llama-3.1-8B-Instruct has a relatively small 1.37-point gap but ranks seventh in fictitious accuracy, and its reasoning performance still decreases significantly.}

\revision{High accuracy can likewise conceal a substantial relative error penalty. Claude Sonnet 4.6 achieves 95.67\% factual accuracy, but this falls to 92.30\% in the fictitious setting (Table 5). The statistically significant 3.37-point gap shows that entity familiarity still affects performance, even for the strongest evaluated model. Its high factual accuracy therefore does not establish benchmark saturation: achieving comparably high accuracy on fictitious tasks while reducing this gap remains an open challenge.}

\vspace{-0.25cm}

\section{Conclusion}

We introduced \textbf{MemoReason}, a human-curated benchmark pairing factual tasks with structurally identical \fictitiousterm{} counterparts to study familiarity effects on contextual reasoning. \revision{Our evaluation reveals statistically significant accuracy drops of up to \unchangednumber{\ResultMaxSignificantVariantDrop\%} in the fictitious setting.} These failures rarely reproduce factual answers directly, suggesting that familiarity affects reasoning beyond simple answer copying. \revision{Identifying the mechanisms underlying these performance differences remains an open question, and MemoReason provides a controlled setting for investigating them. Its regenerable templates support paired comparisons between factual, partially replaced, and fully fictitious documents, as well as targeted interventions on named entities or numerical and temporal values. This enables researchers to test specific explanations for performance changes while preserving task structure and specified reasoning operations. Regenerating fresh fictitious instances also helps mitigate evaluation-data leakage. The accompanying annotation interface as well as the generation and evaluation pipeline make these experiments inspectable, reproducible, and extensible to new documents and models. Together, these resources support a more systematic understanding of when familiar knowledge helps contextual reasoning---and when it interferes with it.}

\subsection*{AI use statement}

\revision{Generative AI was used for benchmark pre-annotation and fictitious entity generation. All generated outputs were reviewed and, where necessary, corrected by human annotators, as detailed in Section~\ref{subsec:template_annotation_process}. Generative-AI tools also provided assistance with some software development.}

\subsection*{Reproducibility statement}

\revision{The dataset and code releases are linked below the authors. Section~\ref{subsec:factual_dataset_construction}--\ref{subsec:fictional_dataset_generation} describes source selection, human-in-the-loop template annotation, and fictitious-data generation. Appendix~\ref{appendix:dataset_details} enumerates the released intervention settings; Appendix~\ref{appendix:experimental_setup} records model identifiers, inference settings, hardware, and licenses; Appendix~\ref{appendix:prompts} provides the generation and evaluation prompts; and Appendix~\ref{sec:judge_calibration} documents evaluation calibration. The pipeline preserves raw outputs, scores, confidence intervals, manifests, and checksums binding each run to its dataset; statistical details are specified in Section~\ref{sec:eval_framework}.}

\section*{Acknowledgements}

We would like to thank BNP Paribas and the French National Association for Research and Technology (ANRT) for funding this project under the CIFRE program (2023/1673). We would also like to thank Andrey Krivonogov, Etienne Boisseau, Gregoire Roullier, Lucas Lima De Carvalho, Mathias Vast, Melanie Bervoets, Randa Elmrabet-Tarmach, and all other contributors who helped annotate and validate \textbf{MemoReason}. Their careful review of entity annotations, replacement rules, and question--answer pairs was essential to improving the quality and consistency of the benchmark.

{
    \small
    \bibliography{references}

@inproceedings{modarressi-etal-2024-consistent,
    title = "Consistent Document-level Relation Extraction via Counterfactuals",
    author = {Modarressi, Ali  and
      K{\"o}ksal, Abdullatif  and
      Schuetze, Hinrich},
    editor = "Al-Onaizan, Yaser  and
      Bansal, Mohit  and
      Chen, Yun-Nung",
    booktitle = "Findings of the Association for Computational Linguistics: EMNLP 2024",
    month = nov,
    year = "2024",
    address = "Miami, Florida, USA",
    publisher = "Association for Computational Linguistics",
    url = "https://aclanthology.org/2024.findings-emnlp.672/",
    doi = "10.18653/v1/2024.findings-emnlp.672",
    pages = "11501--11507"
}

@inproceedings{glockner-etal-2025-neoqa,
    title = "{N}eo{QA}: Evidence-based Question Answering with Generated News Events",
    author = "Glockner, Max  and
      Jiang, Xiang  and
      Ribeiro, Leonardo F. R.  and
      Gurevych, Iryna  and
      Dreyer, Markus",
    editor = "Che, Wanxiang  and
      Nabende, Joyce  and
      Shutova, Ekaterina  and
      Pilehvar, Mohammad Taher",
    booktitle = "Findings of the Association for Computational Linguistics: ACL 2025",
    month = jul,
    year = "2025",
    address = "Vienna, Austria",
    publisher = "Association for Computational Linguistics",
    url = "https://aclanthology.org/2025.findings-acl.616/",
    doi = "10.18653/v1/2025.findings-acl.616",
    pages = "11842--11926",
    ISBN = "979-8-89176-256-5"
}

@misc{wu2024cofcastepwisecounterfactualmultihop,
      title={Cofca: A Step-Wise Counterfactual Multi-hop QA benchmark}, 
      author={Jian Wu and Linyi Yang and Zhen Wang and Manabu Okumura and Yue Zhang},
      year={2024},
      eprint={2402.11924},
      archivePrefix={arXiv},
      primaryClass={cs.CL},
      url={https://arxiv.org/abs/2402.11924}, 
}

@inproceedings{tighidet-etal-2025-context,
    title = "Context Copying Modulation: The Role of Entropy Neurons in Managing Parametric and Contextual Knowledge Conflicts",
    author = "Tighidet, Zineddine  and
      Mogini, Andrea  and
      Ben younes, Hedi  and
      Mei, Jiali  and
      Gallinari, Patrick  and
      Piwowarski, Benjamin",
    editor = "Christodoulopoulos, Christos  and
      Chakraborty, Tanmoy  and
      Rose, Carolyn  and
      Peng, Violet",
    booktitle = "Findings of the Association for Computational Linguistics: EMNLP 2025",
    month = nov,
    year = "2025",
    address = "Suzhou, China",
    publisher = "Association for Computational Linguistics",
    url = "https://aclanthology.org/2025.findings-emnlp.1116/",
    doi = "10.18653/v1/2025.findings-emnlp.1116",
    pages = "20469--20481",
    ISBN = "979-8-89176-335-7"
}

@inproceedings{tighidet-etal-2024-probing,
    title = "Probing Language Models on Their Knowledge Source",
    author = "Tighidet, Zineddine  and
      Mei, Jiali  and
      Piwowarski, Benjamin  and
      Gallinari, Patrick",
    editor = "Belinkov, Yonatan  and
      Kim, Najoung  and
      Jumelet, Jaap  and
      Mohebbi, Hosein  and
      Mueller, Aaron  and
      Chen, Hanjie",
    booktitle = "Proceedings of the 7th BlackboxNLP Workshop: Analyzing and Interpreting Neural Networks for NLP",
    month = nov,
    year = "2024",
    address = "Miami, Florida, US",
    publisher = "Association for Computational Linguistics",
    url = "https://aclanthology.org/2024.blackboxnlp-1.35/",
    doi = "10.18653/v1/2024.blackboxnlp-1.35",
    pages = "604--614"
}

@inproceedings{
mirzadeh2025gsmsymbolic,
title={{GSM}-Symbolic: Understanding the Limitations of Mathematical Reasoning in Large Language Models},
author={Seyed Iman Mirzadeh and Keivan Alizadeh and Hooman Shahrokhi and Oncel Tuzel and Samy Bengio and Mehrdad Farajtabar},
booktitle={The Thirteenth International Conference on Learning Representations},
year={2025},
url={https://openreview.net/forum?id=AjXkRZIvjB}
}

@misc{razeghi2022impactpretrainingtermfrequencies,
      title={Impact of Pretraining Term Frequencies on Few-Shot Reasoning}, 
      author={Yasaman Razeghi and Robert L. Logan IV and Matt Gardner and Sameer Singh},
      year={2022},
      eprint={2202.07206},
      archivePrefix={arXiv},
      primaryClass={cs.CL},
      url={https://arxiv.org/abs/2202.07206}, 
}

@inproceedings{stolfo-etal-2023-causal,
    title = "A Causal Framework to Quantify the Robustness of Mathematical Reasoning with Language Models",
    author = {Stolfo, Alessandro  and
      Jin, Zhijing  and
      Shridhar, Kumar  and
      Sch{\"o}lkopf, Bernhard  and
      Sachan, Mrinmaya},
    editor = "Rogers, Anna  and
      Boyd-Graber, Jordan  and
      Okazaki, Naoaki",
    booktitle = "Proceedings of the 61st Annual Meeting of the Association for Computational Linguistics (Volume 1: Long Papers)",
    month = jul,
    year = "2023",
    address = "Toronto, Canada",
    publisher = "Association for Computational Linguistics",
    url = "https://aclanthology.org/2023.acl-long.32/",
    doi = "10.18653/v1/2023.acl-long.32",
    pages = "545--561"
}

@inproceedings{hong-etal-2025-evaluating,
    title = "Evaluating {LLM}s' Mathematical and Coding Competency through Ontology-guided Interventions",
    author = "Hong, Pengfei  and
      Majumder, Navonil  and
      Ghosal, Deepanway  and
      Aditya, Somak  and
      Mihalcea, Rada  and
      Poria, Soujanya",
    editor = "Che, Wanxiang  and
      Nabende, Joyce  and
      Shutova, Ekaterina  and
      Pilehvar, Mohammad Taher",
    booktitle = "Findings of the Association for Computational Linguistics: ACL 2025",
    month = jul,
    year = "2025",
    address = "Vienna, Austria",
    publisher = "Association for Computational Linguistics",
    url = "https://aclanthology.org/2025.findings-acl.1172/",
    doi = "10.18653/v1/2025.findings-acl.1172",
    pages = "22811--22849",
    ISBN = "979-8-89176-256-5"
}

@misc{monteiro2024repliqaquestionansweringdatasetbenchmarking,
      title={RepLiQA: A Question-Answering Dataset for Benchmarking LLMs on Unseen Reference Content}, 
      author={Joao Monteiro and Pierre-Andre Noel and Etienne Marcotte and Sai Rajeswar and Valentina Zantedeschi and David Vazquez and Nicolas Chapados and Christopher Pal and Perouz Taslakian},
      year={2024},
      eprint={2406.11811},
      archivePrefix={arXiv},
      primaryClass={cs.CL},
      url={https://arxiv.org/abs/2406.11811}, 
}

@misc{shrestha2025mathematicalreasoninglargelanguage,
      title={Mathematical Reasoning in Large Language Models: Assessing Logical and Arithmetic Errors across Wide Numerical Ranges}, 
      author={Safal Shrestha and Minwu Kim and Keith Ross},
      year={2025},
      eprint={2502.08680},
      archivePrefix={arXiv},
      primaryClass={cs.LG},
      url={https://arxiv.org/abs/2502.08680}, 
}

@misc{chen2021codex,
  title={Evaluating Large Language Models Trained on Code},
  author={Mark Chen and Jerry Tworek and Heewoo Jun and Qiming Yuan and Henrique Ponde de Oliveira Pinto and Jared Kaplan and Harri Edwards and Yuri Burda and Nicholas Joseph and Greg Brockman and Alex Ray and Raul Puri and Gretchen Krueger and Michael Petrov and Heidy Khlaaf and Girish Sastry and Pamela Mishkin and Brooke Chan and Scott Gray and Nick Ryder and Mikhail Pavlov and Alethea Power and Lukasz Kaiser and Mohammad Bavarian and Clemens Winter and Philippe Tillet and Felipe Petroski Such and Dave Cummings and Matthias Plappert and Fotios Chantzis and Elizabeth Barnes and Ariel Herbert-Voss and William Hebgen Guss and Alex Nichol and Alex Paino and Nikolas Tezak and Jie Tang and Igor Babuschkin and Suchir Balaji and Shantanu Jain and William Saunders and Christopher Hesse and Andrew N. Carr and Jan Leike and Josh Achiam and Vedant Misra and Evan Morikawa and Alec Radford and Matthew Knight and Miles Brundage and Mira Murati and Katie Mayer and Peter Welinder and Bob McGrew and Dario Amodei and Sam McCandlish and Ilya Sutskever and Wojciech Zaremba},
  year={2021},
  eprint={2107.03374},
  archivePrefix={arXiv},
  primaryClass={cs.LG}
}

@inproceedings{
hendrycks2021measuring,
title={Measuring Mathematical Problem Solving With the {MATH} Dataset},
author={Dan Hendrycks and Collin Burns and Saurav Kadavath and Akul Arora and Steven Basart and Eric Tang and Dawn Song and Jacob Steinhardt},
booktitle={Thirty-fifth Conference on Neural Information Processing Systems Datasets and Benchmarks Track (Round 2)},
year={2021},
url={https://openreview.net/forum?id=7Bywt2mQsCe}
}

@article{lightman2023lets,
      title={Let's Verify Step by Step}, 
      author={Lightman, Hunter and Kosaraju, Vineet and Burda, Yura and Edwards, Harri and Baker, Bowen and Lee, Teddy and Leike, Jan and Schulman, John and Sutskever, Ilya and Cobbe, Karl},
      journal={arXiv preprint arXiv:2305.20050},
      year={2023}
}

@inproceedings{
    jimenez2024swebench,
    title={{SWE}-bench: Can Language Models Resolve Real-world Github Issues?},
    author={Carlos E Jimenez and John Yang and Alexander Wettig and Shunyu Yao and Kexin Pei and Ofir Press and Karthik R Narasimhan},
    booktitle={The Twelfth International Conference on Learning Representations},
    year={2024},
    url={https://openreview.net/forum?id=VTF8yNQM66}
}

@article{jain2024livecodebench,
    title={LiveCodeBench: Holistic and Contamination Free Evaluation of Large Language Models for Code},
    author={Jain, Naman and Han, King and Gu, Alex and Li, Wen-Ding and Yan, Fanjia and Zhang, Tianjun and Wang, Sida and Solar-Lezama, Armando and Sen, Koushik and Stoica, Ion},
    journal={arXiv preprint arXiv:2403.07974},
    year={2024}
}

@inproceedings{
zhuo2025bigcodebench,
title={BigCodeBench: Benchmarking Code Generation with Diverse Function Calls and Complex Instructions},
author={Terry Yue Zhuo and Vu Minh Chien and Jenny Chim and Han Hu and Wenhao Yu and Ratnadira Widyasari and Imam Nur Bani Yusuf and Haolan Zhan and Junda He and Indraneil Paul and Simon Brunner and Chen GONG and James Hoang and Armel Randy Zebaze and Xiaoheng Hong and Wen-Ding Li and Jean Kaddour and Ming Xu and Zhihan Zhang and Prateek Yadav and Naman Jain and Alex Gu and Zhoujun Cheng and Jiawei Liu and Qian Liu and Zijian Wang and David Lo and Binyuan Hui and Niklas Muennighoff and Daniel Fried and Xiaoning Du and Harm de Vries and Leandro Von Werra},
booktitle={The Thirteenth International Conference on Learning Representations},
year={2025},
url={https://openreview.net/forum?id=YrycTjllL0}
}

@article{ye2025verina,
  title={VERINA: Benchmarking Verifiable Code Generation},
  author={Ye, Zhe and Yan, Zhengxu and He, Jingxuan and Kasriel, Timothe and Yang, Kaiyu and Song, Dawn},
  journal={arXiv preprint arXiv:2505.23135},
  year={2025}
}

@misc{srivastava2024functionalbenchmarksrobustevaluation,
      title={Functional Benchmarks for Robust Evaluation of Reasoning Performance, and the Reasoning Gap}, 
      author={Saurabh Srivastava and Annarose M B and Anto P V and Shashank Menon and Ajay Sukumar and Adwaith Samod T and Alan Philipose and Stevin Prince and Sooraj Thomas},
      year={2024},
      eprint={2402.19450},
      archivePrefix={arXiv},
      primaryClass={cs.AI},
      url={https://arxiv.org/abs/2402.19450}, 
}

@article{BAUER_2015, title={English phonotactics}, volume={19}, DOI={10.1017/S1360674315000179}, number={3}, journal={English Language and Linguistics}, author={BAUER, LAURIE}, year={2015}, pages={437–475}}

@techreport{anthropic2026claude46opus,
  title       = {System Card: Claude Opus 4.6},
  author      = {{Anthropic}},
  year        = {2026},
  month       = {February},
  institution = {Anthropic},
  url         = {https://www-cdn.anthropic.com/6a5fa276ac68b9aeb0c8b6af5fa36326e0e166dd.pdf}
}

@techreport{anthropic2026claude46sonnet,
  title       = {System Card: Claude Sonnet 4.6},
  author      = {{Anthropic}},
  year        = {2026},
  month       = {February},
  institution = {Anthropic},
  url         = {https://www-cdn.anthropic.com/bbd8ef16d70b7a1665f14f306ee88b53f686aa75.pdf}
}

@inproceedings{longpre-etal-2021-entity,
    title = "Entity-Based Knowledge Conflicts in Question Answering",
    author = "Longpre, Shayne  and
      Perisetla, Kartik  and
      Chen, Anthony  and
      Ramesh, Nikhil  and
      DuBois, Chris  and
      Singh, Sameer",
    editor = "Moens, Marie-Francine  and
      Huang, Xuanjing  and
      Specia, Lucia  and
      Yih, Scott Wen-tau",
    booktitle = "Proceedings of the 2021 Conference on Empirical Methods in Natural Language Processing",
    month = nov,
    year = "2021",
    address = "Online and Punta Cana, Dominican Republic",
    publisher = "Association for Computational Linguistics",
    url = "https://aclanthology.org/2021.emnlp-main.565/",
    doi = "10.18653/v1/2021.emnlp-main.565",
    pages = "7052--7063"
}

@inproceedings{wu-etal-2024-reasoning,
    title = "Reasoning or Reciting? Exploring the Capabilities and Limitations of Language Models Through Counterfactual Tasks",
    author = {Wu, Zhaofeng  and
      Qiu, Linlu  and
      Ross, Alexis  and
      Aky{\"u}rek, Ekin  and
      Chen, Boyuan  and
      Wang, Bailin  and
      Kim, Najoung  and
      Andreas, Jacob  and
      Kim, Yoon},
    editor = "Duh, Kevin  and
      Gomez, Helena  and
      Bethard, Steven",
    booktitle = "Proceedings of the 2024 Conference of the North American Chapter of the Association for Computational Linguistics: Human Language Technologies (Volume 1: Long Papers)",
    month = jun,
    year = "2024",
    address = "Mexico City, Mexico",
    publisher = "Association for Computational Linguistics",
    url = "https://aclanthology.org/2024.naacl-long.102/",
    doi = "10.18653/v1/2024.naacl-long.102",
    pages = "1819--1862"
}

@inproceedings{zellers-etal-2019-hellaswag,
    title = "{H}ella{S}wag: Can a Machine Really Finish Your Sentence?",
    author = "Zellers, Rowan  and
      Holtzman, Ari  and
      Bisk, Yonatan  and
      Farhadi, Ali  and
      Choi, Yejin",
    editor = "Korhonen, Anna  and
      Traum, David  and
      M{\`a}rquez, Llu{\'i}s",
    booktitle = "Proceedings of the 57th Annual Meeting of the Association for Computational Linguistics",
    month = jul,
    year = "2019",
    address = "Florence, Italy",
    publisher = "Association for Computational Linguistics",
    url = "https://aclanthology.org/P19-1472/",
    doi = "10.18653/v1/P19-1472",
    pages = "4791--4800"
}

@inproceedings{talmor-etal-2019-commonsenseqa,
    title = "{C}ommonsense{QA}: A Question Answering Challenge Targeting Commonsense Knowledge",
    author = "Talmor, Alon  and
      Herzig, Jonathan  and
      Lourie, Nicholas  and
      Berant, Jonathan",
    editor = "Burstein, Jill  and
      Doran, Christy  and
      Solorio, Thamar",
    booktitle = "Proceedings of the 2019 Conference of the North {A}merican Chapter of the Association for Computational Linguistics: Human Language Technologies, Volume 1 (Long and Short Papers)",
    month = jun,
    year = "2019",
    address = "Minneapolis, Minnesota",
    publisher = "Association for Computational Linguistics",
    url = "https://aclanthology.org/N19-1421/",
    doi = "10.18653/v1/N19-1421",
    pages = "4149--4158"
}

@inproceedings{yang-etal-2018-hotpotqa,
    title = "{H}otpot{QA}: A Dataset for Diverse, Explainable Multi-hop Question Answering",
    author = "Yang, Zhilin  and
      Qi, Peng  and
      Zhang, Saizheng  and
      Bengio, Yoshua  and
      Cohen, William  and
      Salakhutdinov, Ruslan  and
      Manning, Christopher D.",
    editor = "Riloff, Ellen  and
      Chiang, David  and
      Hockenmaier, Julia  and
      Tsujii, Jun{'}ichi",
    booktitle = "Proceedings of the 2018 Conference on Empirical Methods in Natural Language Processing",
    month = oct # "-" # nov,
    year = "2018",
    address = "Brussels, Belgium",
    publisher = "Association for Computational Linguistics",
    url = "https://aclanthology.org/D18-1259/",
    doi = "10.18653/v1/D18-1259",
    pages = "2369--2380"
}

@article{10.1098/rsta.2023.0254,
    author = {Katz, Daniel Martin and Bommarito, Michael James and Gao, Shang and Arredondo, Pablo},
    title = {GPT-4 passes the bar exam},
    journal = {Philosophical Transactions of the Royal Society A: Mathematical, Physical and Engineering Sciences},
    volume = {382},
    number = {2270},
    pages = {20230254},
    year = {2024},
    month = {02},
    issn = {1364-503X},
    doi = {10.1098/rsta.2023.0254},
    url = {https://doi.org/10.1098/rsta.2023.0254},
    eprint = {https://royalsocietypublishing.org/rsta/article-pdf/doi/10.1098/rsta.2023.0254/1328474/rsta.2023.0254.pdf},
}

@misc{nori2023capabilitiesgpt4medicalchallenge,
      title={Capabilities of GPT-4 on Medical Challenge Problems}, 
      author={Harsha Nori and Nicholas King and Scott Mayer McKinney and Dean Carignan and Eric Horvitz},
      year={2023},
      eprint={2303.13375},
      archivePrefix={arXiv},
      primaryClass={cs.CL},
      url={https://arxiv.org/abs/2303.13375}, 
}

@inproceedings{alshammari2026mathnet,
  title     = {MathNet: A Global Multimodal Benchmark for Mathematical
               Reasoning and Retrieval},
  author    = {Alshammari, Shaden and Wen, Kevin and Zainal, Abrar and
               Hamilton, Mark and Safaei, Navid and Albarakati, Sultan and
               Freeman, William T. and Torralba, Antonio},
  booktitle = {International Conference on Learning Representations},
  year      = {2026},
  url       = {https://mathnet.mit.edu}
}

@inproceedings{
he2026martingale,
title={Martingale Score: An Unsupervised Metric for Bayesian Rationality in {LLM} Reasoning},
author={Zhonghao He and Tianyi Qiu and Hirokazu Shirado and Maarten Sap},
booktitle={The Thirty-ninth Annual Conference on Neural Information Processing Systems},
year={2026},
url={https://openreview.net/forum?id=BfO6od6JD6}
}

@misc{jain2025consensusmitigatingagreeablenessbias,
      title={Beyond Consensus: Mitigating the Agreeableness Bias in LLM Judge Evaluations}, 
      author={Suryaansh Jain and Umair Z. Ahmed and Shubham Sahai and Ben Leong},
      year={2025},
      eprint={2510.11822},
      archivePrefix={arXiv},
      primaryClass={cs.AI},
      url={https://arxiv.org/abs/2510.11822}, 
}

@article{Virtanen_2020,
   title={SciPy 1.0: fundamental algorithms for scientific computing in Python},
   volume={17},
   ISSN={1548-7105},
   url={http://dx.doi.org/10.1038/s41592-019-0686-2},
   DOI={10.1038/s41592-019-0686-2},
   number={3},
   journal={Nature Methods},
   publisher={Springer Science and Business Media LLC},
   author={Virtanen, Pauli and Gommers, Ralf and Oliphant, Travis E. and Haberland, Matt and Reddy, Tyler and Cournapeau, David and Burovski, Evgeni and Peterson, Pearu and Weckesser, Warren and Bright, Jonathan and van der Walt, Stéfan J. and Brett, Matthew and Wilson, Joshua and Millman, K. Jarrod and Mayorov, Nikolay and Nelson, Andrew R. J. and Jones, Eric and Kern, Robert and Larson, Eric and Carey, C J and Polat, İlhan and Feng, Yu and Moore, Eric W. and VanderPlas, Jake and Laxalde, Denis and Perktold, Josef and Cimrman, Robert and Henriksen, Ian and Quintero, E. A. and Harris, Charles R. and Archibald, Anne M. and Ribeiro, Antônio H. and Pedregosa, Fabian and van Mulbregt, Paul and Vijaykumar, Aditya and Bardelli, Alessandro Pietro and Rothberg, Alex and Hilboll, Andreas and Kloeckner, Andreas and Scopatz, Anthony and Lee, Antony and Rokem, Ariel and Woods, C. Nathan and Fulton, Chad and Masson, Charles and Häggström, Christian and Fitzgerald, Clark and Nicholson, David A. and Hagen, David R. and Pasechnik, Dmitrii V. and Olivetti, Emanuele and Martin, Eric and Wieser, Eric and Silva, Fabrice and Lenders, Felix and Wilhelm, Florian and Young, G. and Price, Gavin A. and Ingold, Gert-Ludwig and Allen, Gregory E. and Lee, Gregory R. and Audren, Hervé and Probst, Irvin and Dietrich, Jörg P. and Silterra, Jacob and Webber, James T and Slavič, Janko and Nothman, Joel and Buchner, Johannes and Kulick, Johannes and Schönberger, Johannes L. and de Miranda Cardoso, José Vinícius and Reimer, Joscha and Harrington, Joseph and Rodríguez, Juan Luis Cano and Nunez-Iglesias, Juan and Kuczynski, Justin and Tritz, Kevin and Thoma, Martin and Newville, Matthew and Kümmerer, Matthias and Bolingbroke, Maximilian and Tartre, Michael and Pak, Mikhail and Smith, Nathaniel J. and Nowaczyk, Nikolai and Shebanov, Nikolay and Pavlyk, Oleksandr and Brodtkorb, Per A. and Lee, Perry and McGibbon, Robert T. and Feldbauer, Roman and Lewis, Sam and Tygier, Sam and Sievert, Scott and Vigna, Sebastiano and Peterson, Stefan and More, Surhud and Pudlik, Tadeusz and Oshima, Takuya and Pingel, Thomas J. and Robitaille, Thomas P. and Spura, Thomas and Jones, Thouis R. and Cera, Tim and Leslie, Tim and Zito, Tiziano and Krauss, Tom and Upadhyay, Utkarsh and Halchenko, Yaroslav O. and Vázquez-Baeza, Yoshiki},
   year={2020},
   month=Feb, pages={261–272} }

@misc{olmo2026olmo3,
      title={Olmo 3}, 
      author={Team Olmo and : and Allyson Ettinger and Amanda Bertsch and Bailey Kuehl and David Graham and David Heineman and Dirk Groeneveld and Faeze Brahman and Finbarr Timbers and Hamish Ivison and Jacob Morrison and Jake Poznanski and Kyle Lo and Luca Soldaini and Matt Jordan and Mayee Chen and Michael Noukhovitch and Nathan Lambert and Pete Walsh and Pradeep Dasigi and Robert Berry and Saumya Malik and Saurabh Shah and Scott Geng and Shane Arora and Shashank Gupta and Taira Anderson and Teng Xiao and Tyler Murray and Tyler Romero and Victoria Graf and Akari Asai and Akshita Bhagia and Alexander Wettig and Alisa Liu and Aman Rangapur and Chloe Anastasiades and Costa Huang and Dustin Schwenk and Harsh Trivedi and Ian Magnusson and Jaron Lochner and Jiacheng Liu and Lester James V. Miranda and Maarten Sap and Malia Morgan and Michael Schmitz and Michal Guerquin and Michael Wilson and Regan Huff and Ronan Le Bras and Rui Xin and Rulin Shao and Sam Skjonsberg and Shannon Zejiang Shen and Shuyue Stella Li and Tucker Wilde and Valentina Pyatkin and Will Merrill and Yapei Chang and Yuling Gu and Zhiyuan Zeng and Ashish Sabharwal and Luke Zettlemoyer and Pang Wei Koh and Ali Farhadi and Noah A. Smith and Hannaneh Hajishirzi},
      year={2026},
      eprint={2512.13961},
      archivePrefix={arXiv},
      primaryClass={cs.CL},
      url={https://arxiv.org/abs/2512.13961}, 
}

@misc{openai2025gptoss120bgptoss20bmodel,
      title={gpt-oss-120b \& gpt-oss-20b Model Card}, 
      author={OpenAI and : and Sandhini Agarwal and Lama Ahmad and Jason Ai and Sam Altman and Andy Applebaum and Edwin Arbus and Rahul K. Arora and Yu Bai and Bowen Baker and Haiming Bao and Boaz Barak and Ally Bennett and Tyler Bertao and Nivedita Brett and Eugene Brevdo and Greg Brockman and Sebastien Bubeck and Che Chang and Kai Chen and Mark Chen and Enoch Cheung and Aidan Clark and Dan Cook and Marat Dukhan and Casey Dvorak and Kevin Fives and Vlad Fomenko and Timur Garipov and Kristian Georgiev and Mia Glaese and Tarun Gogineni and Adam Goucher and Lukas Gross and Katia Gil Guzman and John Hallman and Jackie Hehir and Johannes Heidecke and Alec Helyar and Haitang Hu and Romain Huet and Jacob Huh and Saachi Jain and Zach Johnson and Chris Koch and Irina Kofman and Dominik Kundel and Jason Kwon and Volodymyr Kyrylov and Elaine Ya Le and Guillaume Leclerc and James Park Lennon and Scott Lessans and Mario Lezcano-Casado and Yuanzhi Li and Zhuohan Li and Ji Lin and Jordan Liss and Lily and Liu and Jiancheng Liu and Kevin Lu and Chris Lu and Zoran Martinovic and Lindsay McCallum and Josh McGrath and Scott McKinney and Aidan McLaughlin and Song Mei and Steve Mostovoy and Tong Mu and Gideon Myles and Alexander Neitz and Alex Nichol and Jakub Pachocki and Alex Paino and Dana Palmie and Ashley Pantuliano and Giambattista Parascandolo and Jongsoo Park and Leher Pathak and Carolina Paz and Ludovic Peran and Dmitry Pimenov and Michelle Pokrass and Elizabeth Proehl and Huida Qiu and Gaby Raila and Filippo Raso and Hongyu Ren and Kimmy Richardson and David Robinson and Bob Rotsted and Hadi Salman and Suvansh Sanjeev and Max Schwarzer and D. Sculley and Harshit Sikchi and Kendal Simon and Karan Singhal and Yang Song and Dane Stuckey and Zhiqing Sun and Philippe Tillet and Sam Toizer and Foivos Tsimpourlas and Nikhil Vyas and Eric Wallace and Xin Wang and Miles Wang and Olivia Watkins and Kevin Weil and Amy Wendling and Kevin Whinnery and Cedric Whitney and Hannah Wong and Lin Yang and Yu Yang and Michihiro Yasunaga and Kristen Ying and Wojciech Zaremba and Wenting Zhan and Cyril Zhang and Brian Zhang and Eddie Zhang and Shengjia Zhao},
      year={2025},
      eprint={2508.10925},
      archivePrefix={arXiv},
      primaryClass={cs.CL},
      url={https://arxiv.org/abs/2508.10925}, 
}

@misc{qwen35blog,
    title = {Qwen3.5: Accelerating Productivity with Native Multimodal Agents},
    url = {https://qwen.ai/blog?id=qwen3.5},
    author = {Qwen},
    month = {February},
    year = {2026}
}

@article{dubey2024llama3,
    title = {The Llama 3 Herd of Models},
    author = {Dubey, Abhimanyu and others},
    journal = {arXiv preprint arXiv:2407.21783},
    year = {2024},
    url = {https://arxiv.org/abs/2407.21783}
}
    \bibliographystyle{iclr2027_conference}
}

\appendix

\section*{Appendix}

\section{Related Work}
\label{appendix:related_work}

\subsection{Evaluating Reasoning via Interventions and Robustness}
\label{appendix:related_interventions}

Several studies investigate the reasoning limits of LLMs by applying controlled interventions to the input text. Prior work perturbs numerical values or alters problem constraints in math and coding datasets to test whether models solve the underlying operation or exploit surface regularities \citep{stolfo-etal-2023-causal, hong-etal-2025-evaluating, mirzadeh2025gsmsymbolic}. Similarly, \citet{shrestha2025mathematicalreasoninglargelanguage} show that logical accuracy degrades when numerical entities are scaled outside standard ranges, and \citet{razeghi2022impactpretrainingtermfrequencies} link arithmetic performance to term frequencies in the training data.

These studies show that LLM reasoning can be brittle to input variations. Our objective is narrower: we do not primarily test robustness to arbitrary problem variation, but the dependence of contextual reasoning on parametric familiarity. By swapping factual entities for \fictitiousterm{} ones while preserving the template, the answer expression, and the replacement constraints, we isolate the difference between operating with familiar versus unfamiliar entity anchors.

\subsection{\texorpdfstring{\Fictitiousterm{}}{Fictitious} and Counterfactual Benchmarks}
\label{appendix:related_counterfactual}

To mitigate data contamination and force models to rely on provided context, recent works introduce \fictitiousterm{} documents or counterfactual scenarios \citep{monteiro2024repliqaquestionansweringdatasetbenchmarking, wu2024cofcastepwisecounterfactualmultihop, glockner-etal-2025-neoqa, modarressi-etal-2024-consistent, wu-etal-2024-reasoning}. For instance, NeoQA investigates whether LLMs fall back on parametric knowledge when retrieved documents lack sufficient information \citep{glockner-etal-2025-neoqa}. CofCA provides a step-wise counterfactual multi-hop QA benchmark, making it one of the closest points of comparison for document-level counterfactual reasoning \citep{wu2024cofcastepwisecounterfactualmultihop}.

Previous counterfactual benchmarks nevertheless face a methodological limitation around task parity. Some compare a \fictitiousterm{} dataset against a different factual benchmark, making it difficult to attribute a performance gap strictly to the lack of parametric knowledge rather than to differences in dataset difficulty \citep{monteiro2024repliqaquestionansweringdatasetbenchmarking}. Others regenerate counterfactual passages, which can alter sentence structure, evidence placement, and reasoning path. \textbf{\benchmarkname{}} is designed to remove that confound: the factual and \fictitiousterm{} settings are paired through the same annotated templates, so the document structure and question logic are held fixed.

\revision{The qualitative observation of lower performance on fictitious examples is consistent with \citet{wu-etal-2024-reasoning}. The paired design changes what can be concluded: because their factual and fictitious conditions use different evidence-question pairs, the gap may also reflect uncontrolled task differences, and failed answers in the fictitious condition have no paired factual answer against which to test direct recall. MemoReason holds the task fixed, allowing the gap to be attributed to entity familiarity within this controlled intervention and enabling the Shortcut Rate analysis.}

\subsection{Knowledge-Source Selection and Conflict Mechanisms}

Complementary work investigates how models select between parametric and contextual knowledge when the two conflict. \citet{tighidet-etal-2024-probing} show that internal activations can predict which knowledge source a model uses under controlled knowledge conflicts. \citet{tighidet-etal-2025-context} identify a role for entropy neurons in suppressing context copying and show that ablating these neurons changes model behavior under conflicting information. These mechanistic studies complement MemoReason's behavioral evaluation: our paired templates enable controlled tests of how entity familiarity and knowledge conflict affect contextual reasoning, without attributing the observed performance differences to a specific internal mechanism.

\section{Taxonomy \& Annotation}
\label{appendix:taxonomy_annotation}
\label{appendix:taxonomy}

\subsection{Entity Types}
\label{appendix:entity_types}

\begin{figure}[!htbp]
    \centering
    \ifdefined\CLEANVERSION
      \includegraphics[width=\linewidth]{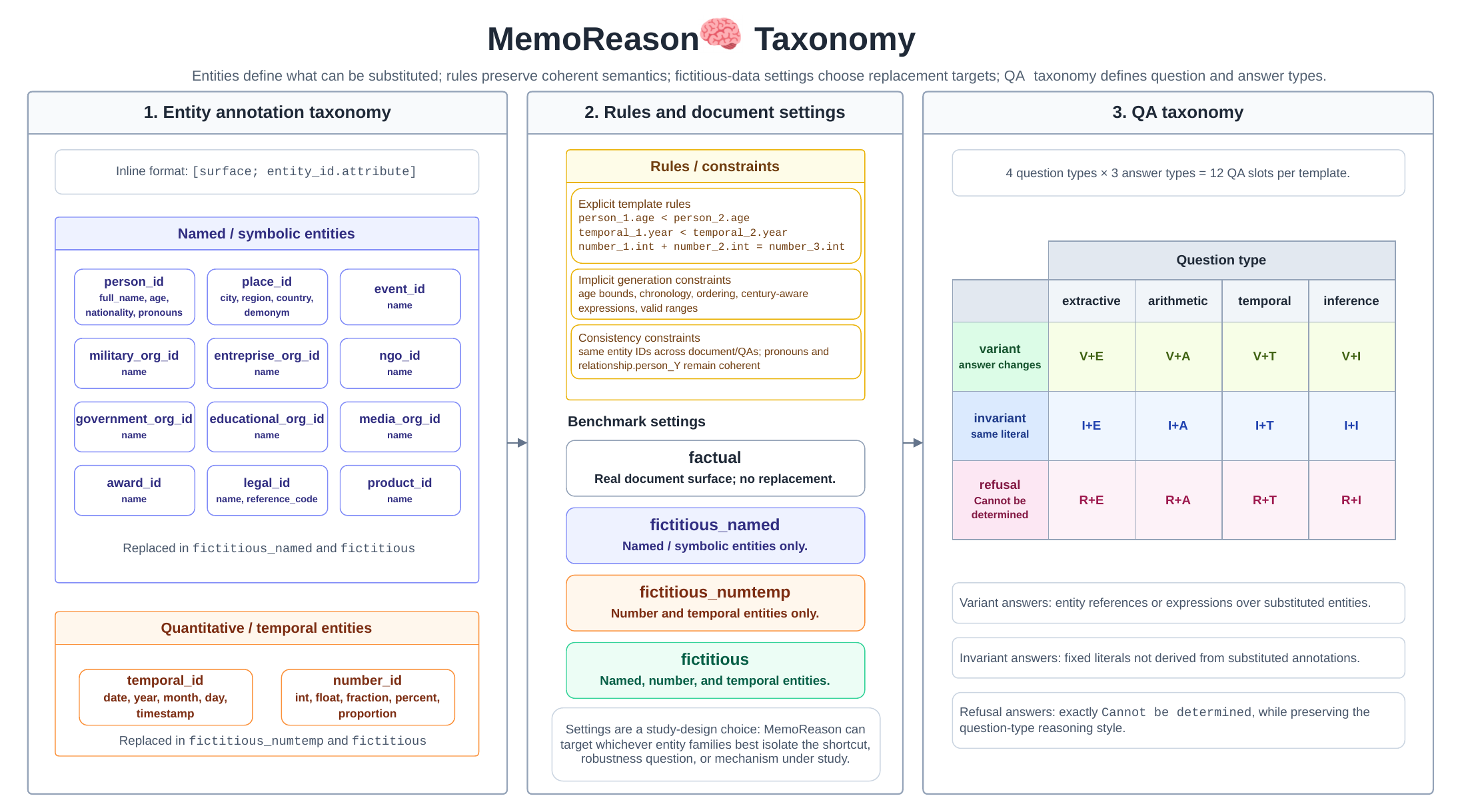}
    \else
      \includegraphics[width=\linewidth]{figures/memoreason_taxonomy/memoreason_taxonomy_fictitious.pdf}
    \fi
    \caption{Overview of the \textbf{\benchmarkname} taxonomy and benchmark construction layers. The taxonomy separates semantically distinct entity families, the rule layer preserves document consistency under replacement, and the QA contract crosses question type with answer type.}
    \label{fig:memoreason_taxonomy}
\end{figure}

The taxonomy defines 14 entity types used to annotate all replaceable spans in the source documents: persons, places, events, military organizations, enterprise organizations, NGOs, government organizations, educational organizations, media organizations, temporals, numbers, awards, legal instruments, and products. Figure~\ref{fig:memoreason_taxonomy} summarizes how these entity families connect to the rule layer and the question-answer contract used by \benchmarkname{}.

The main reason for introducing this taxonomy is experimental control. \benchmarkname{} is not only a \revision{benchmark based on fictitious entity replacement}: it is designed to replace the familiar anchors that models may have memorized while keeping the document structure, question, answer expression, and reasoning path fixed. For this to work, the replacement operation must know which spans can be swapped independently and which spans must remain coupled. Each entity type therefore has a compact set of attributes that determine what can be replaced and what must stay consistent across mentions. For example, a person entity can expose names, age, gendered forms, nationality, and relationship attributes, while a legal entity can expose both its name and reference code.

We chose a medium-grained taxonomy rather than a single generic entity class or a very fine-grained ontology. A generic class would make replacements syntactically easy but semantically unsafe: replacing an educational organization with a media outlet, or a legal instrument with a product, can preserve surface fluency while breaking the factual role played by the entity in the document. Conversely, an overly fine-grained ontology would make annotation brittle and would create many rare categories that are difficult to replace reliably. The selected types reflect the distinctions that most often affect document validity under replacement: named entities with different social or institutional roles, quantitative and temporal entities governed by constraints, and domain-specific referents such as awards, products, and legal instruments.

\subsection{Rules}
\label{appendix:rules}

Rules encode document-specific constraints that must remain true \revision{after entity replacement}. They include arithmetic relations, compatibility constraints, plural constraints, exact offsets, and value couplings that are required by the text. Generic ordering constraints are handled automatically by the generation pipeline when they preserve only the factual order of numbers, dates, years, or ages. The goal is to avoid brittle \fictitiousterm{} variants where the surface replacement is syntactically valid but the document becomes semantically incoherent.

\subsection{Eliminating Unique Factual References}
\label{appendix:unique_factual_references}

Some factual documents contain descriptions that uniquely identify the original entity even after direct entity replacement. During annotation, these spans are either kept literal when they are structurally necessary or softened when they would reintroduce a unique factual cue. For example, a formulation such as ``the greatest sprinter'' can be broadened to avoid forcing the model to reconcile a \fictitiousterm{} name with a world-knowledge fact about a uniquely identifiable person. This step helps ensure that the \fictitiousterm{} setting tests contextual reasoning rather than contradiction handling.

\section{Judge Match Calibration}
\label{sec:judge_calibration}

We model the Judge Match (JM) as a \textit{binary estimator}: given a question, a ground truth answer, and a model prediction as input, it outputs a binary success/failure assessment with respect to the semantic equivalence requirement. Such an estimator is characterized by two recall parameters: its \textbf{positive recall} $r_+$, i.e., the probability that it correctly identifies a semantically equivalent prediction, and its \textbf{negative recall} $r_-$, i.e., the probability that it correctly identifies a non-equivalent one.

\paragraph{Calibration.}
\label{appendix:judge_calibration_protocol}
To estimate $r_+$ and $r_-$, we randomly sampled 200 examples from the training split and had a human annotator label each (question, ground truth, prediction) triple as either a match or a non-match. We then applied JM to the same examples and treated the human annotations as ground truth, yielding $r_+ = 1.0$ and $r_- = 1.0$ (100\% agreement). \postreviewrevision{For \unchangednumber{200} successes in \unchangednumber{200} trials, the two-sided \unchangednumber{95\%} lower bound is \postreviewnumber{98.2\%} for each recall, so perfect observed agreement should not be read as zero calibration uncertainty.}

\paragraph{Corrected performance estimation.}
\label{appendix:judge_corrected_estimation}
When applying JM to $n$ executions, the raw success rate $q$ (i.e., the fraction of predictions labeled as matches) is a biased estimate of the true performance $p$. Given the calibrated recalls, we correct for this bias and report $p$ together with a confidence interval using the following formula:
\begin{equation}
    p \in \frac{q + r_{-} - 1}{r_{+} + r_{-} - 1} \pm \Delta_p, \qquad
    \Delta_p = z\,\frac{\sqrt{\frac{q(1-q)}{n}}}{r_{+} + r_{-} - 1}
\end{equation}
where $z$ is the quantile of the standard normal distribution corresponding to the desired confidence level (e.g., $z = 1.96$ for a 95\% confidence interval). When $r_+ = r_- = 1$, this reduces to $p = q$, confirming that no correction is needed in our case.

\section{Annotation Interface}
\label{appendix:annotation_interface}

We built a custom annotation interface to support the human review loop. The interface is organized around three coupled tasks: document annotation, rule annotation, and question-answer validation. Keeping these tasks in one interface makes it possible for annotators to inspect whether a proposed entity span, rule, or answer expression remains valid under the same template.

\revision{Final quality control is layered rather than delegated to a single automatic score. Each template was independently reviewed by two domain-knowledgeable annotators (and by three for a subset), disagreements were adjudicated, and corrections from audit passes were cross-checked by two additional readers together with a sample of unflagged items. One such audit identified \unchangednumber{15} subtly problematic questions, approximately \unchangednumber{1.4\%} of the then-current question set; all were corrected and independently reviewed before release.}

\subsection{Document Annotation}
\label{appendix:document_annotation}

Annotators review inline entity annotations over the source document, question, and answer expression. They verify the type of entity, the attribute, and the consistency of repeated references throughout the document.

\subsection{Rules Annotation}
\label{appendix:rules_annotation}

Annotators then inspect the rule list attached to the template. The goal is not to encode every factual relation, but to encode only the constraints needed for \fictitiousterm{} replacements to preserve the document logic.

\subsection{Questions/Answers Annotation}
\label{appendix:qa_annotation}

Each document is paired with 12 question-answer slots, crossing four question types with three answer types. The interface helps annotators verify that each question is answerable from the document when it should be, that refusal questions genuinely lack supporting evidence, and that the answer expression can be evaluated after \fictitiousterm{} replacement.

\section{Dataset Details}
\label{appendix:dataset_details}

\subsection{Dataset Themes}
\label{sec:dataset-themes}

The factual dataset is organized into nine thematic categories. These themes are not different tasks: all of them are annotated with the same entity taxonomy, replacement rules, question types, and answer types. Their role is to diversify the factual contexts in which we test whether models rely on document-grounded reasoning or parametric shortcuts.

\begin{itemize}[leftmargin=*,itemsep=3pt]
    \item \textbf{Award Winners} contains articles about prominent public figures whose documents include major awards, distinctions, honors, or record-setting achievements. This theme is dense in person, award, organization, event, and temporal entities.

    \item \textbf{Biographies} contains articles about famous personalities, especially scientists, politicians, and institutional leaders. These documents emphasize career trajectories, appointments, offices, affiliations, and chronological progressions across institutions.

    \item \textbf{Places} contains articles about cities, countries, and broader regions. These documents cover geographic, political, demographic, historical, and administrative facts, making them useful for testing place-based relations and numerical comparisons.

    \item \textbf{Companies} contains articles about corporations and organizations. The documents describe founding histories, mergers, acquisitions, subsidiaries, sectors, headquarters, leadership, and market-related facts.

    \item \textbf{Natural Disasters} contains articles about earthquakes, hurricanes, tsunamis, wildfires, and other large-scale disasters. These documents involve event timelines, affected regions, casualties, magnitudes, damages, and institutional responses.

    \item \textbf{Public Attacks} contains WikiEvent-derived news articles about public attacks, threats, and violent incidents.

    \item \textbf{Retail Banking} contains articles about banking regulations, payment systems, deposit guarantees, capital requirements, and resolution mechanisms. This theme introduces legal and institutional language, with reasoning often depending on policy roles, requirements, and timelines.

    \item \textbf{Space Missions} contains articles about spaceflight programs, spacecraft, missions, and space agencies. These documents combine technical mission descriptions with crews, launch dates, mission outcomes, vehicles, and chronological dependencies.

    \item \textbf{Sport Events} contains articles about major competitions and tournaments. These documents include editions, participants, venues, records, audiences, rankings, and event histories, which provide many numerical, temporal, and relational reasoning cases.
\end{itemize}

\subsection{Released Datasets}
\label{appendix:released_datasets}
\revision{We release the ten dataset settings used in the paper experiments. The factual setting is the reference point; every intervention is paired through the same document templates whenever factual and fictitious performance are compared. Public-facing aliases use the \texttt{fictitious*} prefix below; frozen experimental manifests retain their original keys for hash compatibility.}

\begin{table}[!htbp]
\revisiontablecolor
\revisioncaptionstyle
\centering
\scriptsize
\setlength{\tabcolsep}{4pt}
\begin{tabularx}{\linewidth}{@{}p{0.26\linewidth}p{0.23\linewidth}X@{}}
\toprule
Dataset setting & Experiment role & Description \\
\midrule
\texttt{factual} & Reference setting & Original factual documents and questions after human template review. \\
\fictitiousidentifier{} & Main replacement setting & Full replacement of named, numerical, and temporal entities under the template rules. \\
\fictitiousidentifier{\_named} & \revision{Replacement ablation} & \revision{Replaces named entities while leaving numerical and temporal values unchanged.} \\
\fictitiousidentifier{\_numtemp} & \revision{Replacement ablation} & \revision{Replaces numerical and temporal entities while leaving named entities unchanged.} \\
\fictitiousidentifier{\_10pct} & Partial replacement & Replaces 10\% of eligible entities. \\
\fictitiousidentifier{\_20pct} & Partial replacement & Replaces 20\% of eligible entities. \\
\fictitiousidentifier{\_30pct} & Partial replacement & Replaces 30\% of eligible entities. \\
\fictitiousidentifier{\_50pct} & Partial replacement & Replaces 50\% of eligible entities. \\
\fictitiousidentifier{\_80pct} & Partial replacement & Replaces 80\% of eligible entities. \\
\fictitiousidentifier{\_90pct} & Partial replacement & Replaces 90\% of eligible entities. \\
\bottomrule
\end{tabularx}
\caption{Released dataset settings used in the experiments. Partial-replacement datasets are generated from the same templates as the factual and fully \fictitiousterm{} settings.}
\label{tab:released_datasets}
\end{table}

\section{Experimental Setup Details}
\label{appendix:experimental_setup}

This section reports the inference and scoring settings used for the experiments. Each question is evaluated independently: the model receives the document and a single question, and must return exactly one line of the form \texttt{ANSWER: <answer>}. The system prompt is shown in Appendix~\ref{appendix:evaluation_system_prompt}. We do not request chain-of-thought rationales; when a reasoning-capable model emits hidden or tagged reasoning, the evaluation pipeline stores it separately and parses only the final answer channel or final \texttt{ANSWER:} line.

\subsection{Model Inference Settings}
\label{appendix:model_inference_settings}

\revision{In the main experiments, decoding is deterministic (i.e. greedy), and reasoning-related controls are held fixed:} \texttt{GPT-OSS} API requests use \texttt{reasoning\_effort=low} and do not request reasoning traces, while local \texttt{GPT-OSS} runs use the corresponding low-reasoning setting in the chat template.

\subsection{Hardware}
\label{appendix:compute_resources}
Experiments were performed using NVIDIA H100 and A100 GPUs with 80~GB of VRAM.
Generating the local open-weight model outputs, including ablations, required approximately 200--250 GPU hours. 
This estimate excludes provider-side compute for API-hosted models.

\subsection{Model Licenses}
\label{appendix:models_licenses}

The open-weight models evaluated in this paper are used under the licenses stated by their providers. \texttt{GPT-OSS-20B} and \texttt{GPT-OSS-120B} are released under the Apache 2.0 license \citep{openai2025gptoss120bgptoss20bmodel}. \texttt{Qwen3.5-27B} and \texttt{Qwen3.5-35B-A3B} are released under the Apache 2.0 license \citep{qwen35blog}. The OLMo-family models, including \texttt{OLMo-3-7B-Instruct}, \texttt{OLMo-3-7B-Think}, and \texttt{OLMo-2-32B-Instruct}, are released under the Apache 2.0 license \citep{olmo2026olmo3}. \texttt{Gemma-4-26B-A4B-it} is released under the Gemma license terms available at \url{https://ai.google.dev/gemma/apache_2}. \texttt{\unchangednumber{Llama-3.1-8B-Instruct}} is used under the Llama \unchangednumber{3.1} Community License available from Meta at \url{https://www.llama.com/llama3_1/license/}. For API-only systems, no model weights are redistributed by this work: \texttt{Claude Sonnet 4.6} and \texttt{Opus 4.6} are governed by Anthropic's service terms \citep{anthropic2026claude46sonnet, anthropic2026claude46opus}.

\section{Full Results}
\label{appendix:additional_results}

\subsection{Robustness and Mechanism Checks}
\label{appendix:robustness_checks}

\revision{The factual--fictitious accuracy gap establishes that replacing familiar content affects performance, but does not explain why. We therefore examine several possible explanations. Does performance depend primarily on familiar names or on numerical and temporal information? Which previously correct answers become incorrect after replacement? Could familiar names trigger the retrieval of factual answers that no longer apply? Finally, we examine whether the gap persists under stochastic decoding and whether it is associated with the frequency of named entities in training data. Confidence intervals in these analyses resample complete source documents, preserving the dependence among their questions and variants.}

\subsubsection{\revision{Which substitutions affect performance when the answer stays unchanged?}}
\label{appendix:cue_ablation}

\revision{Replacing all entities changes both names and numerical or temporal values. To distinguish their contributions, we compare four settings: the original factual documents, replacement of names only, replacement of numerical and temporal values only, and full replacement. We evaluate \unchangednumber{OLMo-3-7B-Instruct}, \unchangednumber{OLMo-3-7B-Think}, and \unchangednumber{Qwen3.5-35B-A3B} on the same \unchangednumber{400} invariant-answer questions, covering all four question types. Because the correct answer remains unchanged, this comparison tests sensitivity to changes in the context without also changing the target answer.}

\begin{table}[H]
\centering
\scriptsize
\revisiontablecolor
\revisioncaptionstyle
\renewcommand{\arraystretch}{1.08}
\setlength{\tabcolsep}{4.2pt}
\caption{Accuracy over the 400 invariant-answer questions spanning arithmetic, temporal, inference, and extractive questions. Each non-factual cell reports accuracy and, below it, the setting-minus-factual change with its 95\% CI. $^{*}$ marks an interval excluding zero.}
\label{tab:dataset100-four-setting}
\begin{tabular}{l c c c c}
\toprule
Model & Factual & Values only & Names only & Full fictitious \\
\midrule
OLMO-3 7B-INSTRUCT & 58.25 & \shortstack{57.65\\{\tiny $\Delta=-0.60$ [-3.25, +2.08]}} & \shortstack{54.47\\{\tiny $\Delta=-3.78$ [-6.73, -0.83]$^{*}$}} & \shortstack{53.65\\{\tiny $\Delta=-4.60$ [-7.72, -1.52]$^{*}$}} \\
OLMO-3 7B-THINK & 74.50 & \shortstack{73.75\\{\tiny $\Delta=-0.75$ [-4.08, +2.65]}} & \shortstack{73.17\\{\tiny $\Delta=-1.33$ [-4.92, +2.37]}} & \shortstack{71.70\\{\tiny $\Delta=-2.80$ [-6.30, +0.72]}} \\
QWEN3.5 35B-A3B & 86.50 & \shortstack{82.05\\{\tiny $\Delta=-4.45$ [-6.57, -2.43]$^{*}$}} & \shortstack{84.77\\{\tiny $\Delta=-1.73$ [-3.98, +0.50]}} & \shortstack{82.55\\{\tiny $\Delta=-3.95$ [-6.73, -1.32]$^{*}$}} \\
\bottomrule
\end{tabular}
\end{table}

\revision{Table~\ref{tab:dataset100-four-setting} shows a significant decrease of \unchangednumber{3.78} percentage points for \unchangednumber{OLMo-3-7B-Instruct} under names-only replacement, and \unchangednumber{4.45} points for \unchangednumber{Qwen3.5-35B-A3B} under values-only replacement. Neither isolated intervention produces a statistically detectable decrease for \unchangednumber{OLMo-3-7B-Think}. These results show that both kinds of substitution can affect performance. They do not establish that each model depends exclusively on one kind of information: the remaining isolated-intervention confidence intervals include zero, rather than demonstrating an absence of an effect.}

\subsubsection{\revision{How does replacement change individual answers?}}
\label{appendix:directional_flips}

\revision{Average accuracy does not show which questions a model gains or loses after replacement. We therefore pair each factual prediction with predictions on its \unchangednumber{ten} fully fictitious variants, including all question and answer classes.}

\revision{We report two conditional rates. The \textit{forward flip rate} measures how often a correct factual prediction becomes an incorrect fictitious prediction. The \textit{mirror flip rate} measures how often a correct fictitious prediction has an incorrect factual counterpart. Each rate is therefore calculated among successes in its respective setting; their difference is not the percentage of all questions lost.}

\begin{table}[H]
\centering
\footnotesize
\revisiontablecolor
\postreviewcaptionstyle
\renewcommand{\arraystretch}{1.02}
\setlength{\tabcolsep}{6.5pt}
\captionsetup{justification=raggedright,singlelinecheck=false}
\caption{Directional flips across all question and answer classes; mirror is $P(\text{factual incorrect}\mid\text{fictitious correct})$.}
\label{tab:dataset100-flip-rate-all-questions}
\begin{tabular}{l c c c}
\toprule
Model & Forward flip (\%) & Mirror flip (\%) & Forward $-$ mirror (pp) \\
\midrule
OLMO-3 7B-THINK & \scorerevision{\shortstack{16.87\\{\tiny [15.13, 18.64]}}} & \scorerevision{\shortstack{11.03\\{\tiny [9.49, 12.66]}}} & \scorerevision{\shortstack{5.84\\{\tiny [3.62, 8.03]}}} \\
OLMO-3 7B-INSTRUCT & \scorerevision{\shortstack{15.49\\{\tiny [13.83, 17.18]}}} & \scorerevision{\shortstack{11.89\\{\tiny [10.05, 13.78]}}} & \scorerevision{\shortstack{3.59\\{\tiny [1.18, 5.96]}}} \\
GPT-OSS 20B & \scorerevision{\shortstack{7.62\\{\tiny [6.32, 8.98]}}} & \scorerevision{\shortstack{4.35\\{\tiny [3.27, 5.52]}}} & \scorerevision{\shortstack{3.27\\{\tiny [1.52, 5.02]}}} \\
GEMMA-4 26B-A4B-IT & \scorerevision{\shortstack{5.75\\{\tiny [4.51, 7.07]}}} & \scorerevision{\shortstack{4.28\\{\tiny [3.26, 5.35]}}} & \scorerevision{\shortstack{1.47\\{\tiny [-0.16, 3.11]}}} \\
GPT-OSS 120B & \scorerevision{\shortstack{6.32\\{\tiny [5.03, 7.69]}}} & \scorerevision{\shortstack{2.81\\{\tiny [1.96, 3.72]}}} & \scorerevision{\shortstack{3.51\\{\tiny [1.96, 5.11]}}} \\
QWEN3.5 27B & \scorerevision{\shortstack{5.67\\{\tiny [4.45, 6.97]}}} & \shortstack{2.53\\{\tiny \scorerevision{[1.73, 3.42]}}} & \scorerevision{\shortstack{3.13\\{\tiny [1.58, 4.70]}}} \\
QWEN3.5 35B-A3B & \scorerevision{\shortstack{6.24\\{\tiny [4.95, 7.63]}}} & \scorerevision{\shortstack{3.26\\{\tiny [2.30, 4.31]}}} & \scorerevision{\shortstack{2.98\\{\tiny [1.32, 4.67]}}} \\
CLAUDE SONNET 4.6 & \scorerevision{\shortstack{5.51\\{\tiny [4.36, 6.75]}}} & \scorerevision{\shortstack{2.07\\{\tiny [1.29, 2.96]}}} & \scorerevision{\shortstack{3.45\\{\tiny [2.00, 4.91]}}} \\
LLAMA-3.1 8B-INSTRUCT & \scorerevision{\shortstack{8.78\\{\tiny [7.23, 10.43]}}} & \scorerevision{\shortstack{7.15\\{\tiny [5.68, 8.69]}}} & \scorerevision{\shortstack{1.62\\{\tiny [-0.64, 3.90]}}} \\
\bottomrule
\end{tabular}
\end{table}

\revision{The forward rate exceeds the mirror rate for all \unchangednumber{\ResultAllQuestionFlipModelCount{}} models, with a positive difference whose \unchangednumber{95\%} confidence interval excludes zero for \unchangednumber{\ResultAllQuestionFlipSignificantPositiveCount{}} models (Table~\ref{tab:dataset100-flip-rate-all-questions}). The differences for \unchangednumber{Gemma-4-26B-A4B-IT} and \unchangednumber{Llama-3.1-8B-Instruct} remain inconclusive. This provides an item-level description of the factual advantage, while showing that replacement can also turn failures into successes. It does not, by itself, identify the mechanism behind these changes.}

\subsubsection{\revision{Does retaining familiar names increase the Shortcut Rate?}}

\revision{One possible explanation for the low Shortcut Rates under full replacement is that unfamiliar names no longer trigger the retrieval of memorized facts. We test this explanation by retaining factual names while replacing numerical and temporal values. Familiar names remain available as retrieval cues, even when the information needed to answer the question has changed.}

\revision{We evaluate variant-answer questions using the Shortcut Rate, measuring how often a model's incorrect prediction reproduces the original factual answer rather than the answer required by the modified document. For each question, we compute the fraction of failed variants that reproduce that answer, then average across questions; questions with no failed variants contribute zero. Thus, the reported Shortcut Rate is question-averaged, not a pooled fraction of all failures. Table~\ref{tab:dataset100-numtemp-shortcut} reports arithmetic, temporal, and inference questions, and their aggregate.}

\begin{table}[H]
\centering
\begingroup
\centering
\footnotesize
\revisiontablecolor
\revisioncaptionstyle
\setlength{\tabcolsep}{3.2pt}
\captionsetup{hypcap=false,justification=raggedright,singlelinecheck=false}
\captionof{table}{Shortcut Rate under value-only replacement (\%; failed outputs in parentheses).}
\label{tab:dataset100-numtemp-shortcut}
\begin{tabular}{l c c c c}
\toprule
Model & Arith. & Temp. & Infer. & Reason. \\
\midrule
OLMO-3 7B-INSTRUCT & \shortstack{3.80\\{\tiny (612)}} & \shortstack{3.44\\{\tiny (547)}} & \shortstack{1.86\\{\tiny (375)}} & \shortstack{3.03\\{\tiny (1534)}} \\
OLMO-3 7B-THINK & \shortstack{2.70\\{\tiny (358)}} & \shortstack{2.92\\{\tiny (298)}} & \shortstack{0.25\\{\tiny (198)}} & \shortstack{1.95\\{\tiny (854)}} \\
QWEN3.5 35B-A3B & \shortstack{3.40\\{\tiny (328)}} & \shortstack{1.25\\{\tiny (303)}} & \shortstack{5.00\\{\tiny (194)}} & \shortstack{3.22\\{\tiny (825)}} \\
\postreviewrevision{CLAUDE SONNET \unchangednumber{4.6}} & \postreviewnumber{\shortstack{3.83\\{\tiny (139)}}} & \postreviewnumber{\shortstack{2.50\\{\tiny (70)}}} & \postreviewnumber{\shortstack{2.20\\{\tiny (98)}}} & \postreviewnumber{\shortstack{2.84\\{\tiny (307)}}} \\
\bottomrule
\end{tabular}
\endgroup

\end{table}

\revision{Across the \unchangednumber{\ResultReverseModelCount{}} evaluated models, reasoning Shortcut Rates range from \unchangednumber{\ResultReverseShortcutReasoningMin\%} to \unchangednumber{\ResultReverseShortcutReasoningMax\%} (Table~\ref{tab:dataset100-numtemp-shortcut}). The largest question-type rate is \unchangednumber{\ResultReverseShortcutAtomicMax\%}, for \unchangednumber{Qwen3.5-35B-A3B} on inference questions. Retaining familiar names therefore does not produce high Shortcut Rates under this diagnostic. Familiarity may still influence intermediate reasoning without causing the final response to reproduce the factual answer.}

\subsubsection{\revision{Is the factual advantage specific to deterministic decoding?}}
\label{appendix:temperature_robustness}

\revision{The main comparison uses deterministic decoding. To check whether the observed gap depends on this choice, we evaluate \unchangednumber{OLMo-3-7B-Instruct} and \unchangednumber{Qwen3.5-27B} at temperatures of \unchangednumber{0}, \unchangednumber{0.5}, and \unchangednumber{1.0}. We use the same factual and fully fictitious questions, averaging \unchangednumber{three} generation seeds at each nonzero temperature. We compare both the factual--fictitious gap at each temperature and its change relative to temperature \unchangednumber{0}.}

\begin{table}[H]
\begingroup
\centering
\scriptsize
\revisiontablecolor
\revisioncaptionstyle
\renewcommand{\arraystretch}{1.08}
\setlength{\tabcolsep}{3.0pt}
\captionsetup{hypcap=false,font=small,justification=raggedright,singlelinecheck=false,skip=5pt}
\caption{\textbf{Decoding-temperature robustness.} Accuracy (\%) and factual--fictitious gap (percentage points). Brackets are 95\% CIs. Nonzero temperatures average three generation seeds; $T=0$ uses the deterministic run. Bold gaps have an interval excluding zero.}
\label{tab:dataset100-temperature-robustness}
\begin{tabular*}{\linewidth}{@{\extracolsep{\fill}}l c c c c@{}}
\toprule
Model & $T$ & Factual acc. & Fictitious acc. & Gap (pp) \\
\midrule
\multirow{3}{*}{\shortstack[l]{OLMo-3\\7B-Instruct}} & 0.0 & \postreviewnumber{\shortstack{69.67\\{\tiny [67.42, 71.92]}}} & \postreviewnumber{\shortstack{66.48\\{\tiny [64.55, 68.38]}}} & \postreviewnumber{\shortstack{\textbf{3.19}\\{\tiny \textbf{[1.35, 5.02]}}}} \\
 & 0.5 & \postreviewnumber{\shortstack{69.42\\{\tiny [67.03, 71.78]}}} & \postreviewnumber{\shortstack{66.02\\{\tiny [64.15, 67.87]}}} & \postreviewnumber{\shortstack{\textbf{3.40}\\{\tiny \textbf{[1.62, 5.16]}}}} \\
 & 1.0 & \postreviewnumber{\shortstack{68.81\\{\tiny [66.42, 71.19]}}} & \postreviewnumber{\shortstack{65.33\\{\tiny [63.47, 67.16]}}} & \postreviewnumber{\shortstack{\textbf{3.48}\\{\tiny \textbf{[1.69, 5.27]}}}} \\
\midrule
\multirow{3}{*}{\shortstack[l]{Qwen3.5\\27B}} & 0.0 & \postreviewnumber{\shortstack{91.67\\{\tiny [90.17, 93.17]}}} & \postreviewnumber{\shortstack{88.78\\{\tiny [87.13, 90.38]}}} & \postreviewnumber{\shortstack{\textbf{2.88}\\{\tiny \textbf{[1.43, 4.35]}}}} \\
 & 0.5 & \postreviewnumber{\shortstack{91.61\\{\tiny [90.14, 93.06]}}} & \postreviewnumber{\shortstack{88.69\\{\tiny [87.10, 90.24]}}} & \postreviewnumber{\shortstack{\textbf{2.92}\\{\tiny \textbf{[1.52, 4.35]}}}} \\
 & 1.0 & \postreviewnumber{\shortstack{91.00\\{\tiny [89.44, 92.53]}}} & \postreviewnumber{\shortstack{88.06\\{\tiny [86.49, 89.61]}}} & \postreviewnumber{\shortstack{\textbf{2.94}\\{\tiny \textbf{[1.54, 4.37]}}}} \\
\bottomrule
\end{tabular*}
\endgroup

\end{table}

\revision{The gap remains positive in all \unchangednumber{\ResultTemperatureSignificantGapCellCount{}} model--temperature combinations, with \unchangednumber{95\%} confidence intervals excluding zero (Table~\ref{tab:dataset100-temperature-robustness}). Changes relative to temperature \unchangednumber{0} range from \unchangednumber{\ResultTemperatureGapChangeMin{}} to \unchangednumber{\ResultTemperatureGapChangeMax{}} percentage points, and all \unchangednumber{\ResultTemperatureInconclusiveContrastCount{}} corresponding confidence intervals include zero. The factual advantage therefore persists under the tested stochastic settings. This does not establish that the gap is identical across temperatures or across other decoding methods.}

\subsubsection{\revision{Do more frequent names predict a larger factual advantage?}}
\label{appendix:training_frequency}

\revision{If repeated exposure strengthens access to factual associations, documents containing frequently encountered names might show a larger performance drop when those names are replaced. We examine this prediction for \unchangednumber{OLMo-3-7B-Think} using available Infini-gram counts for named entities in its training mix.}

\begin{figure}[H]
  \centering
  \includegraphics[width=0.65\linewidth]{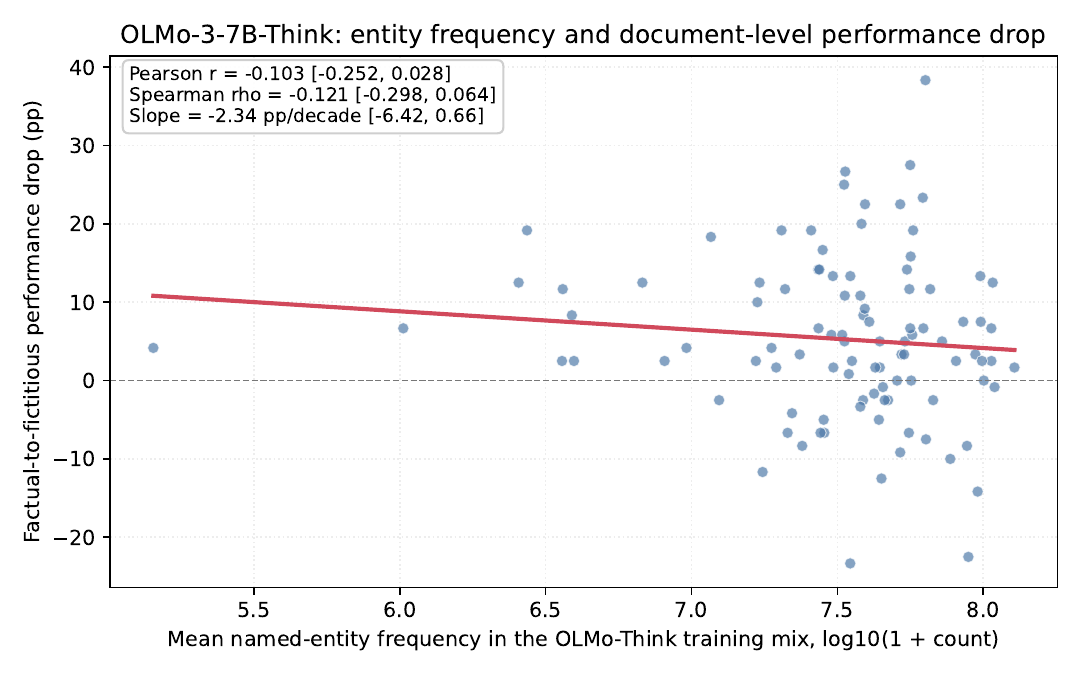}
  \caption{\revision{Named-entity frequency and factual--fictitious accuracy gap for \unchangednumber{OLMo-3-7B-Think}. The horizontal axis uses $\log_{\unchangednumber{10}}(\unchangednumber{1}+x)$, where $x$ is the mean occurrence count. The line is an OLS fit; brackets in the inset are \unchangednumber{95\%} CIs.}}
  \label{fig:training_frequency_drop}
\end{figure}

\revision{For each of the \unchangednumber{100} source documents, we average the occurrence counts of its replaced named entities. We relate $\log_{\unchangednumber{10}}(\unchangednumber{1}+\text{mean count})$ to the document's factual--fictitious accuracy gap, computed over its \unchangednumber{12} questions and \unchangednumber{ten} fictitious variants (Figure~\ref{fig:training_frequency_drop}). This measures an association with a proxy for training exposure, not with memorized knowledge directly.}

\revision{The estimated association is weakly negative: Pearson $r=\unchangednumber{\ResultFrequencyPearson}$, with a \unchangednumber{95\%} confidence interval of \unchangednumber{$[\ResultFrequencyPearsonLow,\ResultFrequencyPearsonHigh]$}. Spearman correlation and the fitted slope likewise have intervals containing zero. We therefore find no clear relationship between this document-level frequency proxy and the performance drop. This result does not rule out an influence of training exposure; it shows that average name frequency does not provide a clear predictor in this analysis.}

\clearpage
\section{Annotation Interface}
\label{sec:annotation_interface}

\begin{figure}[H]
    \centering
    \includegraphics[width=\linewidth]{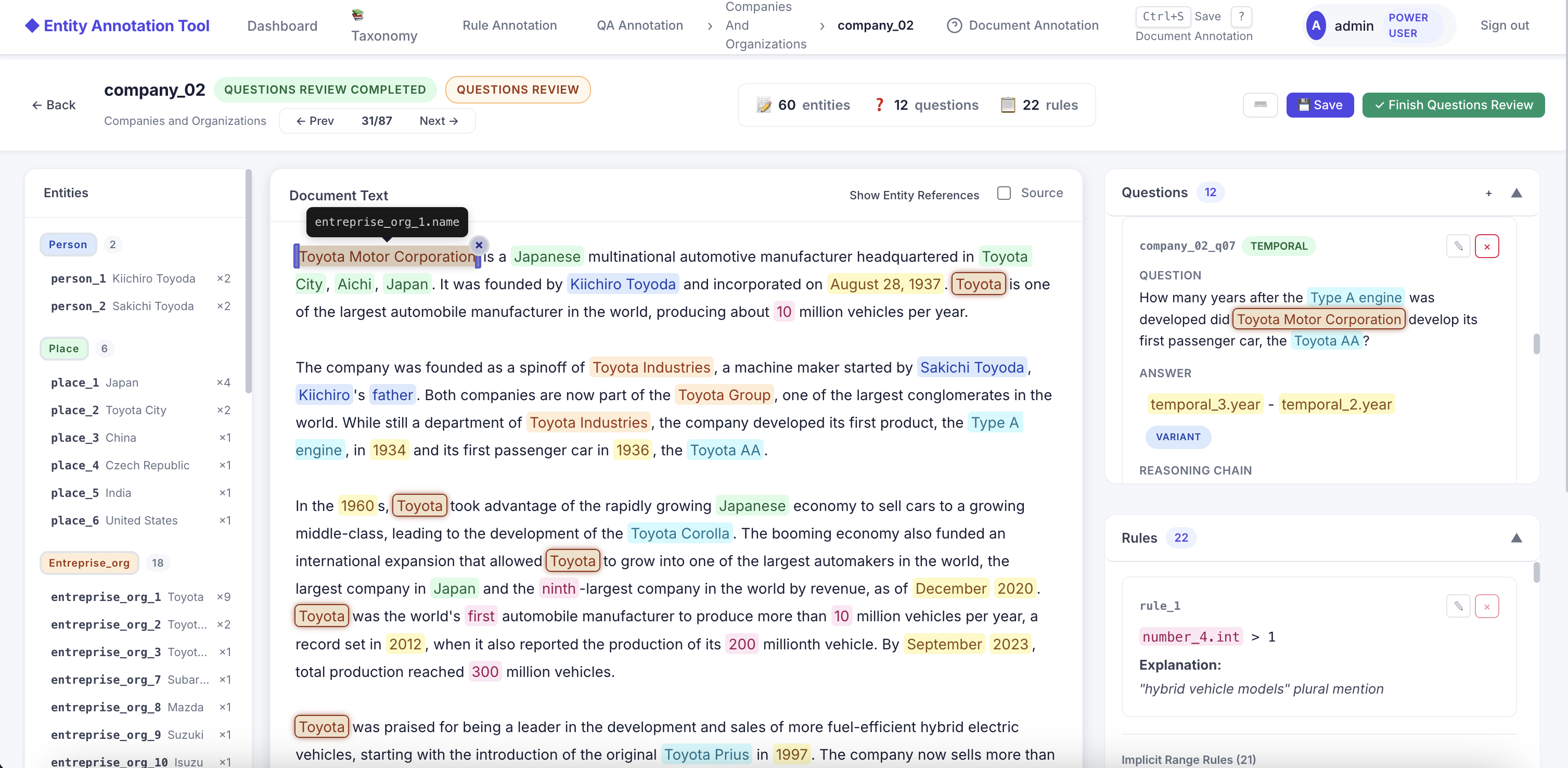}
    \caption{Screenshot of the annotation interface used to inspect entity spans, replacement rules, and question-answer fields for the \texttt{Toyota} template.}
    \label{fig:annotation_interface}
\end{figure}

\revision{We specifically developed a web interface customized for the annotation tasks of MemoReason and make the code to run it publicly available. We show in Figure \ref{fig:annotation_interface} a screenshot for the \texttt{Toyota} template.}

\clearpage
\subsection{Partial Replacements}
\label{appendix:partial_replacement_overlays}

\begin{figure}[H]
    \centering
    \includegraphics[width=\linewidth]{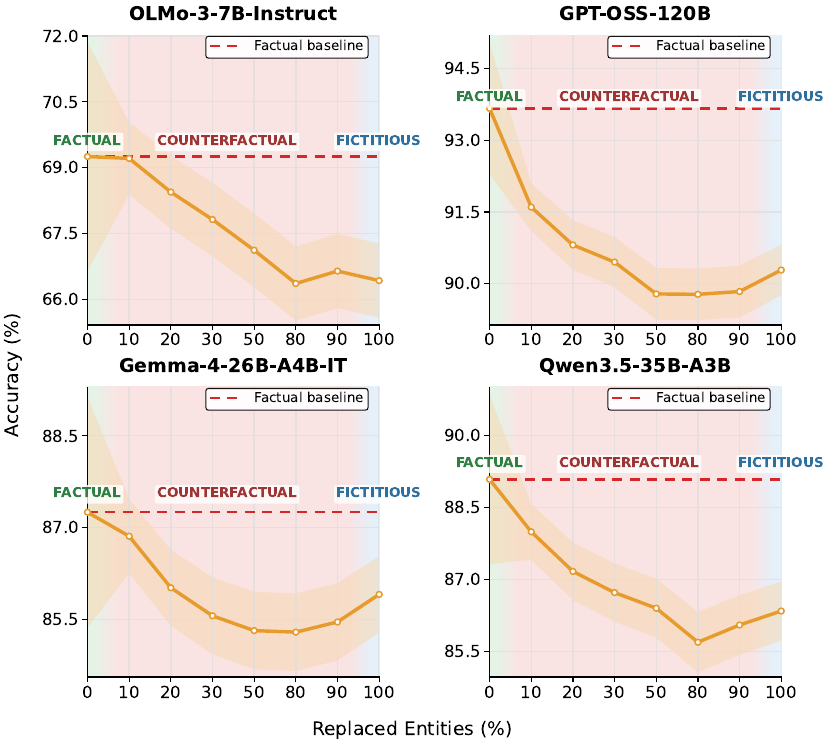}
    \caption{\scorerevision{Partial-replacement accuracy for the four models not shown in Figure~\ref{fig:accuracy_by_fictiobal_replacement_proportions}. Each point represents a setting where a given percentage of known entities have been replaced with fictitious ones, ranging from the factual baseline (0\%, dashed red lines) to the fully fictitious counterpart (100\%). Orange shaded bands denote the 95\% confidence intervals.}}
    \label{fig:accuracy_by_fictional_replacement_proportions_all_models}
\end{figure}

\postreviewrevision{Figures~\ref{fig:accuracy_by_fictiobal_replacement_proportions} and~\ref{fig:accuracy_by_fictional_replacement_proportions_all_models} together show the \greennumber{\ResultPartialModelCount{}} complete partial-replacement sweeps. \greenrevision{Six} curves recover \scorerevision{descriptively} from a \unchangednumber{50\%--80\%} minimum toward the fully fictitious endpoint, by \scorerevision{\ResultPartialClearRecoveryMin{}}\greennumber{--\ResultPartialRecoveryMax{}} points; \unchangednumber{OLMo-3-7B-Instruct} changes by only \unchangednumber{\ResultPartialRecoveryMin{}} points and is better described as plateauing. Thus, the mixed-context pattern is common but not universal in magnitude, \scorerevision{and these point-estimate recoveries alone establish neither statistical significance nor a causal mechanism.}}

\greenrevision{Figure~\ref{fig:accuracy_by_fictiobal_replacement_proportions} includes the complete partial-replacement sweep for \texttt{Claude Sonnet \unchangednumber{4.6}}. Accuracy falls from \greennumber{\ResultClaudePartialFactual\%} in the factual setting to \scorerevision{\ResultClaudePartialMinimum\%} at \greennumber{\ResultClaudePartialMinimumPercent\%} replacement, then recovers by \scorerevision{\ResultClaudePartialRecovery} points at the fully fictitious endpoint.}

\scorerevision{For partial replacements, deterministic accepted-answer rules supplement EM/JM evaluation using reference answers derived from the evaluated documents.}

\clearpage
\section{Qualitative Analysis}
\label{appendix:qualitative_analysis}

\subsection{\postreviewrevision{Qualitative Thinking-Checkpoint Contrast over \texorpdfstring{\Fictitiousterm{}}{Fictitious} Contexts}}
\label{appendix:qualitative_cot}
\definecolor{cotblue}{HTML}{DCEBFF}
\definecolor{cotgreen}{HTML}{DFF6E4}
\definecolor{cotred}{HTML}{FFE4E1}
\definecolor{cotorange}{HTML}{FFE8C2}
\definecolor{cotgray}{HTML}{F6F6F6}

\begin{tcolorbox}[
  breakable,
  colback=white,
  colframe=black!70,
  boxrule=0.8pt,
  arc=2mm,
  title={Qualitative analysis: chain-of-thought helps \fictitiousterm{} reasoning},
  fonttitle=\bfseries
]
\label{example:cot_vs_nocot}
\textbf{Question.}
How many years passed between the liberation of \colorbox{cotblue}{Velanthor}
and the formation of the \colorbox{cotblue}{Sovereign Pact Stability Council}?

\vspace{0.4em}
\textbf{Relevant document evidence.}
[...] Following its liberation in \colorbox{cotgreen}{1938}, the short-lived
Tessarine Commonwealth was established and later dissolved over the course of
the Telvoran Conflict. [...] The current \colorbox{cotblue}{Sovereign Pact Stability Council}, a
semi-presidential system, was formed in \colorbox{cotgreen}{1949} by Tarvien
Molquesp.
\vspace{0.4em}\\

\textbf{Gold answer.}
\colorbox{cotorange}{$1949 - 1938 = 11$}

\vspace{0.6em}
\begin{tcolorbox}[colback=cotred, colframe=red!45, title={OLMo-3-7B-Instruct}]
\textbf{Model response:} \texttt{ANSWER: 9}

\textbf{Outcome:} Incorrect.
\end{tcolorbox}

\vspace{0.4em}
\begin{tcolorbox}[colback=cotgray, colframe=green!45!black, title={OLMo-3-7B-Think}]
\small
\textbf{Model chain-of-thought and answer:}

Okay, let's see. The question is asking how many years passed between the
liberation of Velanthor and the formation of the Sovereign Pact Stability
Council.

First, I need to find the dates mentioned in the document for both events.
Let's start with the liberation. The document says, ``Following its liberation
in \colorbox{cotgreen}{1938}\ldots'' So the liberation happened in
\colorbox{cotgreen}{1938}.

Now, the Sovereign Pact Stability Council was formed in
\colorbox{cotgreen}{1949}. Wait, the exact phrase is ``formed in
\colorbox{cotgreen}{1949} by Tarvien Molquesp.'' Hmm, the year here is written
as 1949. That might be a typo or maybe it's supposed to be 1949? Usually, years
are four digits, so 1949 makes sense as 1949. So the council was formed in
1949, which I'll take as 1949.

So the time between 1938 and 1949. Let me calculate the difference. From 1938
to 1949 is \colorbox{cotorange}{11 years}. Because
\colorbox{cotorange}{$1949 - 1938 = 11$}.

Wait, but I should check if there's any other dates involved. The liberation is
clearly 1938. The formation of the council is 1949. The question is
straightforward subtraction. The answer should be 11 years.

I need to make sure there's no other events in between that might affect the
count. The document mentions the collapse in 1931, then the Commonwealth in
1938? Wait no, let me re-read that part.

The liberation was in 1938. Then the next part says ``the short-lived Tessarine
Commonwealth was established and later dissolved over the course of the
Telvoran Conflict.'' The Telvoran Conflict's timeline isn't given, but the next
event mentioned is the formation of the Sovereign Pact in 1949. So between 1938
and 1949, that's 11 years.

Yes, so 1949 minus 1938 is 11. The answer is 11. The document doesn't mention
any other years in between, so I think that's it. Just subtract the two years
given. So the answer should be 11.

\medskip
\textbf{Final answer:} \texttt{ANSWER: 11}

\textbf{Outcome:} Correct.
\end{tcolorbox}

\end{tcolorbox}

\section{Responsible Release}
\label{appendix:limitations_release}

The release contains benchmark examples and evaluation code, not a trained model. We anonymize publication identifiers, remove annotator names from released data, include a canary field in dataset rows, and publish the dataset as explicit splits so users can distinguish factual, \fictitiousterm{}, partial-replacement, and temporal-perturbation settings. The main intended use is diagnostic evaluation of contextual faithfulness; because the dataset contains \fictitiousterm{} passages, users should avoid presenting individual \fictitiousterm{} documents as real-world claims.

\section{Prompts}
\label{appendix:prompts}

\subsection{Evaluation System Prompt}
\label{appendix:evaluation_system_prompt}

\begin{figure}[H]
    \begin{tcolorbox}[title=Evaluation System Prompt]
    \ttfamily\small
    \raggedright
    \obeylines
    You answer questions using only the provided document.

      Rules:
    
      - Use only the document. Do not use outside knowledge.
    
      - Output exactly one line in the format: ANSWER: <answer>
    
      - Do not explain your reasoning.
    
      - Do not output a calculation or equation.
    
      - Do not restate the question.
    
      - Do not add any extra text before or after the answer line.
    
      - If the document does not determine the answer, output exactly: ANSWER: Cannot be determined
    
      - Otherwise, return only the shortest final answer.
    \end{tcolorbox}
\end{figure}

\subsection{Taxonomy Specification Prompt}
\label{appendix:taxonomy_prompt}

\begin{tcolorbox}[title=Taxonomy Specifications, breakable]
\footnotesize
\textbf{List of entity types with their associated attributes:}
\begin{itemize}[leftmargin=1.5em]
    \item person\_ID
    \item place\_ID
    \item event\_ID
    \item military\_org\_ID
    \item entreprise\_org\_ID
    \item ngo\_ID
    \item government\_org\_ID
    \item educational\_org\_ID
    \item media\_org\_ID
    \item temporal\_ID
    \item number\_ID
    \item award\_ID
    \item legal\_ID
    \item product\_ID
\end{itemize}

Where ID is a unique identifier that is given to each entity type instance that is mentioned in the document. This identifier should be used consistently across all the references of the same entity in the document.

\vspace{0.5em}
\textbf{List of attributes for each entity type:}

\vspace{0.5em}
\textbf{person\_ID} \\
(Individual human being mentioned in the document)
\begin{itemize}[leftmargin=1.5em]
    \item full\_name: Complete name of the person (first, middle, and/or last). Example: [John Smith; person\_1.full\_name], [Marie Curie; person\_1.full\_name]
    \item first\_name: Person's given name. Example: [John; person\_1.first\_name], [Marie; person\_1.first\_name]
    \item last\_name: Person's family name. Example: [Smith; person\_1.last\_name], [Curie; person\_1.last\_name]
    \item age: Person's age in years. Example: [25; person\_1.age], [45; person\_1.age]
    \item gender: Person's gender. Example: [male; person\_1.gender], [female; person\_1.gender]
    \item nationality: Person's nationality or citizenship. Example: [American; person\_1.nationality], [French; person\_1.nationality]
    \item ethnicity: Person's ethnic or cultural background. Example: [Hispanic; person\_1.ethnicity], [Asian; person\_1.ethnicity]
    \item subj\_pronoun: Subject pronoun (he/she/they). Example: [he; person\_1.subj\_pronoun], [she; person\_1.subj\_pronoun]
    \item obj\_pronoun: Object pronoun (him/her/them). Example: [him; person\_1.obj\_pronoun], [her; person\_1.obj\_pronoun]
    \item poss\_det\_pronoun: Possessive determiner (his/her/their). Example: [his; person\_1.poss\_det\_pronoun], [her; person\_1.poss\_det\_pronoun]
    \item poss\_pro\_pronoun: Possessive pronoun (his/hers/theirs). Example: [his; person\_1.poss\_pro\_pronoun], [hers; person\_1.poss\_pro\_pronoun]
    \item refl\_pronoun: Reflexive pronoun (himself/herself/themselves). Example: [himself; person\_1.refl\_pronoun], [herself; person\_1.refl\_pronoun]
    \item honorific: Title or honorific. Example: [Mr.; person\_1.honorific], [Ms.; person\_1.honorific]
    \item relationship: Special attribute for relationships between people (use relationship.person\_Y). Example: [mother; person\_1.relationship], [brother; person\_1.relationship]
    \item middle\_name: Person's middle name. Example: [Marie; person\_1.middle\_name], [Fitzgerald; person\_1.middle\_name]
\end{itemize}

\textbf{place\_ID} \\
(Geographic location or landmark)
\begin{itemize}[leftmargin=1.5em]
    \item city: Name of a city or town. Example: [New York; place\_1.city], [Paris; place\_1.city]
    \item region: Geographic region or area. Example: [New England; place\_1.region], [Midwest; place\_1.region]
    \item state: State or province. Example: [California; place\_1.state], [Ontario; place\_1.state]
    \item country: Nation or country. Example: [United States; place\_1.country], [France; place\_1.country]
    \item street: Street name or address. Example: [Main Street; place\_1.street], [5th Avenue; place\_1.street]
    \item natural\_site: Natural landmark or feature. Example: [Mount Everest; place\_1.natural\_site], [Amazon River; place\_1.natural\_site]
    \item continent: An entity referring to a continent. Example: [Europe; place\_1.continent], [Africa; place\_1.continent]
    \item demonym: The demonym of a place entity. Example: [Syrian; place\_1.demonym], [African; place\_1.demonym]
\end{itemize}

\textbf{event\_ID} \\
(Named event or occurrence)
\begin{itemize}[leftmargin=1.5em]
    \item name: Name of the event. Example: [World War II; event\_1.name], [Olympic Games; event\_1.name]
    \item type: Type or category of event. Example: [war; event\_1.type], [conference; event\_1.type]
\end{itemize}

\textbf{military\_org\_ID} \\
(Military organization)
\begin{itemize}[leftmargin=1.5em]
    \item name: Name of the military organization. Example: [IRA; military\_org\_1.name], [Royal Guard Command; military\_org\_1.name]
\end{itemize}

\textbf{entreprise\_org\_ID} \\
(Private/company organization (enterprise))
\begin{itemize}[leftmargin=1.5em]
    \item name: Name of the enterprise organization. Example: [Google; entreprise\_org\_1.name], [BNP Paribas; entreprise\_org\_1.name]
\end{itemize}

\textbf{ngo\_ID} \\
(NGO / non-governmental organization)
\begin{itemize}[leftmargin=1.5em]
    \item name: Name of the NGO. Example: [Amnesty International; ngo\_1.name], [Médecins Sans Frontières; ngo\_1.name]
\end{itemize}

\textbf{government\_org\_ID} \\
(Government institution or agency)
\begin{itemize}[leftmargin=1.5em]
    \item name: Name of the government organization. Example: [U.S. State Department; government\_org\_1.name], [European Commission; government\_org\_1.name]
\end{itemize}

\textbf{educational\_org\_ID} \\
(Educational institution)
\begin{itemize}[leftmargin=1.5em]
    \item name: Name of the educational organization. Example: [Harvard University; educational\_org\_1.name], [Sorbonne Université; educational\_org\_1.name]
\end{itemize}

\textbf{media\_org\_ID} \\
(Media/journalism organization)
\begin{itemize}[leftmargin=1.5em]
    \item name: Name of the media organization. Example: [BBC; media\_org\_1.name], [Reuters; media\_org\_1.name]
\end{itemize}

\textbf{temporal\_ID} \\
(Time-related information (dates, times, etc.))
\begin{itemize}[leftmargin=1.5em]
    \item day: Day of the week. Example: [Monday; temporal\_1.day], [Friday; temporal\_1.day]
    \item date: Full date. Example: [January 1, 2024; temporal\_1.date], [2024-01-01; temporal\_1.date]
    \item year: Year. Example: [2024; temporal\_1.year], [1999; temporal\_1.year]
    \item month: Month name or number. Example: [January; temporal\_1.month], [12; temporal\_1.month]
    \item timestamp: Specific time or timestamp. Example: [3:30 PM; temporal\_1.timestamp], [15:30; temporal\_1.timestamp]
    \item day\_of\_month: Day number within a month. Example: [15; temporal\_1.day\_of\_month], [1st; temporal\_1.day\_of\_month]
\end{itemize}

\textbf{number\_ID} \\
(Numerical values and quantities)
\begin{itemize}[leftmargin=1.5em]
    \item int: Integer number. Example: [42; number\_1.int], [100; number\_1.int]
    \item str: Number expressed as text. Example: [forty-two; number\_1.str], [one hundred; number\_1.str]
    \item float: Decimal number. Example: [3.14; number\_1.float], [99.9; number\_1.float]
    \item fraction: Fractional value. Example: [1/2; number\_1.fraction], [three quarters; number\_1.fraction]
    \item percent: A number representing a percentage (must be between 0 and 100). Example: [20; number\_1.percent], [10; number\_1.percent]
    \item proportion: A number that represents a proportion (must be between 0.0 and 1.0). Example: [0.1; number\_1.proportion], [0.00012; number\_1.proportion]
\end{itemize}

\textbf{award\_ID} \\
(An entity that represents any kind of awards)
\begin{itemize}[leftmargin=1.5em]
    \item name
\end{itemize}

\textbf{legal\_ID} \\
(Legal or regulatory instrument (law, directive, regulation, treaty, policy framework))
\begin{itemize}[leftmargin=1.5em]
    \item name: Legal or regulatory instrument (law, directive, regulation, treaty, policy framework). Example: [Markets in Financial Instruments Directive 2014; legal\_1.name], [Capital Requirements Regulation; legal\_1.name]
    \item reference\_code: Official legal citation or identifier. Example: [2014/65/EU; legal\_1.reference\_code], [(EU) No 575/2013; legal\_1.reference\_code]
\end{itemize}

\textbf{product\_ID} \\
(An entity describing a product)
\begin{itemize}[leftmargin=1.5em]
    \item name: The name of the product as mentioned in the text. Example: [iPhone; product\_1.name], [Eliquis (apixaban); product\_1.name]
\end{itemize}
\end{tcolorbox}

    \label{prompt:taxonomy_specifications}

\subsection{Entity Annotation Prompt}
\label{appendix:entity_prompt}

\begin{figure}[H]
    \begin{tcolorbox}[title=Entity Annotation Prompt]
    \ttfamily\small
    \raggedright
    \obeylines
        \# Entity Annotator\\
    
        You are an annotation agent. Annotate the document, questions, and answers below using inline annotations only.\\
        
        Output format:\\
        
        Render the annotated document, questions, and answers in a YAML form with the following structure:\\
        
        \texttt{document\_id: <doc\_id>}\\
        \texttt{annotated\_document: |}\\
        \texttt{  <full document with inline annotations [span; entity.attr]>}\\
        \texttt{annotated\_questions:}\\
        \texttt{  - question\_id: <id>}\\
        \texttt{    question: <question with inline annotations>}\\
        \texttt{    question\_type: <extractive|arithmetic|temporal|inference>}\\
        \texttt{    answer: <answer\_expression>}\\
        \texttt{decision\_log:}\\
        \texttt{  - span: "<text span as it appears in document>"}\\
        \texttt{    action: ANNOTATE | KEEP\_LITERAL}\\
        \texttt{    reason\_code: SAFE\_TO\_TEMPLATE | OUT\_OF\_TAXONOMY | STRUCTURAL\_CRITICAL\_SPAN | CORE\_SEMANTIC\_CRITICAL\_SPAN | COUPLED\_DEPENDENCY | AMBIGUOUS}\\
        \texttt{    short\_reason: "<one short sentence>"}\\
        
        Here are a few rules to follow when annotating the document, questions, and answers:\\
        
        1. Do NOT modify the document text except by inserting inline annotations.\\
        
        2. Do NOT modify question/answer text except by inserting inline annotations.\\
        
        3. Inline annotation format MUST be exactly: \texttt{[span\_text; entity\_id.attribute]}\\
        
        4. After removing annotations (keeping only span\_text), the text must be identical to the original document.\\
        
        5. Use only the taxonomy below. If an annotation would violate it, do NOT annotate it.\\
        
        \{\{TAXONOMY\_SPECIFICATIONS\_PROMPT\ref{prompt:taxonomy_specifications}\}\}
    \end{tcolorbox}
    \caption{The prompt used to draft AI Agent pre-annotations to help human annotators.}
    \label{prompt:entity_annotation}
\end{figure}

\subsection{Rule Generation Prompt}
\label{appendix:rule_prompt}

\begin{tcolorbox}[title=Rule Generation Prompt,breakable]
\ttfamily\small
\raggedright
\obeylines

    Your are a rigorous rule generation agent and your task is to generate a list of rules based on an annotated document.

    Purpose of these rules:
    - The annotated template will later be converted into factual and fictional documents.
    - During fictionalization, the code replaces annotated entities with sampled fictional ones (i.e. non existing entities) and regenerates numbers, temporals, and gender-related attributes (e.g. pronouns, honorific, etc.).
    - The problem is that these new fictional/non-existing entities may not be suitable for the context of the document, we therefore need to define sampling rules/constraints.
    - Your task is to produce logical constraints (rules) that will be used to sample fictional entities to make sure that semantic/linguistic coherence is still maintained accros the document after the replacement.
    - The rule list tells the generator which document-specific semantic constraints must still hold after replacement.
    - Basically those rules serve as safe guards to avoid replacing with fictional entities that break the semantic and coherence of the document.
    - Without these rules, the generator could produce a fictional document that is no longer coherent.
    
    How to think about the task:
    - You are identifying the minimal set of explicit constraints that must be satisfied by the entity replacement.
    - A good rule protects a fact that is stated or directly implied by the document and that could otherwise break after replacement.
    - Focus specifically on the annotated entities (i.e. text spans between brackets with an entity type, ID, and attribute [text span; entity\_type\_ID.attribute] -- a detailed description of the entity taxonomy is provided below.)
    
    General notes:
    - Use only entities and attributes that are actually annotated in the document.
    - The code performs automatic generation for numbers, temporals, age, and gendered forms (pronouns, gender, honorific, etc.) and sampling from a pre-defined pool for the rest of attributes such as names.
    - Do not write rules whose only purpose is to preserve order for numbers, ages, dates, or years. This is already automatically handled by the code.
    - Never write pure chronology/order rules between temporal entities (for example `temporal\_2.year < temporal\_3.year`).
    - Never write pure order rules between number entities (for example `number\_1.int < number\_2.int`) or age entities (for example `person\_1.age < person\_2.age`) when the goal is only to preserve factual ordering.
    - Important: if the document implies an exact gap/offset, you should encode that relation explicitly (for example consecutive years: `temporal\_3.year - temporal\_2.year == 1`).
    - Century mentions are annotated as `number` entities, not `temporal` entities. When a century mention is semantically tied to an exact year/date and changing them independently could create an obvious contradiction, add an explicit century rule.
    - Allowed century helper functions:
      - `century\_of(temporal\_X.year)` or `century\_of(temporal\_X.date)` when an exact year/date must stay inside a stated century
      - `century\_start(number\_X.int)` and `century\_end(number\_X.int)` when a stated century must stay before/after an exact year/date
    - Typical century rules:
      - `century\_of(temporal\_6.year) == number\_8.int`
      - `century\_end(number\_9.int) < temporal\_7.year`
    - Crucial: Never use the factual values of entities but only use the actual entity reference, for example in the sentence *They reported [9; number\_3.int] deaths at the Oxford Avenue* never use the factual value 9 explicitly but rather use its associated reference `number\_3.int`. This applies to all the other entities/attributes such as temporals and `person.age`. This also applies to sums, don't force them to sum to a certain factual fixed value, but rather they must sum to the entity reference that is used to annotate the factual value. For example in the sentence *There were [9; number\_3.int] deaths and [3; number\_4.int] injured resulting in [12; number\_5.int] casualties.* never use the factual value 12 to represent the result of summing `number\_3.int` and `number\_4.int` but rather use its reference `number\_5.int` (`number\_3.int + number\_4.int == number\_5.int`).
    - Constants are allowed when they encode a structural relation (for example `== 1` for consecutive years).
    - Don't fix entities to a have specific value that is mentionned in the text (e.g. `number\_10.int == 10`), remember that we actually want to replace entities with different values so fixing them does make sense.
    - add a small concise comment next to each rule to quickly explain why, using inline `\#` format (e.g. `number\_5.int > 1 \# plural mention`)
    
    What a good rule captures:
    - an arithmetic relation that must stay true
    - an explicit comparison or bound
    - a document-specific dependency that would become incoherent if broken
    - a semantic constraint whose violation would make the fictionalized document inconsistent
    - an explicit interval/offset equation between temporals, numbers, or ages when the text implies a specific gap
    - a sum of numbers, temporals, or ages that must sum to a certain value
    - a century/year compatibility rule when a stated century must remain compatible with an exact year or date
    - if you notice that two entities are the same and should have the same values but are annotated with different entity IDs then indicate this with a rule (e.g. `number\_2.int == number\_3.int`)
    - a rule that indicates that a number should be plural because there is a plural mention in the text (e.g. )
    
    What should not become a rule:
    - generic number ordering
    - generic temporal ordering
    - string-content constraints over name fields
    - repeated-value matching just because two annotations happen to share the same factual value (you have to make sure that they are the same entities because sharing the same value doesn't mean that it's the same entities)
    - date decomposition rules that only restate that a full date includes its year or month
    - biography chronology chains that simply mirror the order in which events happened
    - anything based on outside knowledge
    - vague or speculative constraints
    - trivial restatements of the annotations
    - facts that the generation code already enforces automatically
    
    Bad rule patterns to avoid:
    - `temporal\_7.year < temporal\_8.year`
      - bad because it only encodes chronology.
    - `number\_1.int < number\_2.int`
      - bad because it only preserves factual ordering.
    - `person\_1.age < person\_2.age`
      - bad because it only preserves factual ordering.
    
    When to return no rule: if the only candidate rules are chronology/order rules, string-content rules over names, repeated-value rules, or date/number decomposition rules, return `rules: []`.
    
    In [2022; temporal\_4.year], [Nora; person\_1.name], who was [6; person\_1.age] years old, joined the [Oak Street kids' science; entreprise\_org\_1.name] club. In [2023; temporal\_5.year], a year after [she; person\_1.subj\_pronoun] joined the club, [her; person\_1.obj\_pronoun] older [brother; person\_2.relationship.person\_1] [Sam; person\_2.name] joined too. [Sam; person\_2.name] was [8; person\_2.age] years old, so [he; person\_1.subj\_pronoun] was [2; number\_7.int] years older than [Nora; person\_1.name]. During the spring tournament, the club won [3; number\_2.int] robotics rounds and [2; number\_3.int] quiz rounds, for a total of [5; number\_4.int] wins. At the summer fair, [Sam; person\_2.name] arrived with [4; number\_5.int] friends; together with the [3; number\_2.int] club children already waiting, their group included [7; number\_6.int] children. The club archives say a reform charter was written in the [19th; number\_8.int] century and officially adopted in [1875; temporal\_6.year]. A related dispute started in the [18th; number\_9.int] century and was settled in [1802; temporal\_7.year].
    
    Examples:
    - `number\_2.int + number\_3.int = number\_4.int`
    - `temporal\_5.year - temporal\_4.year == 1 \# consecutive calendar years`
    - `person\_2.age - person\_1.age == number\_7.int \# age gap stated in the document`
    - `number\_5.int > 1 \# plural mention`
    - `4 < person\_2.age < 12 \# age constraints for specific profiles, here for example the text mentions that a person is a child and kids`
    - `number\_6.int = 2 + number\_5.int`
    - `century\_of(temporal\_6.year) == number\_8.int \# the adoption year must stay inside the stated century`
    - `century\_end(number\_9.int) < temporal\_7.year \# the settlement year must come after the stated starting century`
    
    Inputs you will receive:
    - `document\_id`
    - `annotated\_document`
    
    OUTPUT FORMAT (STRICT YAML ONLY):
    ```yaml
    document\_id: <doc\_id>
    rules:
      - <rule\_expression> \# <short\_comment>
\end{tcolorbox}

\subsection{\texorpdfstring{\Fictitiousterm{}}{Fictitious} Entity Pool Generation Prompt}
\label{appendix:fictional_pool_prompt}

The \fictitiousterm{} entity pool prompt instructs the generation agent to produce document-specific pools of non-existing named entities, while leaving numbers, dates, years, pronouns, and other automatically generated fields to the deterministic generator. \revision{Prompts are reproduced verbatim; their original terminology is retained to preserve the exact experimental record.}

\begin{tcolorbox}[title=\Fictitiousterm{} Pool Generation Prompt, breakable]
\begin{lstlisting}[basicstyle=\ttfamily\scriptsize,breaklines=true,columns=fullflexible,keepspaces=true]

    You generate document-specific pools of fictional named-entity variants for benchmark construction.
    
    Purpose:
    - The benchmark replaces factual entities with fictional ones and then generates fictional document versions from those pools.
    - Each required entity reference must receive its own list of fictional variants.
    - The same document variant index is shared across all references generated in one response.
    - If rules connect several references in this batch, variant `1` for all of those references must jointly satisfy the rules, variant `2` must jointly satisfy the rules, and so on up to variant `15`.
    - Your output is not the final fictional document. It is the reference-specific candidate pool that later code will sample from.
    
    What must stay outside the pool:
    - Do NOT generate numbers.
    - Do NOT generate dates, years, weekdays, months, timestamps, or any other temporals.
    - Do NOT generate person gender attributes, pronouns, honorifics, or relationship fields.
    - Those values are generated later by Python code with fixed seeds and rule checking.
    
    Why the rules matter here:
    - The rules describe constraints that the later document generator must preserve.
    - Your pool should make those constraints easy to satisfy.
    - Example: if a rule requires `person_1.nationality == place_2.demonym`, generate person nationalities and place demonyms that can be matched cleanly.
    - Example: if a legal item needs a `name` and a `reference_code`, generate values that plausibly belong to the same fictional legal instrument.
    - Ignore rules that only concern numbers, temporals, or other automatically generated fields.
    - When you invent a place and a corresponding demonym or nationality adjective, they must belong to the same invented place.
    - Do not pair a country, city, region, or state with a demonym that clearly belongs to some other invented place.
    - If a place variant contains both `demonym` and `nationality`, keep them aligned unless the document explicitly requires different surface forms.
    
    Entity taxonomy reference for this prompt:
    - The list below is injected from the project taxonomy so the entity meanings and attribute meanings match the rest of the codebase exactly.
    {{ENTITY_TAXONOMY_REFERENCE}}
    
    Pool construction requirements:
    - Every invented value must be fictional.
    - In this benchmark, "fictional" means "non-existing": the value must not refer to a real entity that already exists.
    - Treat any candidate that appears on Wikipedia or in a web search as invalid for this task.
    - Before finalizing the pool, explicitly check whether the names you generated correspond to real entities. If your environment gives you web search or browsing tools, use them. If it does not, regenerate any candidate that seems plausibly real or widely used.
    - Do NOT use ordinary attested first names, surnames, city names, country names, demonyms, award names, legal names, or organization names.
    - Common human names are invalid even if you combine them with other fictional fields.
    - If a candidate looks like a standard French, English, Spanish, Portuguese, German, Italian, Arabic, Slavic, or otherwise attested real-world name, reject it and invent a new one.
    - Bad examples of invalid outputs: `Eliane`, `Lucien`, `Margaux`, `Renaud`, `Marcelo`, `Gaston`, `Adrienne`, `Henrik`, `Mallaby`, `Stembridge`, `Redwick`, `Marbleton`.
    - Good outputs should feel pronounceable but unattested: they should read like plausible names while still looking clearly invented.
    - Every invented name must be pronounceable for an English speaker.
    - Follow ordinary English phonotactics. Avoid impossible consonant clusters, unreadable punctuation, excessive doubled letters, and fantasy-style spellings.
    - Use ASCII only.
    - Avoid accents, diacritics, emoji, and decorative punctuation.
    - Avoid names that obviously match real well-known entities.
    - Avoid names that are only tiny edits of famous real entities.
    - Avoid slight edits of ordinary real first names or surnames.
    - Avoid ordinary English surname or town-style endings such as `-ton`, `-bridge`, `-wick`, `-bury`, `-ford`, or `-by` when the result looks like an attested real place or family name.
    - Avoid slight edits of real demonyms, historical labels, dynasties, eras, empires, or treaty names.
    - Do not use standalone Roman numerals or ordinal dynastic labels as entity names.
    - Do not invent names by taking a real root and adding a thin suffix like `-an`, `-ian`, `-ish`, `-ic`, or `-a`.
    - Do not use generic suffix-marker tokens such as `Alt`, `Astra`, `Nova`, `Prime`, or `Sigma` anywhere in generated values.
    - Avoid joke names, placeholders, and nonsense strings.
    - Keep the pool diverse. Do not output many near-duplicates that only differ by one letter or one generic suffix.
    - If a place entry contains several attributes, they must be internally coherent inside the same object.
    - If a place entry contains both a place name and a demonym, the demonym must clearly match that exact fictional place.
    - If an organization entry is tagged with a subtype, the name must sound plausible for that subtype.
    
    Semantic consistency guardrails (critical):
    - Do not generate names that semantically contradict explicit cues in the document text.
    - Preserve ideological/polarity cues when those cues are explicitly stated.
    - Avoid lexical markers that imply the opposite of the described role.
    - If the text gives a clear stance, mission, alignment, or institutional function, generated names must remain compatible with that context.
    - Use these examples as strict constraints:
      - If a political party is described as left-wing/progressive, do not generate a name that strongly signals right-wing/ultra-conservative/monarchist alignment.
      - If a party is described as conservative/right-wing, do not generate a name that strongly signals socialist/leftist alignment.
      - If an organization is described as humanitarian, relief-focused, or pacifist, do not generate a militaristic or combat-framed name.
      - If an organization is described as environmental/climate-focused, do not generate a name that suggests fossil-fuel expansion or anti-environment positioning.
      - If an entity is a court, ministry, or regulator, do not generate a name that sounds like a private company brand.
      - If an entity is a company/commercial operator, do not generate a name that sounds like a government ministry or tribunal.
      - If an entity is a university/school, do not generate a name that sounds like a bank, military unit, or political party.
      - If a media outlet is described as local/regional, do not generate a name implying global or official state-agency status unless the text supports it.
      - If the document is about countries, empires, wars, revolutions, or historical periods, do not output names that look like real-world historical labels or near-variants such as `Frankish`, `Gallican`, `Europan`, `Bourbon Restoration`, `Golden Era`, or bare numerals like `III`.
    
    Reference-level requirements:
    - allocate {{TARGET_CANDIDATES_PER_ENTITY}} fictional candidates for each unique entity reference requested below.
    - Generate exactly {{TARGET_CANDIDATES_PER_ENTITY}} fictional variants for every required entity reference shown below.
    - Treat the requested reference ids as opaque keys for this call. Do not rename them, renumber them, substitute different ids from the document, or output a nearby reference id.
    - If this call requests `place_7` and `place_8`, then output exactly `place_7` and `place_8`. Outputting `place_2`, `place_4`, or any other reference id is invalid.
    - Do not merge entity references together, even when they share the same type.
    - Keep variants globally distinct across entity references of the same bucket. Do not reuse the same fictional person/place/event/etc. under two different reference ids.
    - Only include the attributes required for that specific reference. Do not add unrelated optional attributes.
    - For organization entities, write values into the taxonomy bucket that matches the annotation type.
    - Within one response, list order matters: `variants[0]` across linked references describes one coherent fictional document version, `variants[1]` describes another, and so on.
    - When linked references involve nationality or demonym fields, keep the same variant index aligned so the person/place pair still matches at that index.
    
    Inputs:
    - `document_id`: {{DOCUMENT_ID}}
    - `document_theme`: {{DOCUMENT_THEME}}
    - `annotated_document_excerpt_for_this_call`:
    {{ANNOTATED_DOCUMENT}}
    
    - `requested_reference_mentions`:
    {{REQUESTED_REFERENCE_MENTIONS}}
    
    - `rules_relevant_to_pool_generation`:
    {{POOL_RELEVANT_RULES}}
      - Treat this as the actionable rule subset for pool construction.
    
    - `required_entities_summary`:
    {{REQUIRED_ENTITIES_SUMMARY}}
    
    - `reference_ids_for_this_call`:
    {{REFERENCE_IDS_FOR_THIS_CALL}}
    
    - `reference_target_counts`:
    {{REFERENCE_TARGET_COUNTS}}
    
    Output format (STRICT YAML ONLY):
    ```yaml
    persons:
      person_1:
        required_attributes:
          - full_name
          - first_name
        count: 15
        variants:
          - full_name: <fictional full name>
            first_name: <fictional first name>
      person_2:
        required_attributes:
          - full_name
        count: 15
        variants:
          - full_name: <fictional full name>
    places:
      place_1:
        required_attributes:
          - country
          - demonym
        count: 15
        variants:
          - country: <fictional country>
            demonym: <fictional demonym>
    events:
      event_1:
        required_attributes:
          - name
        count: 15
        variants:
          - name: <fictional event name>
            type: <event type if needed>
    military_orgs:
      military_org_1:
        required_attributes:
          - name
        count: 15
        variants:
          - name: <fictional military organization name>
    entreprise_orgs:
      entreprise_org_1:
        required_attributes:
          - name
        count: 15
        variants:
          - name: <fictional enterprise organization name>
    ngos:
      ngo_1:
        required_attributes:
          - name
        count: 15
        variants:
          - name: <fictional NGO name>
    government_orgs:
      government_org_1:
        required_attributes:
          - name
        count: 15
        variants:
          - name: <fictional government organization name>
    educational_orgs:
      educational_org_1:
        required_attributes:
          - name
        count: 15
        variants:
          - name: <fictional educational organization name>
    media_orgs:
      media_org_1:
        required_attributes:
          - name
        count: 15
        variants:
          - name: <fictional media organization name>
    awards:
      award_1:
        required_attributes:
          - name
        count: 15
        variants:
          - name: <fictional award name>
    legals:
      legal_1:
        required_attributes:
          - name
          - reference_code
        count: 15
        variants:
          - name: <fictional legal entity name>
            reference_code: <fictional legal code if needed>
    products:
      product_1:
        required_attributes:
          - name
        count: 15
        variants:
          - name: <fictional product name>
    ```
    
    Output rules:
    
    Return YAML only.
    Output only the buckets and entity references requested in required_entities_summary.
    Output only the reference ids listed in reference_ids_for_this_call. Any other id is invalid.
    Keep only relevant keys in each variant object. Do not write null values.
    For organization buckets, write only {name: ...} entries inside each variant. The bucket name already carries the taxonomy type.
    Set count to the number of distinct valid variants you actually provide for that reference.
    Respect reference_target_counts and aim to make every requested count equal 15.
    Use the rules to make the pool compatible with the later replacement step.
    Prefer varied, clean fictional values over tiny spelling variations of the same item.
\end{lstlisting}
\end{tcolorbox}

\end{document}